\pdfoutput=1 
\documentclass[10pt,journal,compsoc]{IEEEtran}
 
\usepackage{amsmath,amssymb,amsfonts}
\usepackage{algorithm}
\usepackage{algorithmic}
\usepackage{array}
\usepackage[caption=false,font=normalsize,labelfont=sf,textfont=sf]{subfig}
\usepackage{textcomp}
\usepackage{stfloats}
\usepackage{url}
\usepackage{graphicx}
\usepackage{cite}
\usepackage{booktabs}
\usepackage{multirow}
\usepackage{CJKutf8} 
\usepackage[table]{xcolor}
\usepackage{enumitem} 
\usepackage{verbatim} 
\usepackage{bbding} 
\usepackage{pifont}
\usepackage[hidelinks]{hyperref} 

\definecolor{mygray}{RGB}{230,230,230}

\usepackage[utf8]{inputenc}        
\usepackage[LGR,T2A,T1]{fontenc}   
\usepackage[greek,russian,english]{babel} 
\usepackage{newunicodechar}
\DeclareUnicodeCharacter{FFFD}{?}

\newlength\savedwidth 
\newcommand\whline{\noalign{\global\savedwidth\arrayrulewidth
                            \global\arrayrulewidth 1.5pt}
                   \hline
                   \noalign{\global\arrayrulewidth\savedwidth}}
                   
\newlength\savewidth

\begin{document} 

\title{Weight Patching: Toward Source-Level Mechanistic Localization in LLMs}

\author{Chenghao~Sun,
        Chengsheng~Zhang,
        Guanzheng~Qin,
        Rui~Dai,
        and~Xinmei~Tian%
\IEEEcompsocitemizethanks{%
\IEEEcompsocthanksitem All authors are with the Key Laboratory of Ministry of Education for Brain Inspired Intelligent Perception and Cognition, University of Science and Technology of China, Hefei, Anhui 230026, China.
\IEEEcompsocthanksitem Corresponding author: Xinmei Tian (e-mail: xinmei@ustc.edu.cn).
\IEEEcompsocthanksitem This work has been submitted to the IEEE for possible publication. Copyright may be transferred without notice, after which this version may no longer be accessible. 
}
}

\markboth{Preprint submitted to IEEE Transactions on Pattern Analysis and Machine Intelligence}%
{Sun \MakeLowercase{\textit{et al.}}: Weight Patching: Toward Source-Level Mechanistic Localization in LLMs}

\IEEEtitleabstractindextext{%
 
\begin{abstract}
Mechanistic interpretability seeks to localize model behavior to the internal components that causally realize it.
Prior work has advanced activation-space localization and causal tracing, but modules that appear important in activation space may merely aggregate or amplify upstream signals rather than encode the target capability in their own parameters.
To address this gap, we propose Weight Patching, a parameter-space intervention method for source-oriented analysis in paired same-architecture models that differ in how strongly they express a target capability under the inputs of interest.
Given a base model and a behavior-specialized counterpart, Weight Patching replaces selected module weights from the specialized model into the base model under a fixed input.
We instantiate the method on instruction following and introduce a framework centered on a vector-anchor behavioral interface that provides a shared internal criterion for whether a task-relevant control state has been formed or recovered in open-ended generation.
Under this framework, the analysis reveals a hierarchy from shallow candidate source-side carriers to aggregation and routing modules, and further to downstream execution circuits. 
The recovered component scores can also guide mechanism-aware model merging, improving selective fusion across the evaluated expert combinations and providing additional external validation.

\end{abstract}

}

\maketitle
\IEEEdisplaynontitleabstractindextext

 
\section{Introduction}
\label{sec:introduction}  
\IEEEPARstart{M}{echanistic} interpretability aims to localize model behavior to the internal components that causally realize it, rather than only describing correlations in activations or outputs~\cite{olah2020zoom,geiger2023causal_abstraction}. Previous work in large language models (LLMs)  has made substantial progress in activation-space localization and causal tracing, including path-level analysis and circuit discovery~\cite{activation_patching,path_patching,circuit_discovery_nips2023,syed2024attribution_patching_outperforms}. Yet if the goal is to determine where a target capability is actually written in parameters, activation-space evidence alone is not enough. A module may appear critical because it aggregates or amplifies decisive upstream signals, even when the capability is not primarily implemented in that module’s own parameters~\cite{lan2024shared_circuits,heimersheim2024interpret_activation_patching,zhang2024best_practices_activation_patching}. 
This distinction between activation-side importance and parameter-side implementation matters not only for mechanistic explanation, but also for downstream parameter-space analyses and operations that depend on identifying capability-relevant components~\cite{nips2022locating_knowledge,wortsman2022model_soups,matena2022fisher_merging,yadav2023ties,rame2023model_ratatouille,Yu2023LanguageMA,goddard2024mergekit}.

\begin{figure}[t]
    \centering
    \includegraphics[width=0.95\linewidth]{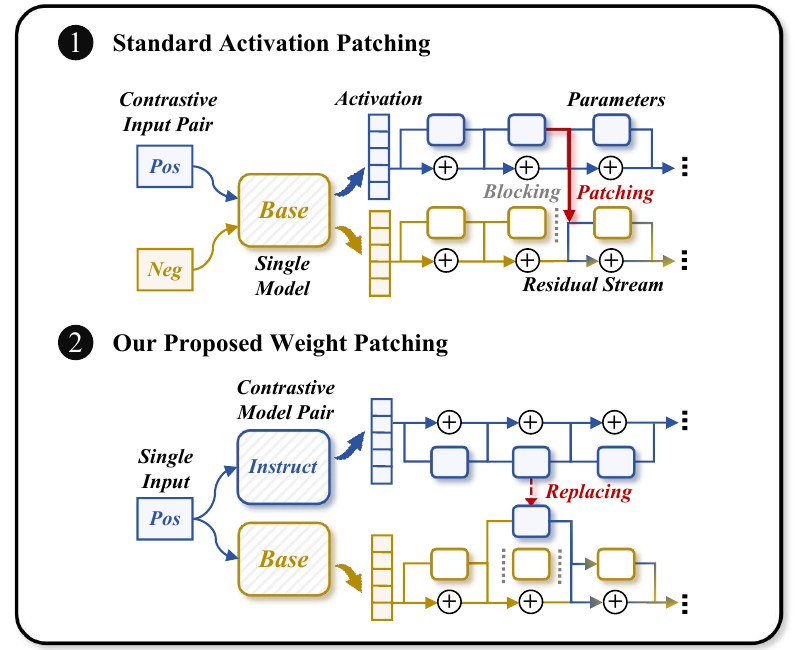}
    \caption{Weight Patching versus standard Activation Patching. Standard activation patching intervenes on activations under clean/corrupted inputs, whereas Weight Patching intervenes on parameters under a fixed input. In our instruction-following analyses, the activation-side companion intervention uses a fixed-input cross-model form with the same restoration logic.}
    \label{fig:wp-vs-ap}
\end{figure}

Framed this way, the problem becomes source-oriented: it shifts attention from inference-time signal flow to the parameter differences introduced by training or post-training. In same-architecture models, differences in how strongly a target capability is expressed are realized through such parameter differences, making paired checkpoints with different levels of capability expression a natural object of analysis. 
Here, source-level localization refers to identifying candidate source-side carriers under the paired-model setting and head-/neuron-level intervention granularity—that is, components whose specialized parameter slices help recover a target internal control state when transplanted into the base model.
We therefore propose \emph{Weight Patching}, a parameter-space intervention method for source-level mechanistic localization in LLMs. Given a base model and a behavior-specialized counterpart, it replaces selected module weights under a fixed input and measures whether the transplanted parameters recover a capability-relevant internal state.
Because exhaustive exact intervention is expensive at fine granularity, we further develop a gradient-based first-order approximation that makes fine-grained screening practical at model scale.

We instantiate Weight Patching on instruction following because it provides a natural setting for studying how post-training writes control into model parameters. 
In this setting, post-training substantially affects how natural-language instructions are converted into internal control and used to guide downstream generation, making it a useful testbed for asking which specialized parameters most directly support the resulting capability difference. 
Recent work has shown that alignment induces structured control directions, steerable behavioral signals, and systematic routing changes~\cite{llm_know_instruction_iclr2024,task_vector_head_iclr2024,task_vector_iclr2025,steering_2024,repe_zou2023,instr_behavior_change_naacl2024}, but these studies mainly characterize representations or inference-time activations rather than the parameter subsets that carry the resulting specialization. A further challenge is that the relevant control must be established and maintained across generations, while standard token-level utilities often do not provide a stable criterion for determining whether that state has been formed and sustained.

\begin{figure}[!ht]
    \centering
    \includegraphics[width=0.95\linewidth]{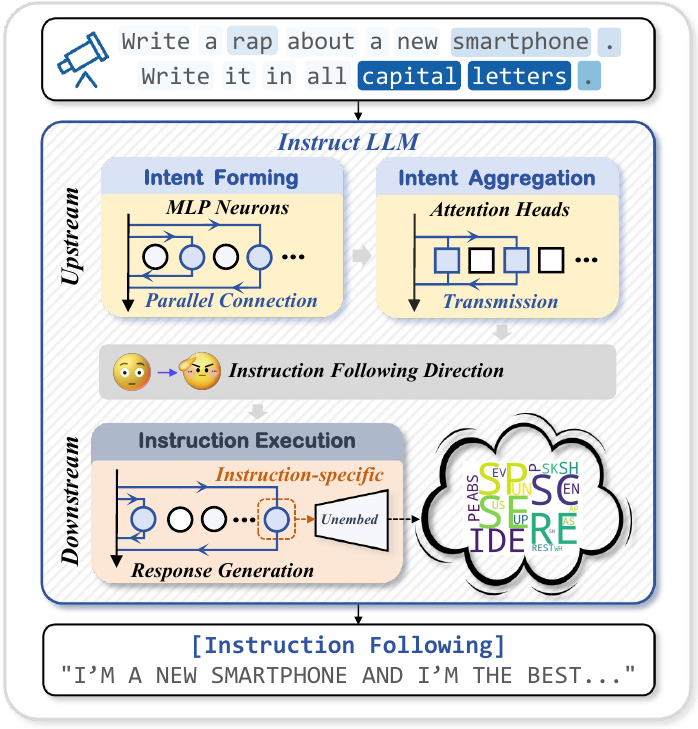}
    \vspace{-5pt}
    \caption{Mechanistic summary of instruction following under the proposed framework. 
    The recovered hierarchy suggests that shallow components act as candidate source-side carriers of instruction-conditioned control features, mid-layer attention heads aggregate and route them into an instruction-following direction, and downstream instruction-specific modules help unfold the recovered control state into response generation.
    }
    \label{fig:if_hierarchy}
    \vspace{-5pt}
\end{figure}

To make this problem operational for generative behaviors, we introduce a framework centered on a vector-anchor behavioral interface that provides a shared internal criterion for whether a task-relevant control state has been formed or recovered. Under task-aligned instruction-following inputs, this criterion lets us evaluate paired-model interventions without requiring the paired models to differ only in the studied capability. Within this framework, Weight Patching tests candidate source-side carriers in parameter space, while activation-side analyses identify where the recovered state is aggregated, routed, and translated into output behavior. Together, these analyses reveal a hierarchy from shallow source-side carriers to mid-layer aggregation/routing modules and downstream execution circuits, as summarized in Fig.~\ref{fig:if_hierarchy}.
 
A further benefit of this analysis is that the recovered component scores can be reused for mechanism-aware model merging, providing structured priors for component-wise fusion rather than uniform averaging. Empirically, this reuse yields the strongest overall average performance across the evaluated expert combinations.

In summary, our contributions are four-fold as follows:
\begin{itemize}
    \item We propose Weight Patching, a parameter-space intervention method for source-level mechanistic localization in LLMs, together with a gradient-based approximation that makes fine-grained screening practical.
    
    \item We introduce a framework for mechanistic analysis of generative behaviors, in which a vector-anchor behavioral interface provides a shared internal criterion for open-ended generation.
    
    \item Under this framework, we recover a hierarchical mechanistic account of instruction following, from shallow source-side carriers to aggregation and routing modules and downstream execution circuits.
    
    \item We demonstrate that the recovered source-level scores can be leveraged for mechanism-aware model merging, where they guide component-wise expert fusion to yield superior overall performance across the evaluated expert combinations.
\end{itemize}

  \section{Preliminaries}
\label{sec:preliminaries}

\subsection{Transformer Components in a Residual-Stream Coordinate System}
\label{sec:prelim_components}

We study decoder-only transformers through a residual-stream view, in which modules write additively into a shared residual-stream space. 
This shared residual stream provides a common coordinate system for both activation-space and parameter-space analysis; throughout the paper, attention heads and individual MLP neurons serve as the basic component-level intervention units.
 Let the input sequence be $x=(x_1,\dots,x_T)$, with input embeddings $\mathbf Z_0=\mathrm{Embed}(x)\in\mathbb{R}^{T\times d_{\text{model}}}$. Under a pre-norm decoder, layer $l\in\{1,\dots,L\}$ updates the residual stream as
\begin{align}
\mathbf Z'_l &= \mathbf Z_{l-1} + \mathrm{Attn}_l\!\bigl(\mathrm{Norm}(\mathbf Z_{l-1})\bigr),\\
\mathbf Z_l  &= \mathbf Z'_l + \mathrm{MLP}_l\!\bigl(\mathrm{Norm}(\mathbf Z'_l)\bigr),
\end{align}
where $\mathbf Z_l\in\mathbb{R}^{T\times d_{\text{model}}}$ denotes the residual-state matrix after layer $l$, and $\mathbf z_l^{(t)}\in\mathbb{R}^{d_{\text{model}}}$ denotes its row at position $t$.

Within this shared residual-stream space, an attention head and an MLP neuron can both be treated as decomposable write-back units.
For head $H^{(l,h)}$, let $\mathbf O^{(l,h)}$ denote its head output; its contribution to the residual stream is $\Delta\mathbf Z_{\mathrm{attn}}^{(l,h)}=\mathbf O^{(l,h)}\mathbf W_O^{(l,h)}$. For the SwiGLU MLP, neuron $N^{(l,j)}$ induces a write-back term $\Delta\mathbf Z_{\mathrm{mlp}}^{(l,j)}=\mathbf a^{(l,j)}\mathbf W_{\mathrm{down}}^{(l)}[j,:]$, where $\mathbf a^{(l,j)}$ denotes the corresponding hidden activation. Thus, both attention heads and MLP neurons appear as additive component-level updates in the same residual-stream space. Let $\mathcal C$ denote the candidate component set consisting of attention heads and MLP neurons; these form the basic units for the activation-space and parameter-space interventions introduced later. Detailed component interfaces and architecture-specific replaceable parameter slices are deferred to the supplementary material.

\subsection{Standard Activation Patching as a Reference Template}
\label{sec:prelim_intervention}

Activation patching provides a standard activation-space intervention template for testing whether a component’s clean-run activation can restore a target behavior or control signal in a corrupted run. Given a clean input $x_{\mathrm{clean}}$, a corrupted input $x_{\mathrm{corr}}$, and a component $c\in\mathcal C$, it replaces the corrupted activation of $c$ with its clean counterpart and measures the resulting restoration effect $\Delta \mathcal F_{\mathrm{patch}}(c)$ under a utility $\mathcal F$. Activation patching is therefore useful for identifying where behavior-relevant signals become causally effective during inference, including restoration, routing, and aggregation sites.
Here, however, we use the clean/corrupted formulation only as a canonical restoration template, not as the exact activation-side protocol in our main analyses. While $\mathcal F$ can often be defined directly in token space for narrowly defined prediction tasks, this is less suitable for capability-level generative behaviors such as instruction following: changing the instruction may alter task semantics, and token-level readouts are often too narrow to determine whether an instruction-conditioned control state has been formed and sustained. 
We therefore preserve the same restoration logic but instantiate it through the shared vector-anchor behavioral interface introduced later. In the paired-model setting studied here, the activation-side analysis takes a fixed-input cross-model form under this shared criterion; details are given in Sec.~\ref{sec:hierarchy_recovery} and the supplementary material.

\begin{figure*}[t]
    \centering
    \includegraphics[width=\textwidth]{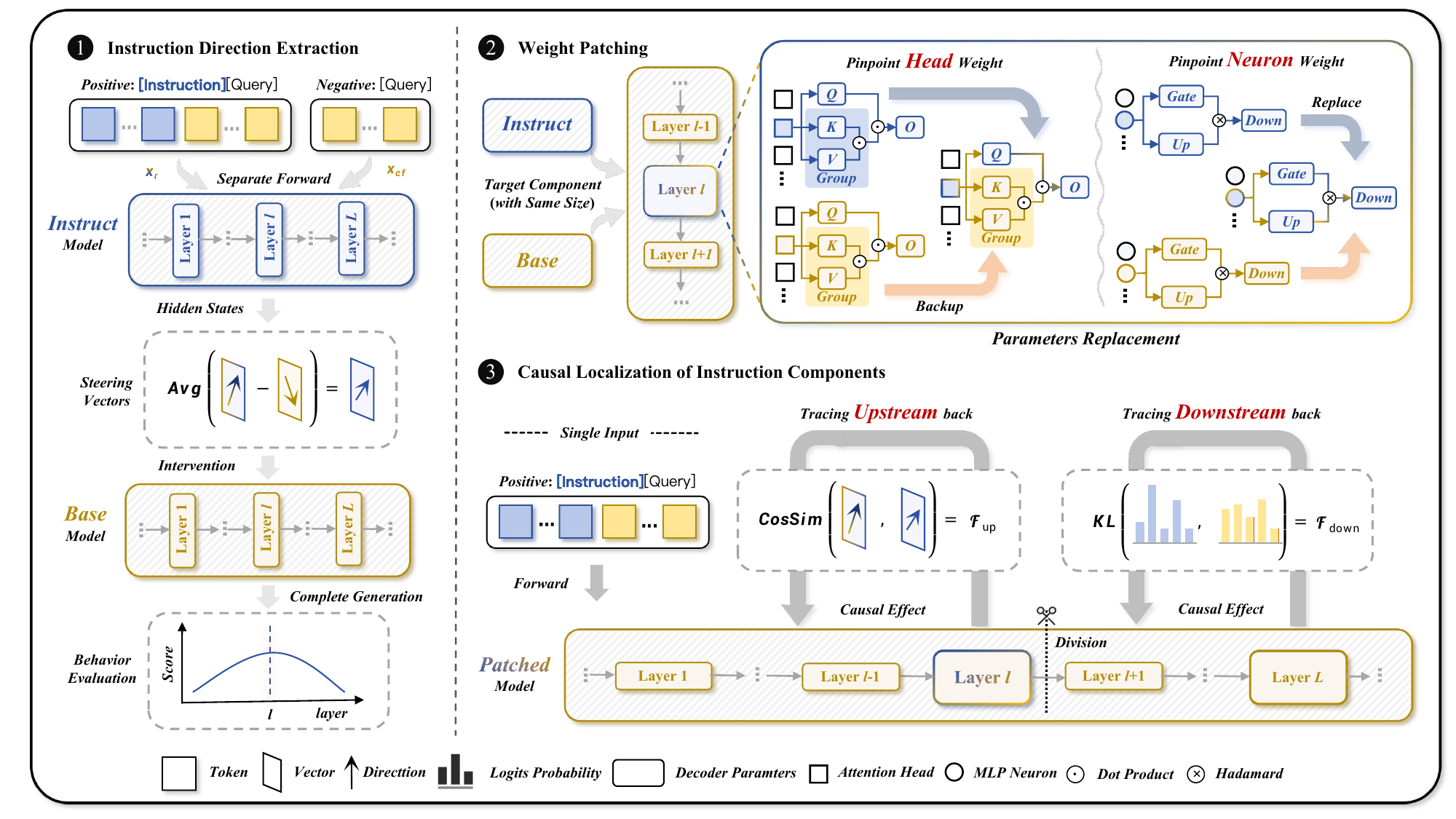}
    \vspace{-15pt}
    \caption{Overview of the proposed framework. We first extract an instruction-direction anchor as a shared behavioral interface, then apply Weight Patching for source-level localization, and finally combine parameter-space and activation-space evidence to 
    identify source-side carriers, aggregation/routing modules, and downstream execution circuits.
    }
    \label{fig:wp-pipeline}
\end{figure*}

\section{Method}
\label{sec:method}
Fig.~\ref{fig:wp-pipeline}  summarizes the framework. We first construct a vector-anchor behavioral interface as a shared internal criterion. We then use Weight Patching as the primary source-level test and a first-order approximation for scalable screening. Finally, we combine parameter-space and activation-space evidence to organize localized components into a source–aggregation–execution hierarchy.

\subsection{Paired-Model Setting}
\label{sec:paired_setting}

The primary setting is a paired-model setting consisting of a pretrained base model $M_{\mathrm{base}}$ with parameters $\theta_{\mathrm{base}}$ and a behavior-specialized counterpart $M_{\mathrm{sft}}$ with parameters $\theta_{\mathrm{sft}}$.
This setting is natural for source-level analysis because the capability difference of interest is realized through post-training parameter change under a shared architecture.
The two models share the same architecture and differ only in parameter values, with post-training parameter change $\Delta\theta=\theta_{\mathrm{sft}}-\theta_{\mathrm{base}}$. In the main empirical setting, the input takes the form $x=[I;C_{\mathrm{ctx}}]$, where $I$ denotes the natural-language instruction and $C_{\mathrm{ctx}}$ denotes the accompanying context. Throughout the method, the candidate component set $\mathcal C$ consists of attention heads and individual MLP neurons.
We do not assume that $M_{\mathrm{sft}}$ differs from $M_{\mathrm{base}}$ only in the studied capability; rather, under task-aligned inputs, the analysis asks which specialized parameterized components consistently help recover the capability-associated anchor state.
Throughout, `source-level' is interpreted relative to the paired-model difference and the intervention granularity used here.

\subsection{Vector-Anchor Behavioral Interface}
\label{sec:vector_anchor_interface}

Direct text-level evaluation is often unstable for mechanistic localization in generative tasks such as instruction following, where control must persist across extended generation, and narrow token-level readouts are often insufficient to determine whether the relevant control state has been formed. 
To make source-level localization operational in this setting, we construct a vector-anchor behavioral interface as a shared internal criterion,  using an existing task-vector extraction and steering procedure~\cite{task_vector_iclr2025}. 

Let $\mathcal D_{\mathrm{inst}}=\{(x_r^{(i)},x_{cf}^{(i)})\}_{i=1}^N$ denote a paired dataset, where $x_r^{(i)}$ is an instructed input and $x_{cf}^{(i)}$ is its instruction-removed counterpart with matched context. For each candidate layer $l\in\mathcal L$, an instruction direction is extracted by averaging residual-stream differences at a designated anchor position $t_a$:
\begin{equation}
\mathbf v^{(l)}
=
\frac{1}{N}
\sum_{i=1}^{N}
\left[
\mathbf z_l^{(t_a)}(x_r^{(i)}\mid M_{\mathrm{sft}})
-
\mathbf z_l^{(t_a)}(x_{cf}^{(i)}\mid M_{\mathrm{sft}})
\right].
\label{eq:layer_direction}
\end{equation}
The anchor layer is then selected as
\begin{equation}
l_a
=
\arg\max_{l\in\mathcal L}
\mathrm{Recover}\!\left(\mathbf v^{(l)}\right),
\label{eq:anchor_layer_selection}
\end{equation}
where $\mathrm{Recover}(\cdot)$ denotes the behavioral recovery score under steering. The final task vector is set to $\mathbf v=\mathbf v^{(l_a)}$. 

Based on this construction, the anchor utility is defined as
\begin{equation}
\mathcal F_a(M,x)
=
\operatorname{sim}\!\bigl(\mathbf z_a(x\mid M),\mathbf v\bigr).
\label{eq:anchor_utility}
\end{equation}
Here, $\mathbf z_a(x\mid M)\equiv \mathbf z_{l_a}^{(t_a)}(x\mid M)$ denotes the anchor representation.
The specialized-to-base anchor gap is
\begin{equation}
G(x)
=
\mathcal F_a(M_{\mathrm{sft}},x)-\mathcal F_a(M_{\mathrm{base}},x).
\label{eq:anchor_gap}
\end{equation}
This quantity serves as the normalization term for both exact and first-order restoration scores. The resulting interface provides the shared internal criterion under which parameter-space and activation-space evidence can be compared in the remainder of the method; full anchor construction, layer selection, and input-adaptive calibration are deferred to the supplementary material.

\subsection{Weight Patching for Identifying Candidate Source-Side Carriers}
\label{sec:weight_patching}


Weight Patching shifts the intervention target from activations to parameters. Under a fixed input, it tests whether transplanting a component’s specialized parameters into the base model recovers the target control representation at the anchor, thereby helping identify candidate source-side carriers and separate them from downstream aggregation/routing and execution sites under the paired-model setting.

For a component $c\in\mathcal C$, let $\Theta^{(c)}$ denote its replaceable parameter slice. The corresponding single-component parameter-patched model is defined as
\begin{equation}
M_{\mathrm{base}}^{(c\leftarrow \mathrm{sft})}
:=
\operatorname{Replace}(M_{\mathrm{base}},M_{\mathrm{sft}};\Theta^{(c)}),
\label{eq:component_patch_model}
\end{equation}
meaning that only the parameters in $\Theta^{(c)}$ are taken from $M_{\mathrm{sft}}$, while all remaining parameters are kept from $M_{\mathrm{base}}$. When evaluating multi-component restoration or ablation, the same replacement operator is applied to a set of components:
\begin{equation}
M_{\mathrm{base}}^{(\mathcal S\leftarrow \mathrm{sft})}
:=
\operatorname{Replace}\!\left(
M_{\mathrm{base}},M_{\mathrm{sft}};
\bigcup_{c\in\mathcal S}\Theta^{(c)}
\right).
\label{eq:set_patch_model}
\end{equation}
The basic intervention unit throughout source localization, however, remains a single component.

The parameter slice is defined at the same structural granularity as the mechanistic unit under study:
\begin{equation}
\begin{aligned}
\Theta^{(H^{(l,h)})}
&=
\{W_Q^{(l,h)},\, W_O^{(l,h)}\},\\
\Theta^{(N^{(l,j)})}
&=
\{W_{\mathrm{gate}}^{(l)}[:,j],\,
  W_{\mathrm{up}}^{(l)}[:,j],\,
  W_{\mathrm{down}}^{(l)}[j,:]\}.
\end{aligned}
\label{eq:component_slices}
\end{equation}
For attention heads, replacing only the query and output slices avoids ambiguity when key and value projections are shared across heads. For MLP neurons, the gate column, up-projection column, and down-projection row together preserve the full feature-detection, gating, and write-back interface of the neuron. Detailed architecture-specific implementations, including grouped-query attention, are deferred to the supplementary material.

The source-level effect of component $c$ is quantified by how much of the anchor gap is restored when only that component's parameters are replaced:
\begin{equation}
E_w(c)
=
\mathbb E_{x\sim\mathcal D_{\mathrm{inst}}}
\left[
\frac{
\mathcal F_a(M_{\mathrm{base}}^{(c\leftarrow \mathrm{sft})},x)
-
\mathcal F_a(M_{\mathrm{base}},x)
}{
G(x)
}
\right].
\label{eq:wp_effect}
\end{equation}

A large $E_w(c)$ 
provides evidence that the component’s specialized parameters make a substantial contribution to recovering the target internal control representation.
Such components are therefore supported as candidate source-side parameter carriers, rather than merely downstream sites where behavior-relevant information becomes visible.

\subsection{Efficient Screening via First-Order Weight Attribution}
\label{sec:efficient_screening}

Exact Weight Patching provides the most direct source-level causal test, but exhaustive evaluation over all heads and especially all MLP neurons is computationally prohibitive. We therefore introduce a first-order weight-attribution approximation not as a replacement for exact intervention, but as a scalable surrogate for full-model screening and neuron-scale localization. This design is conceptually analogous to attribution patching in activation space, which approximates exact patching by locally linearizing the restoration utility around a reference point and using a gradient--difference inner product as a fast screening score~\cite{attribution_patching,kramar2024atpstar}. Here, we apply the same approximation logic in parameter space, while retaining exact Weight Patching as the primary exact interventional test.

For a component $c$, exact Weight Patching replaces the base-model slice $\Theta_{\mathrm{base}}^{(c)}$ by $\Theta_{\mathrm{sft}}^{(c)}=\Theta_{\mathrm{base}}^{(c)}+\Delta\theta^{(c)}$, where $\Delta\theta=\theta_{\mathrm{sft}}-\theta_{\mathrm{base}}$. Under a first-order Taylor expansion of the anchor utility around $M_{\mathrm{base}}$, the resulting change in anchor utility is approximated by
\[
\mathcal F_a(M_{\mathrm{base}}^{(c\leftarrow \mathrm{sft})},x)-\mathcal F_a(M_{\mathrm{base}},x)
\approx
\sum_{p\in\Theta^{(c)}} \Delta\theta_p
\frac{\partial \mathcal F_a(M_{\mathrm{base}},x)}{\partial \theta_p}.
\]
Normalizing this local approximation by the same anchor gap $G(x)$ used in exact Weight Patching yields the first-order weight-attribution score:
\begin{equation}
\mathrm{Attr}_{\mathrm{wt}}(c)
=
\mathbb E_{x\sim\mathcal D_{\mathrm{inst}}}
\left[
\frac{
\sum_{p\in\Theta^{(c)}}
\Delta\theta_p
\frac{\partial \mathcal F_a(M_{\mathrm{base}},x)}{\partial \theta_p}
}{
G(x)
}
\right].
\label{eq:wap_score}
\end{equation}
Here, $\Theta^{(c)}$ denotes the parameter slice of component $c$. The score measures how well the post-training parameter change on $c$ aligns with the local sensitivity of the anchor utility, and thus serves as a first-order surrogate to the exact restoration effect rather than an independent heuristic.

In practice, exact Weight Patching is used whenever exhaustive intervention remains feasible, especially for head-level analyses and for validating top-ranked candidates. For neuron-scale localization, however, exhaustive exact patching is computationally prohibitive, so first-order weight attribution serves as the practical screening score that makes fine-grained localization operational at model scale. The fidelity of this approximation is evaluated separately in Sec.~\ref{sec:exp_approximation}.
 
\subsection{From Candidate Source Carriers to Aggregation/Routing and Execution Modules}
\label{sec:hierarchy_recovery}
  
Under the shared anchor utility, 
WP serves as the primary parameter-space sufficiency test for candidate source-side carriers, while AP provides complementary activation-space evidence for where the recovered state becomes effective during inference.
Here AP is instantiated in the fixed-input cross-model form introduced in Sec.~\ref{sec:prelim_intervention}: for the same input $x$, we replace the activation of component $c$ in the base-model run with its counterpart from $M_{\mathrm{sft}}$ and measure the extent to which the anchor-side state is recovered. Although this differs from the canonical clean/corrupted formulation, it preserves the same restoration logic under controlled component replacement. Relative to the anchor layer, components before the anchor are treated as potential upstream suppliers because they contribute to forming the recovered anchor-side control state, whereas components after the anchor are treated as downstream execution modules because they determine how that recovered state is translated into output behavior. 
For the downstream stage, we considered both vector-projection and KL-divergence readouts and found qualitatively similar patterns; in the main text, we report the KL-based results because execution-stage modules are most directly characterized by changes in the next-token distribution.
This yields a staged hierarchy rather than a single global ranking. Detailed activation-side intervention and hierarchy-recovery procedures, including the projection-based variant, are deferred to the supplementary material.

For a component $c\in\mathcal C$, let $\widetilde{M}^{(c)}_{\mathrm{base}}$ denote the base-model run in which the activation of $c$ is replaced by its counterpart from $M_{\mathrm{sft}}$. Its activation-side restoration effect is
\begin{equation}
E_a(c)=\mathbb{E}_{x\sim\mathcal D_{\mathrm{inst}}}\!\left[
\frac{
\mathcal F_a(\widetilde{M}^{(c)}_{\mathrm{base}},x)-\mathcal F_a(M_{\mathrm{base}},x)
}{
G(x)
}
\right].
\label{eq:ap_restore}
\end{equation}

\begin{figure}[!ht]
    \centering
    \includegraphics[width=0.95\linewidth]{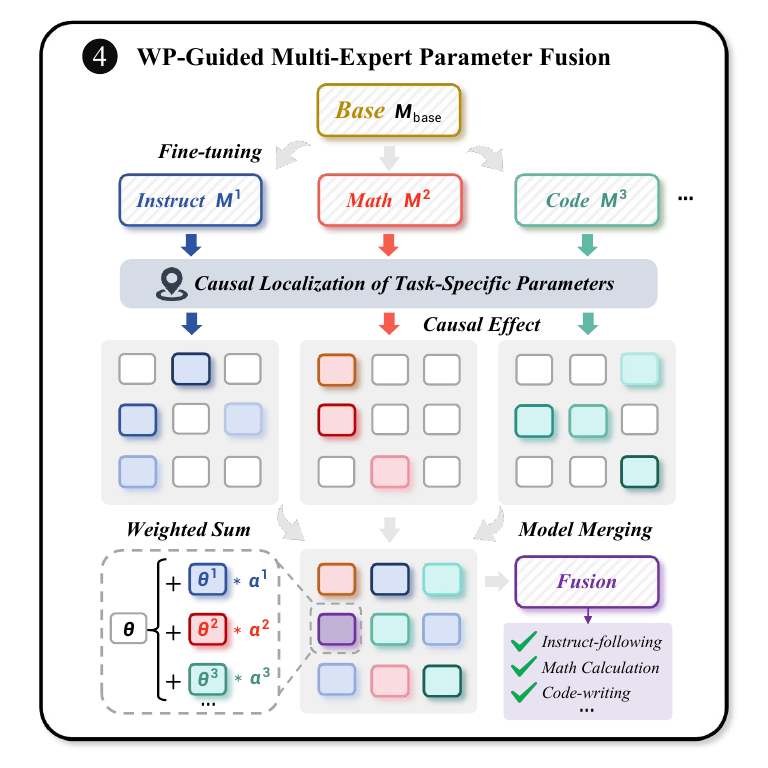}
    \vspace{-5pt}

    \caption{Reusing WP-recovered component scores for mechanism-aware model merging.}
    \label{fig:model_merging}
    \vspace{-5pt}
\end{figure}

Large $E_a(c)$ indicates that restoring the specialized activation of $c$ helps recover the anchor-side control representation under the shared criterion. This supports $c$ as 
an activation-side restoration site or aggregation/routing module, but does not by itself show that the behavior difference is written in c’s own parameters.

To trace candidate upstream suppliers for a chosen target component $c_{\mathrm{tar}}$, let
\begin{equation}
\mathbf g_{\mathrm{need}}(c_{\mathrm{tar}})
=
\nabla_{\Psi_{c_{\mathrm{tar}}}(\mathbf Z)}
\mathcal F_a(M_{\mathrm{base}},x)
\label{eq:need_direction}
\end{equation}
denote the need direction at the residual-stream input of $c_{\mathrm{tar}}$, where $\Psi_{c_{\mathrm{tar}}}(\mathbf Z)$ extracts that input. For a candidate supplier component $c_{\mathrm{sup}}$, define its weight-side support to $c_{\mathrm{tar}}$ by
\begin{equation}
s_{\mathrm{wt}}(c_{\mathrm{tar}},c_{\mathrm{sup}})
=
\sum_{p\in\Theta^{(c_{\mathrm{sup}})}}
\Delta\theta_p
\frac{
\partial \mathcal F^{(c_{\mathrm{tar}})}_{\mathrm{need}}(M_{\mathrm{base}},x)
}{
\partial \theta_p
},
\label{eq:weight_supply}
\end{equation}
where
\begin{equation}
\mathcal F^{(c_{\mathrm{tar}})}_{\mathrm{need}}(M,x)
=
\left\langle
\Psi_{c_{\mathrm{tar}}}(\mathbf Z(M,x)),
\mathbf g_{\mathrm{need}}(c_{\mathrm{tar}})
\right\rangle.
\label{eq:need_utility}
\end{equation}
Their functional link strength is then
\begin{equation}
E_{\mathrm{link}}(c_{\mathrm{tar}},c_{\mathrm{sup}})
=
\mathbb{E}_{x\sim\mathcal D_{\mathrm{inst}}}\!\left[
\frac{
s_{\mathrm{wt}}(c_{\mathrm{tar}},c_{\mathrm{sup}})
}{
G(x)
}
\right].
\label{eq:link_strength}
\end{equation}
A large $E_{\mathrm{link}}(c_{\mathrm{tar}},c_{\mathrm{sup}})$ indicates that the post-training change on $c_{\mathrm{sup}}$ is well aligned with increasing the target-side need utility. In practice, target components for tracing may be nominated by activation-side evidence, parameter-side evidence, or prior mechanistic hypotheses. Retrieved suppliers are retained as source candidates only when they also exhibit strong exact WP effects $E_w(c)$, so that supplier tracing complements rather than replaces the primary parameter-side source criterion.

\subsection{Reusing WP-Recovered Component Scores for Mechanism-Aware Model Merging}
\label{sec:wp_guided_fusion}

WP-recovered component scores can be reused for component-wise model merging by weighting experts at the component level rather than averaging expert deltas uniformly.
Let $M_{\mathrm{base}}$ denote a shared pretrained base model, and let $\{M^{(k)}\}_{k=1}^K$ denote expert models obtained by finetuning the same base toward different capabilities. Their expert-specific parameter deltas are $\Delta\theta^{(k)}=\theta^{(k)}-\theta_{\mathrm{base}}$. 
Rather than averaging these deltas uniformly over the full parameter space, we propose a precise component-wise fusion over the same head- and neuron-level units used in Weight Patching.

For each expert model $M^{(k)}$, a component score $s^{(k)}(c)$ is computed for every component $c\in\mathcal C$, using either the exact Weight Patching effect or its first-order approximation. Only positive evidence is retained, $\bar s^{(k)}(c)=\max(s^{(k)}(c),0)$, and the component-wise fusion weight of expert $k$ on component $c$ is defined as
\begin{equation}
\alpha^{(k)}(c)
=
\begin{cases}
\dfrac{\bar s^{(k)}(c)}{\sum_{m=1}^{K}\bar s^{(m)}(c)}, & \sum_{m=1}^{K}\bar s^{(m)}(c)>0, \\[8pt]
\dfrac{1}{K}, & \sum_{m=1}^{K}\bar s^{(m)}(c)=0.
\end{cases}
\label{eq:fusion_weight}
\end{equation}

The fused model $M_{\mathrm{fuse}}$ is then defined component-wise. For each component $c$, its fused parameter slice is
\begin{equation}
\theta_{\mathrm{fuse}}^{(c)}
=
\theta_{\mathrm{base}}^{(c)}
+
\sum_{k=1}^{K}\alpha^{(k)}(c)\,\Delta\theta^{(k,c)},
\label{eq:fused_slice}
\end{equation}
where $\Delta\theta^{(k,c)}=\theta^{(k,c)}-\theta_{\mathrm{base}}^{(c)}$. 
This yields a mechanism-aware model-merging rule implemented through component-wise fusion, in which different experts dominate different components based on source-level support rather than a globally uniform parameter average.
Detailed slice-level fusion and implementation are provided in the supplementary material.

\section{Experiments}
\label{sec:experiments}
\subsection{Experimental Setup}
\label{sec:exp_setup}

We evaluate the proposed framework on instruction-following behaviors that require persistent control over generation. For mechanistic analysis, we use IFEval~\cite{IFEval_zhou2023instruction}, whose constraints are explicit and automatically checkable. We study six representative tasks---\emph{No Comma}, \emph{Title}, \emph{Multiple Sections}, \emph{Quotation}, \emph{Number Highlighted Sections}, and \emph{English Capital}---using \emph{English Capital} as the main running example.
This choice is driven by interface reliability: using the existing task-vector extraction and steering procedure of~\cite{task_vector_iclr2025}, we obtain sufficiently recoverable anchor directions on these six tasks, whereas the remaining IFEval tasks yield directions that are too weak for reliable anchor-based causal analysis.
Our interpretability experiments use three paired base/instruction-tuned Llama settings: Llama-3.2-3B, Llama-3.1-8B~\cite{llama3}, and Llama-2-13B~\cite{touvron2023llama2}. The main paper focuses on Llama-3.2-3B, with selected cross-scale results on 8B and 13B. To broaden architectural coverage beyond the Llama family, the supplementary material additionally reports selected results on paired Qwen2.5, Mistral-7B-v0.3, and Gemma 2 settings~\cite{qwen25tech,mistral7bv03,gemma2tech}. 
For IFEval data usage and steering-vector extraction, we follow~\cite{task_vector_iclr2025}. We evaluate upstream localization by anchor similarity, downstream execution by KL divergence to the instruction-tuned model, and end-to-end behavior by IFEval strict accuracy; a vector-projection variant gives similar patterns and is deferred to the supplementary material. Unless otherwise noted, exact Weight Patching is used when exhaustive intervention is tractable, while the first-order score is used mainly to screen neuron-scale candidates before exact validation. Additional implementation details are provided in the supplementary material.

For mechanism-aware model merging, we follow the Llama-2-13B expert-pool setup used in DARE, WIDEN, and AIM~\cite{Yu2023LanguageMA,yu2024widen,nobari2025aim}. We use WizardLM-13B, WizardMath-13B, and Llama-2-13B-Code-Alpaca as the instruction, math, and code experts~\cite{wizardlm,wizardmath,codealpaca}, and study four combinations: \textit{Code + Instruction}, \textit{Code + Math}, \textit{Instruction + Math}, and \textit{Code + Instruction + Math}. We evaluate merged models on HumanEval~\cite{chen2021codex}, MBPP~\cite{austin2021program_synthesis}, MMLU~\cite{mmlu}, MATH~\cite{hendrycks2021math}, GSM8K~\cite{cobbe2021gsm8k}, and IFEval~\cite{IFEval_zhou2023instruction}, and compare against Avg Baseline~\cite{wortsman2022model_soups}, Task Arithmetic~\cite{task_arithmetic_ilharco2023}, TIES-Merging~\cite{yadav2023ties}, DARE Task Arithmetic, DARE TIES~\cite{Yu2023LanguageMA}, and WIDEN~\cite{yu2024widen}. Except for WIDEN, the baselines are implemented with MergeKit~\cite{goddard2024mergekit}. Additional implementation details are provided in the supplementary material. 

\begin{figure}[!ht]
  \centering
  \includegraphics[width=0.95\linewidth]{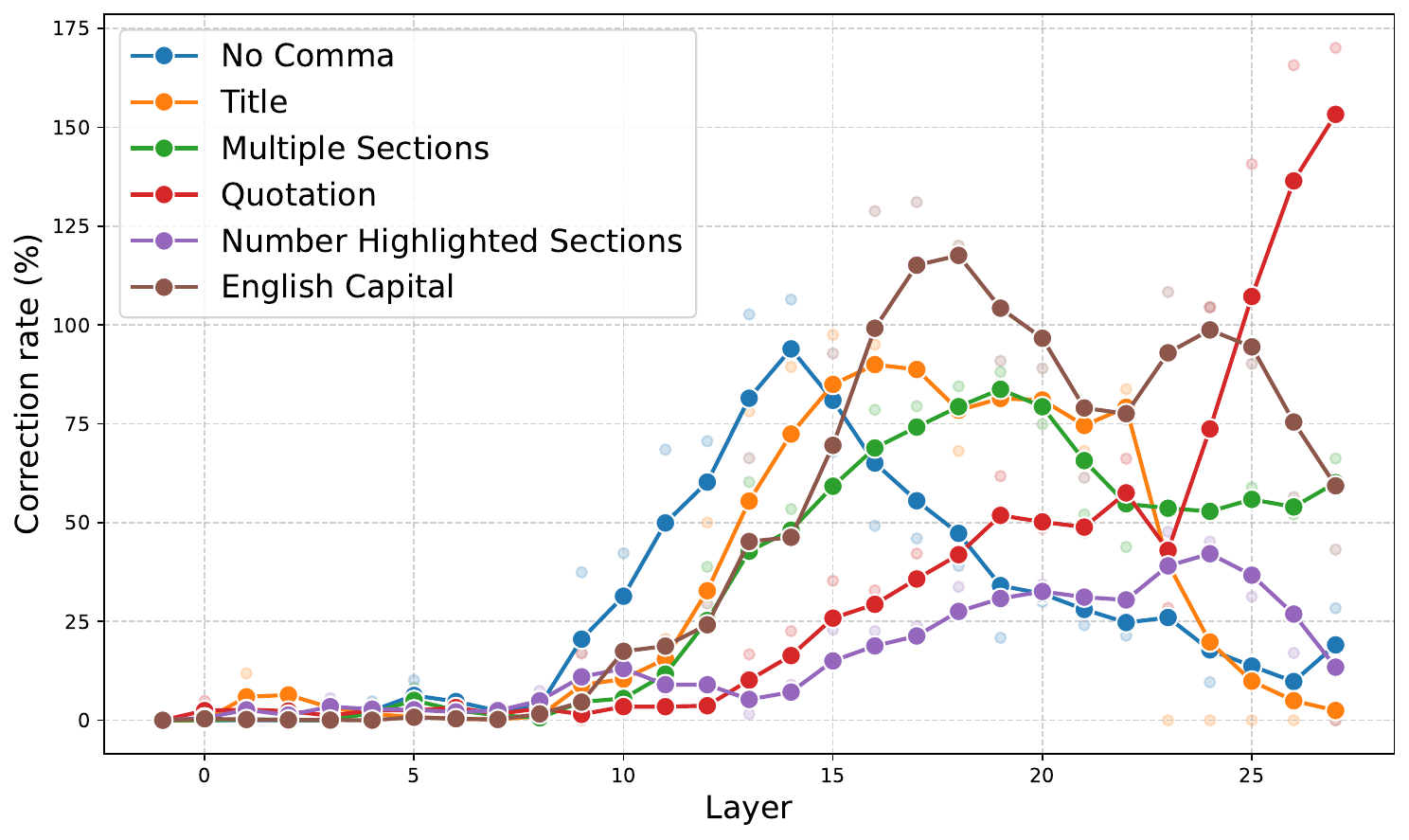}
  \vspace{-10pt}
  \caption{Layer-wise recovery under task-vector injection. Injecting the extracted task vector into different layers of Llama-3.2-3B-Base yields clear recovery peaks across six IFEval tasks, indicating that the anchor-level control representation emerges primarily in middle-to-late layers.}
  \label{fig:steering_plot}
\end{figure}

\subsection{Validating the Vector-Anchor Interface}
\label{sec:exp_anchor}

Before comparing AP and WP under a shared internal criterion, we first verify that the extracted task vector serves as a usable behavioral interface for instruction following rather than a fragile text-level proxy. To this end, we inject each task vector into different layers of Llama-3.2-3B-Base and measure the \emph{correction rate}, defined as the normalized fraction of the original base-to-instruct performance gap recovered by steering,
$R=(\mathrm{Acc}_{\mathrm{steer}}-\mathrm{Acc}_{\mathrm{base}})/(\mathrm{Acc}_{\mathrm{inst}}-\mathrm{Acc}_{\mathrm{base}})$.
As shown in Fig.~\ref{fig:steering_plot}, recovery peaks consistently emerge in mid-to-late layers across all six IFEval tasks rather than in shallow lexical layers, and the selected anchor layers lie in the same range. 
A vocabulary-space projection of the best-layer vectors offers only limited interpretability: structurally explicit tasks show partially recognizable format-related cues, whereas other tasks remain diffuse and abstract. Together, these results suggest that the extracted vector is not merely a prompt-level lexical residue, but a relatively stable intermediate control representation.
These checks are intended to validate the interface instantiation used in this paper, rather than to claim a new task-vector extraction method.

\begin{figure}[!ht]
    \centering

    \includegraphics[width=0.95\linewidth]{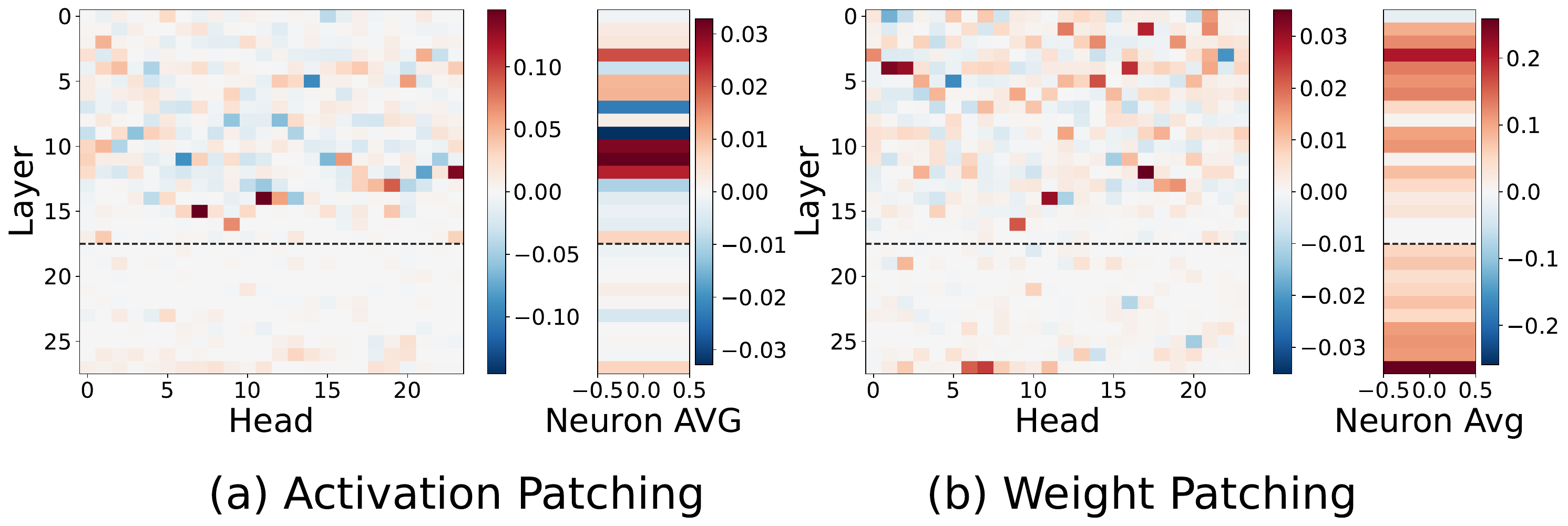}
    \vspace{-10pt}

    \caption{Component-importance heatmaps under Activation Patching and Weight Patching on the English Capital task.}

    \label{fig:heatmap}
    \vspace{-10pt}
    
\end{figure}

\begin{figure}[!th]
    \centering

    \includegraphics[width=0.95\linewidth]{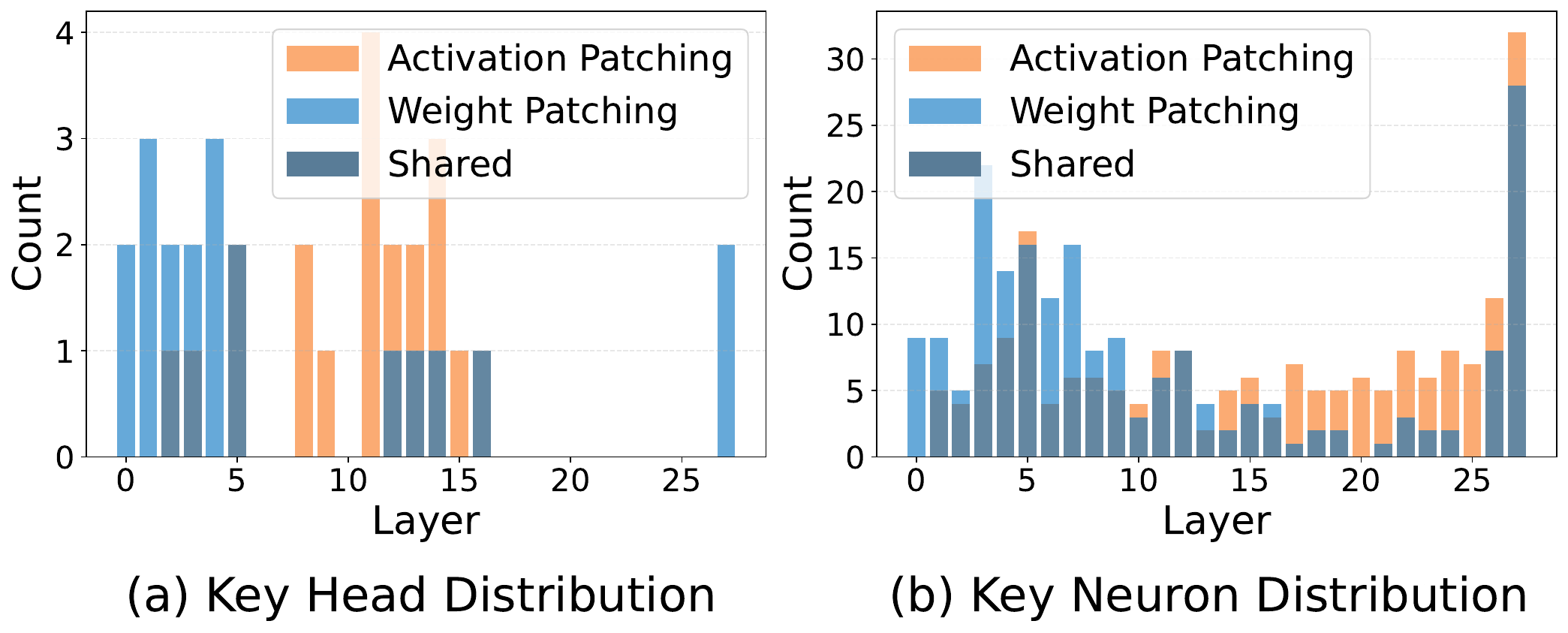}
    \vspace{-10pt}
    \caption{Layer distributions and overlaps of top-ranked heads and neurons identified by Activation Patching and Weight Patching on the English Capital task. }

    \label{fig:distribution}
    \vspace{-10pt}
    
\end{figure}

\subsection{Weight Patching vs. Activation Patching: Source Carriers versus aggregation/Routing Modules}
\label{sec:exp_source_convergence}

With the vector anchor validated as a shared criterion, we compare Activation Patching (AP) and Weight Patching (WP) under the same task and restoration objective.
Although both interventions are evaluated by anchor recovery, they probe different causal objects. AP is more sensitive to modules that become important once instruction-conditioned information is already in motion, whereas WP more directly tests whether a module’s specialized parameters are sufficient to reintroduce the target control signal.
Accordingly, limited overlap between AP- and WP-localized components should not be read as inconsistency; rather, it is expected if source-level carriers and downstream aggregation or routing modules are functionally distinct.

\begin{table}[!th]
    \centering
    \vspace{-5pt}
    \caption{Overlap between neurons localized directly by Weight Patching and neurons recovered by tracing upstream from the Top-20 AP-ranked heads. We report the maximum overlap (Max), the mean overlap over the Top-5 heads (Top-5 Avg), and the mean overlap over all 20 heads (Top-20 Avg).}
    \vspace{-8pt}
    \label{tab:neuron_intersection}
    
    \setlength{\aboverulesep}{0pt}
    \setlength{\belowrulesep}{0pt}
    \renewcommand{\arraystretch}{1.15} 
    \setlength{\tabcolsep}{4.5pt} 
    \begin{tabular}{lccc} 
        \whline
        \rowcolor{mygray}
        \textbf{Task} & \textbf{Max} & \textbf{Top-5 Avg} & \textbf{Top-20 Avg} \\
        \hline

        \multicolumn{4}{c}{\cellcolor{mygray!40}\textbf{Llama-3.2-3B}} \\
        \hline
        English Capital             & $0.775$ & $0.734$ & $0.507$ \\
        Multiple Sections           & $0.405$ & $0.371$ & $0.264$ \\
        Number Highlighted Sections & $0.485$ & $0.442$ & $0.305$ \\
        Title                       & $0.435$ & $0.422$ & $0.321$ \\
        No Comma                    & $0.445$ & $0.363$ & $0.246$ \\
        Quotation                   & $0.705$ & $0.628$ & $0.476$ \\
        \hline
        \textit{Avg}                & $0.542$ & $0.493$ & $0.353$ \\
        \hline
        
        \multicolumn{4}{c}{\cellcolor{mygray!40}\textbf{Llama-3.1-8B}} \\
        \hline
        English Capital             & $0.760$ & $0.654$ & $0.422$ \\
        Multiple Sections           & $0.620$ & $0.519$ & $0.348$ \\
        Number Highlighted Sections & $0.380$ & $0.298$ & $0.205$ \\
        Title                       & $0.495$ & $0.458$ & $0.313$ \\
        No Comma                    & $0.600$ & $0.551$ & $0.387$ \\
        Quotation                   & $0.575$ & $0.492$ & $0.375$ \\
        \hline
        \textit{Avg}                & $0.572$ & $0.495$ & $0.342$ \\
        \hline

        \multicolumn{4}{c}{\cellcolor{mygray!40}\textbf{Llama-2-13B}} \\
        \hline
        English Capital             & $0.700$ & $0.642$ & $0.411$ \\
        Multiple Sections           & $0.450$ & $0.427$ & $0.342$ \\
        Number Highlighted Sections & $0.400$ & $0.390$ & $0.287$ \\
        Title                       & $0.575$ & $0.498$ & $0.393$ \\
        No Comma                    & $0.530$ & $0.507$ & $0.355$ \\
        Quotation                   & $0.605$ & $0.566$ & $0.405$ \\
        \hline
        \textit{Avg}                & $0.543$ & $0.505$ & $0.365$ \\
  
        \whline
    \end{tabular}
    \vspace{-5pt}
\end{table}

The localization patterns differ systematically. On the \emph{English Capital} task, Fig.~\ref{fig:heatmap} shows that AP concentrates on middle-layer heads and nearby neurons, consistent with modules involved in routing or consolidating instruction-conditioned information during inference, whereas WP shifts more strongly toward shallow neurons, suggesting earlier and more neuron-centric source-level carriers. Fig.~\ref{fig:distribution} shows that this difference is global rather than local: the layer distributions are misaligned, and the overlap between AP- and WP-ranked components remains limited. Thus, the two interventions do not simply provide two views of the same important modules, but instead capture different functional stages of the mechanism.

These results are consistent with a source--aggregation separation in the present setting: activation patching more strongly surfaces aggregation and routing bottlenecks, 
whereas weight patching more directly highlights candidate source-side parameter carriers.
At this stage, however, the evidence is still distributional. The next subsection, therefore, tests whether this separation corresponds to a genuine causal division of labor through targeted ablation, restoration, and upstream linkage analyses.

\begin{figure*}[!th]
  \centering

  \includegraphics[width=0.95\linewidth]{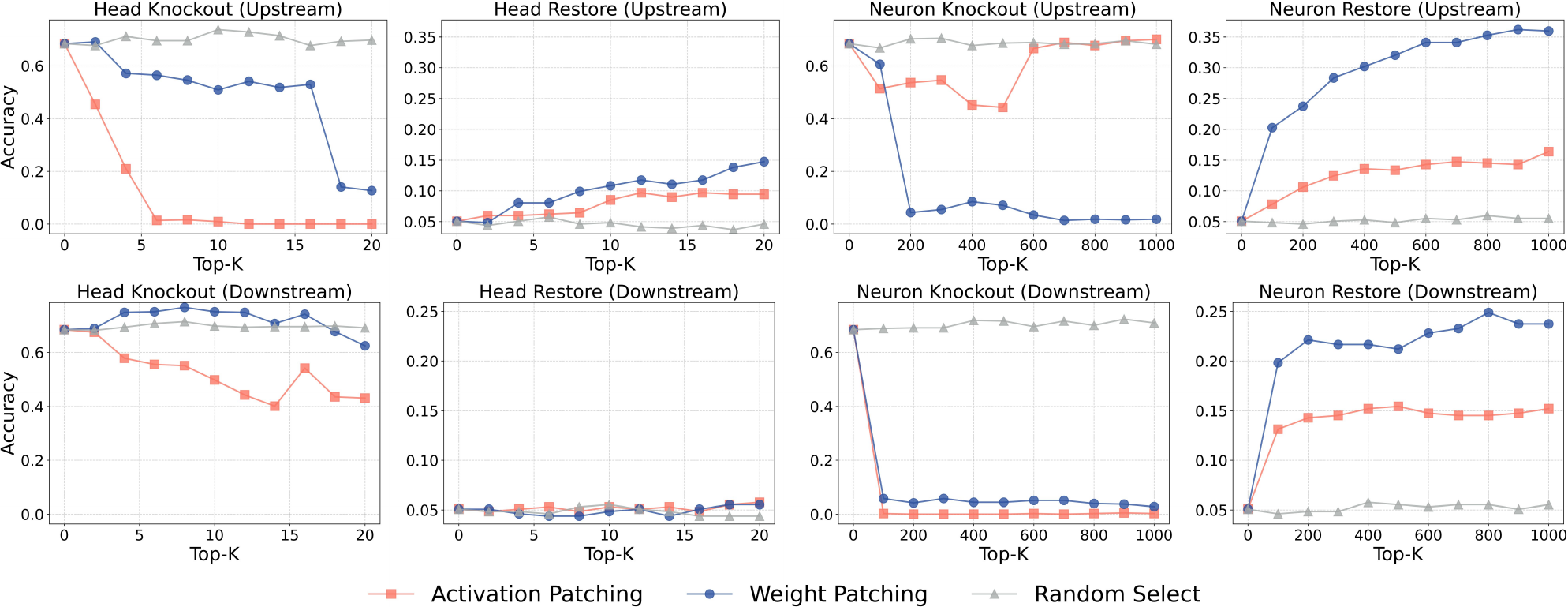}
  \vspace{-10pt}
  \caption{Causal validation of upstream (top row) and downstream (bottom row) modules on the English Capital task. The results support a source–convergence separation in upstream modules and a neuron-dominant pattern in downstream execution.}
  \label{fig:validation-upstream}
\end{figure*}

\begin{figure}[!th]

    \centering
      \includegraphics[width=0.95\linewidth]{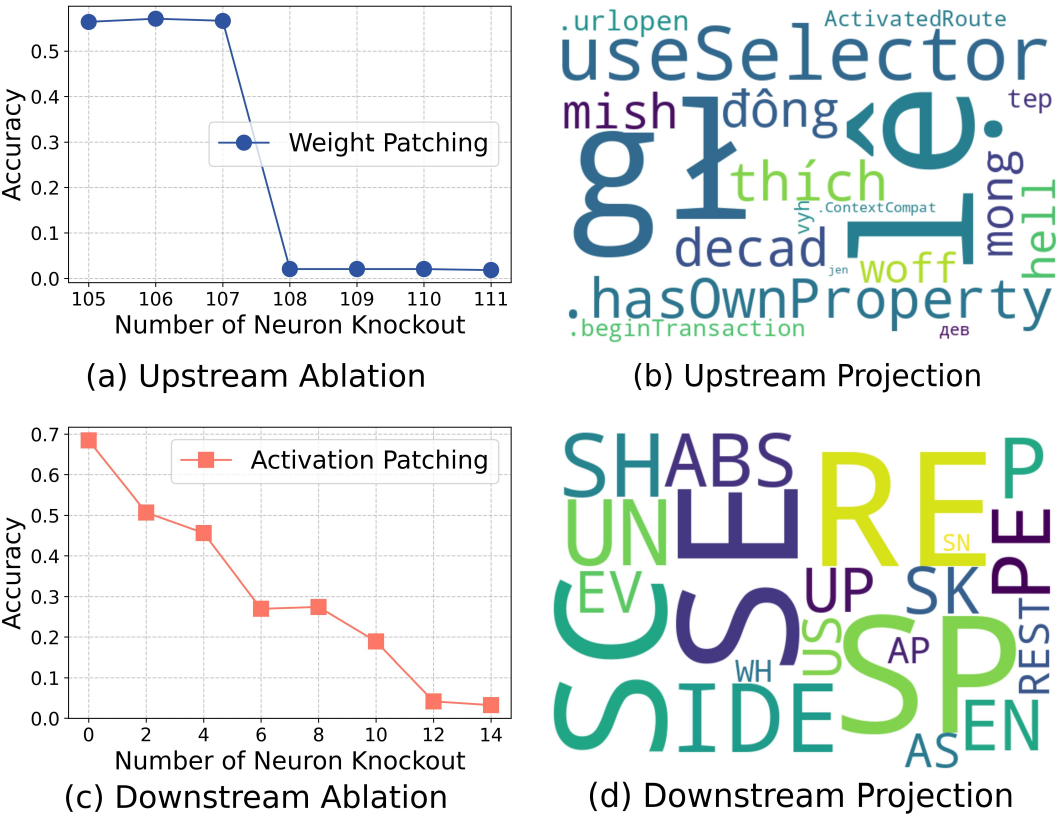}
      \vspace{-10pt}
    \caption{
    Fine-grained comparison of upstream and downstream neurons on the English Capital task. Left panels show ablation behavior, and right panels show output-space projections. Upstream neurons exhibit threshold-like collective effects and more abstract projections, whereas downstream neurons behave more additively and show more directly interpretable output-space features.
    }
    \label{fig:study}
    \vspace{-5pt}
    
\end{figure}

\begin{figure}[!ht]
  \centering
  \includegraphics[width=0.92\linewidth]{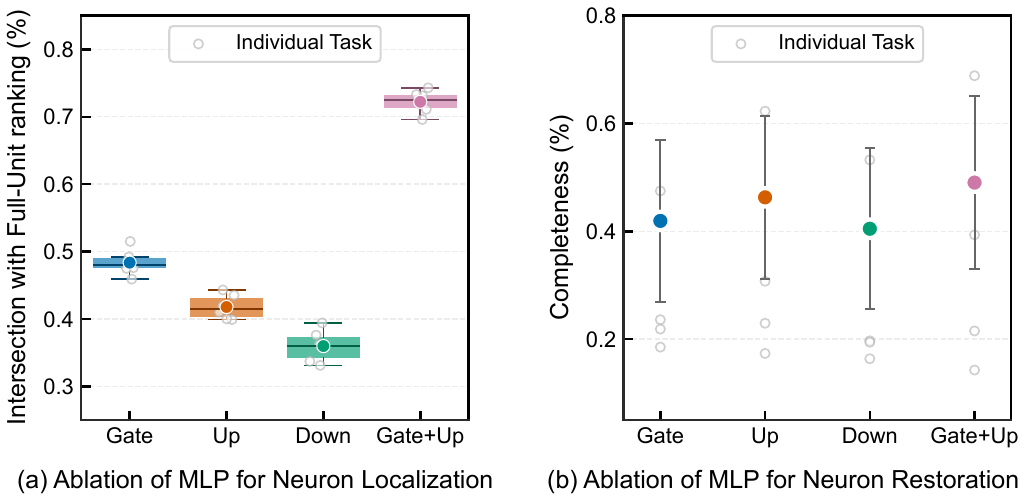}
  \vspace{-10pt}

  \caption{MLP submodule ablation for neuron localization. (a) Overlap between neurons localized from gate, up, down, or gate+up and the full-unit neuron ranking under WP. (b) Restoration completeness after globally restoring neurons selected under each setting, averaged over six IFEval tasks. }

  \label{fig:mlp-ablation}
\end{figure}

\subsection{Causal Validation of the Recovered Hierarchy}
\label{sec:exp_validation}

The source--aggregation separation in Sec.~\ref{sec:exp_source_convergence} is still, by itself, a distributional observation.
We therefore ask whether the modules highlighted by AP and WP also differ in their functional roles under intervention.
Fig.~\ref{fig:validation-upstream} provides this test for upstream modules on the \emph{English Capital} task. The contrast between heads and neurons is especially clear. Ablating a small number of AP-ranked upstream heads sharply degrades performance, yet restoring those heads alone yields only limited recovery.  
This pattern is consistent with aggregation or routing bottlenecks: these heads are important once the instruction-conditioned signal is already present, but are not by themselves sufficient to recreate it in the base model.
In contrast, WP-ranked upstream neurons show both stronger degradation under ablation and substantially stronger recovery under restoration. 
This supports interpreting the shallow neuron subsets identified by WP as candidate source-side carriers of the control representation, rather than merely downstream sites where the signal becomes visible.

A related but distinct pattern appears downstream. As shown in Fig.~\ref{fig:validation-upstream}, downstream heads remain only weakly restorative, even when some of them are behaviorally important under ablation, suggesting a more conditional routing role. Downstream neurons, however, exhibit both strong necessity and substantial sufficiency: ablating them rapidly destroys task performance, whereas restoring them recovers a much larger fraction of the target behavior.  
This result suggests that, once an abstract control representation has been formed and routed forward, 
later neurons appear to play a major role in translating the recovered control state into concrete output behavior.
In other words, the downstream evidence extends the hierarchy beyond the upstream source--aggregation distinction and points to a later execution stage centered on neurons.

Table~\ref{tab:neuron_intersection} further connects these two views by tracing the upstream suppliers of the top AP-ranked heads. Across tasks, the neurons recovered by tracing from the top-20 AP heads show substantial overlap with the neurons directly identified by WP, under both signed and absolute-value variants. The overlap is stable not only at the best-matching head ({Max}), but also in the {Top-5 Avg} and {Top-20 Avg}, indicating that the traced neurons are not random upstream artifacts. Instead, they align systematically with the WP-localized source neurons. This provides structural support for the interpretation that WP-recovered neurons are not merely another set of important modules; they are plausible upstream suppliers of the aggregation heads identified by AP. 
Taken together, these results are consistent with a hierarchical division of labor, with shallow neurons emerging as candidate source-side carriers, middle-layer heads as aggregation/routing interfaces, and later neurons as downstream execution units.
The next subsection examines whether this hierarchy also remains visible in finer-grained component signatures.

\subsection{Fine-Grained Component Analysis}
\label{sec:exp_finegrained}

The hierarchy is most convincing if it remains visible not only in aggregate intervention scores, but also in the fine-grained behavior of individual components. We therefore examine whether the recovered components exhibit distinct signatures consistent with the source--aggregation--execution organization, mainly through the \emph{English Capital} case, with Fig.~\ref{fig:mlp-ablation} providing a task-averaged MLP ablation across the six IFEval tasks. Fig.~\ref{fig:study} shows a clear contrast between upstream and downstream neurons: upstream neurons exhibit threshold-like collective effects, where ablating a relatively small set of WP-ranked neurons can trigger an abrupt collapse even though no single neuron alone accounts for the full behavior, whereas downstream neurons behave more additively and align more closely with output-space behavior. This supports a distinction between control formation and control execution. Fig.~\ref{fig:mlp-ablation} further sharpens this picture within SwiGLU: gate-only and up-only localization each recover a substantial portion of the full-unit neuron ranking, while down-only localization is markedly weaker; restoration is strongest when gate and up are combined, with up alone outperforming gate alone. This indicates that neuron localization depends primarily on the activation-forming branches of the MLP rather than on the write-back branch alone.

\begin{figure}[t]
  \centering
  \includegraphics[width=0.95\linewidth]{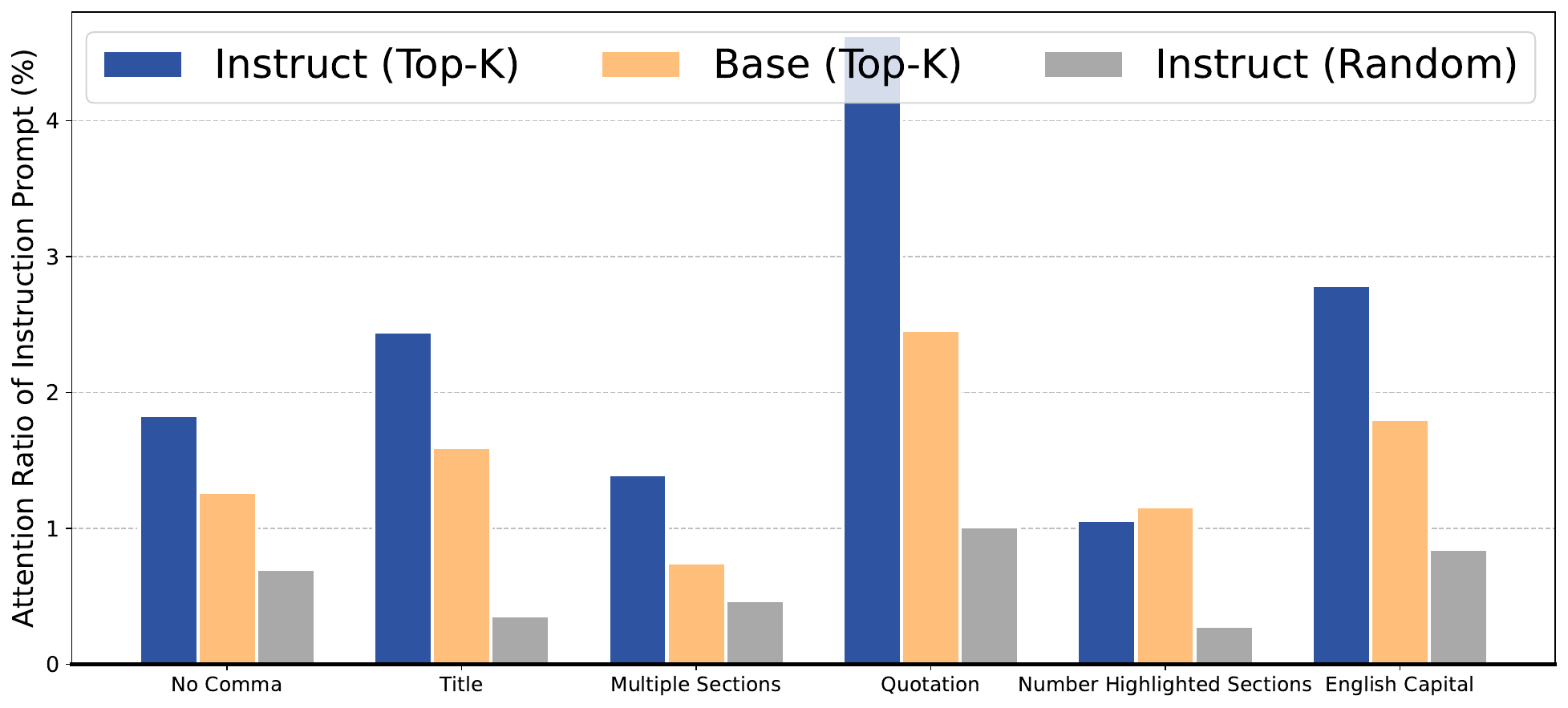}
  \vspace{-10pt}
  
  \caption{Selective attention of critical heads to instruction tokens. Critical heads in the instruction-tuned model allocate substantially more attention to instruction regions than matched heads in the base model or random heads, supporting their routing role.}
  \label{fig:attn-instruct}
\end{figure}

The head-level evidence is complementary. In Fig.~\ref{fig:attn-instruct}, the critical heads identified in the instruction-tuned model allocate substantially more attention to instruction-bearing regions than matched heads in the base model or random heads in the same model, indicating a routing role rather than a storage role. Thus, the fine-grained evidence refines rather than replaces the hierarchy recovered in Sec.~\ref{sec:exp_validation}: 
Sparse neurons are more closely tied to upstream control formation and downstream execution, whereas critical heads are more closely tied to selective reading, routing, and aggregation. Having established these fine-grained signatures---primarily in the \emph{English Capital} case, with complementary task-averaged evidence from Fig.~\ref{fig:mlp-ablation}---we next ask whether the recovered organization generalizes across instruction-following tasks.

\subsection{Cross-Task Generalization and Shared-vs.-Specific Structure}
\label{sec:exp_generalization}

\begin{figure}[!t]
  \centering
  \includegraphics[width=0.95\linewidth]{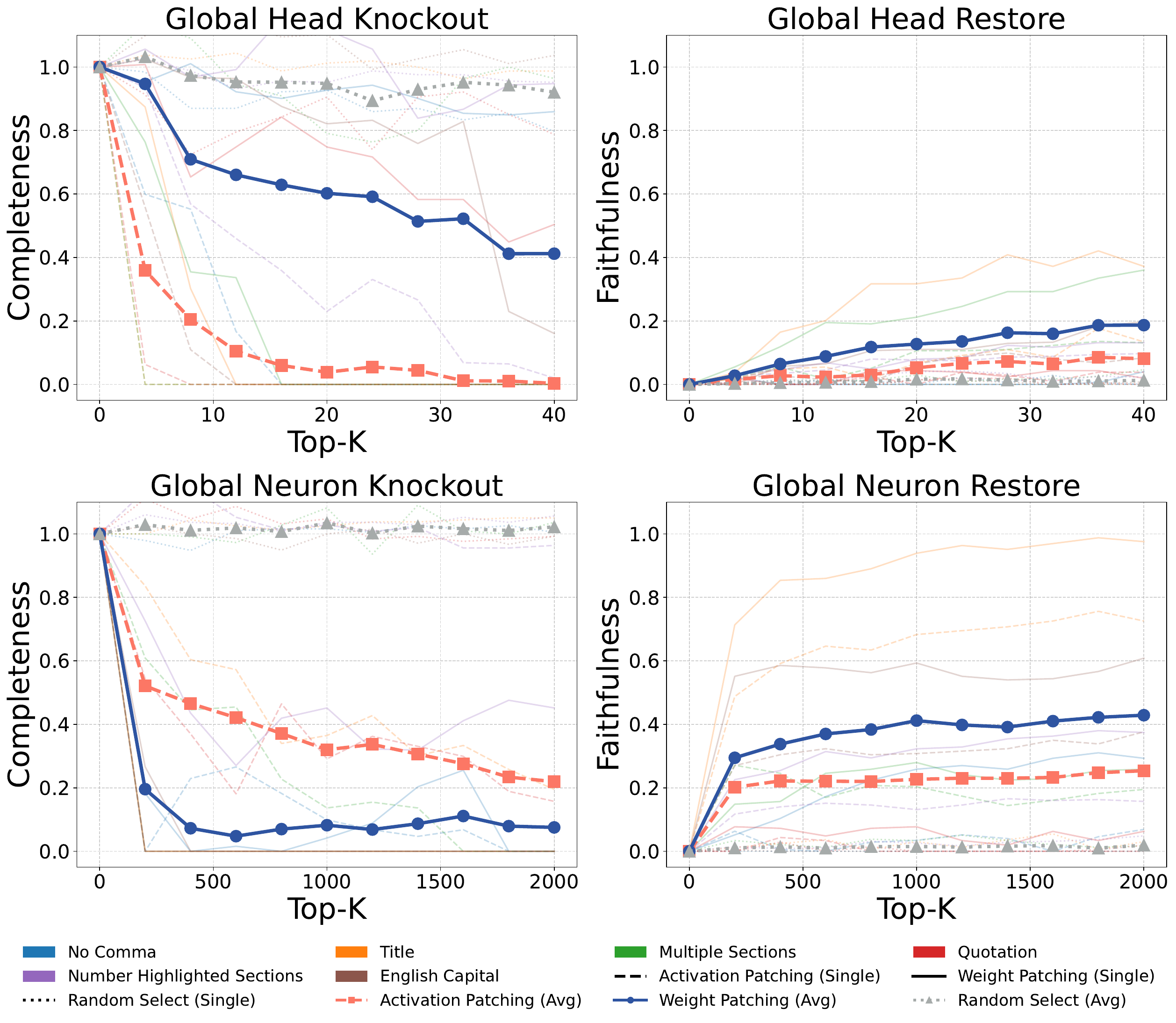}
  
  \vspace{-10pt}
  \caption{Cross-task ablation and restoration across six IFEval tasks. The plots compare head-level and neuron-level knockout/restoration under Activation Patching, Weight Patching, and random selection.} 
  \label{fig:task-avg}
\end{figure}

The analyses above establish a source--aggregation--execution hierarchy on the \emph{English Capital} task. We next ask whether this organization reflects a broader pattern of instruction following rather than a property of a single case.  
Fig.~\ref{fig:task-avg} shows that, although absolute intervention effects vary across tasks, the overall division of labor remains stable: 
Attention heads are more strongly associated with aggregation/routing effects, whereas neurons are more strongly associated with upstream source-side storage and downstream execution.
Thus, what generalizes across these tasks is not a single intervention profile, but a recurring hierarchical organization of how instruction-conditioned control is formed, routed, and enacted.

\begin{figure}[th]
  \centering
  \includegraphics[width=\linewidth]{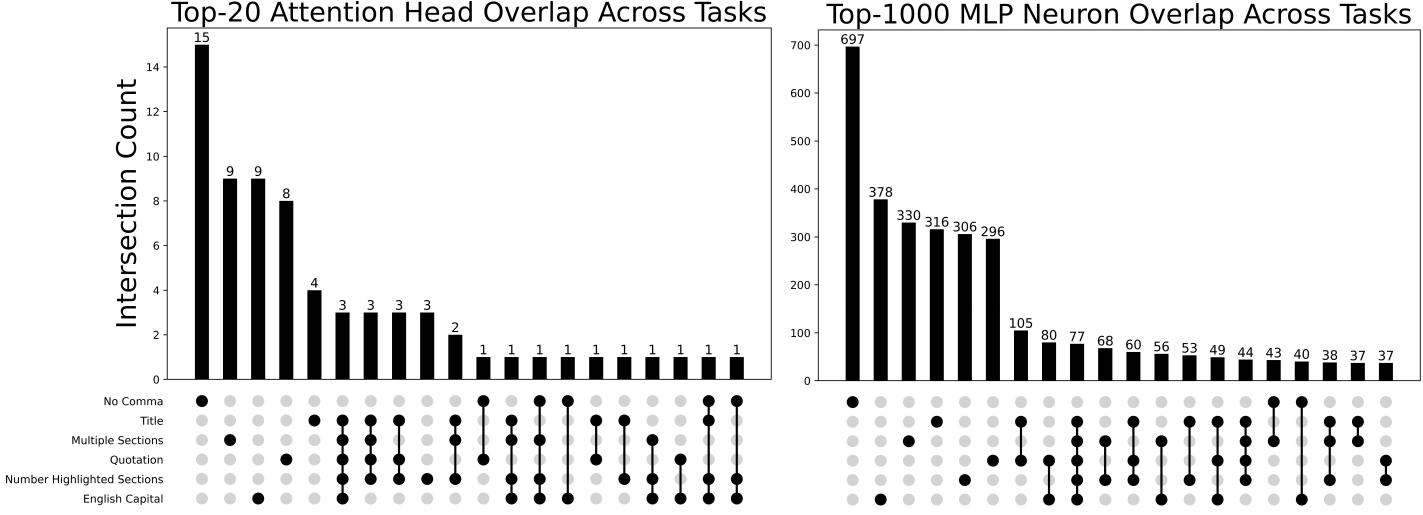}
  \vspace{-15pt}
  \caption{Overlap of critical modules across tasks. Intersection counts are shown for top-ranked heads and neurons across six IFEval tasks.}  
  \label{fig:task-overlap}
   \vspace{-5pt}
  
\end{figure}

\begin{figure}[!th]
  \centering
  \includegraphics[width=\linewidth]{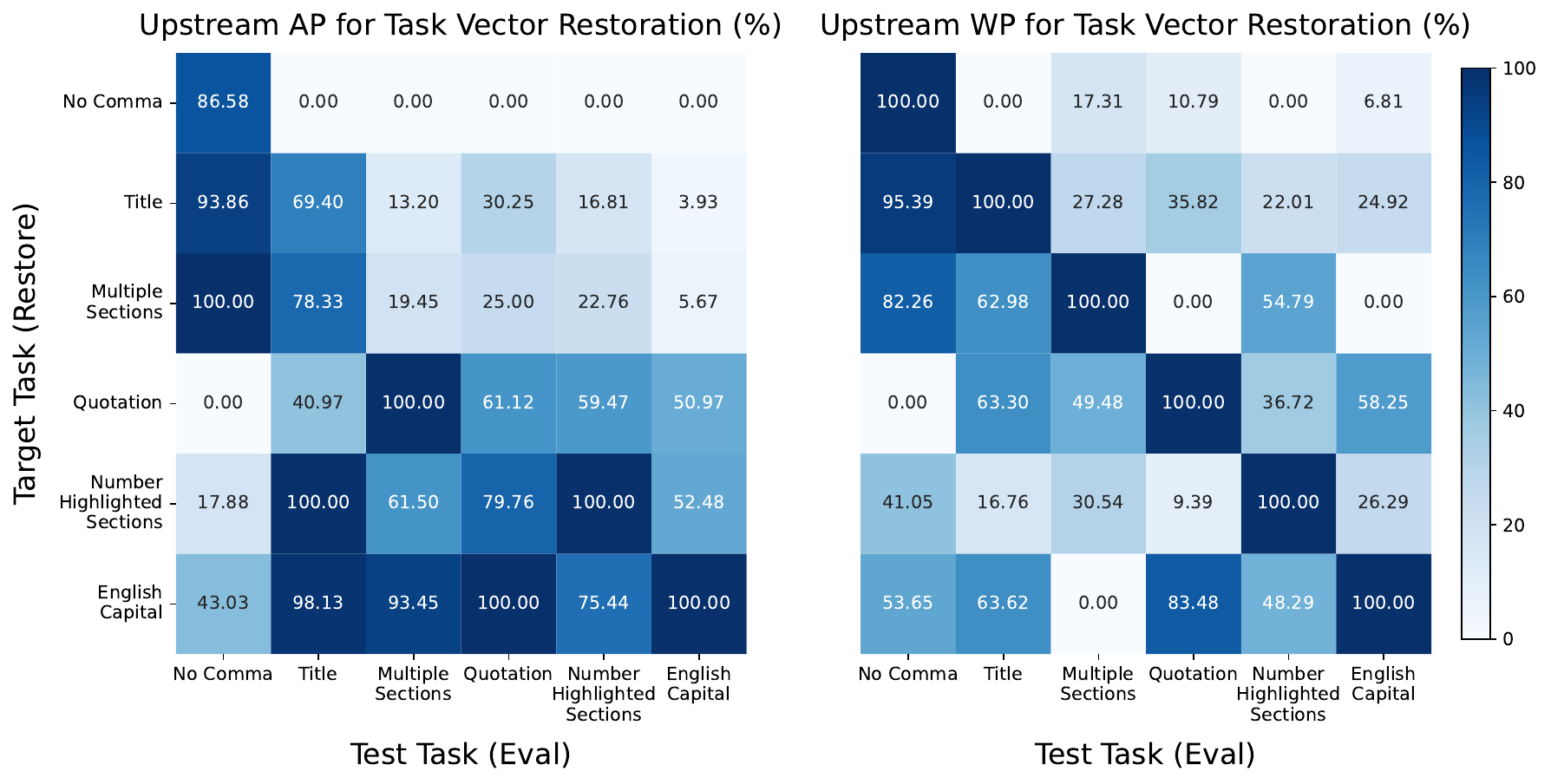}
  \vspace{-15pt} 
    \caption{Cross-task specificity of restored upstream neurons. Rows denote the restored task and columns the evaluation task; values are normalized restoration completeness. Left: AP-traced upstream neurons. Right: WP-localized upstream neurons.}  

  \label{fig:task-specific}
   \vspace{-5pt}
  
\end{figure}

At the same time, this shared organization does not imply a universal set of reused local components. Fig.~\ref{fig:task-overlap} shows that the overlap among top-ranked heads and neurons is limited, and Fig.~\ref{fig:task-specific} further shows that this distinction is functional as well as structural: when the Top-2000 upstream neurons localized for one task are globally restored and tested across all six IFEval tasks, the strongest recovery typically remains on the task from which they were localized, especially under WP. The remaining off-diagonal responses indicate partial sharing rather than fully isolated circuits. Together, these results suggest that instruction-following tasks share a common mechanistic template while instantiating it through partially task-specific local components. 

\begin{figure}[!ht]
  \centering
  \includegraphics[width=0.92\linewidth]{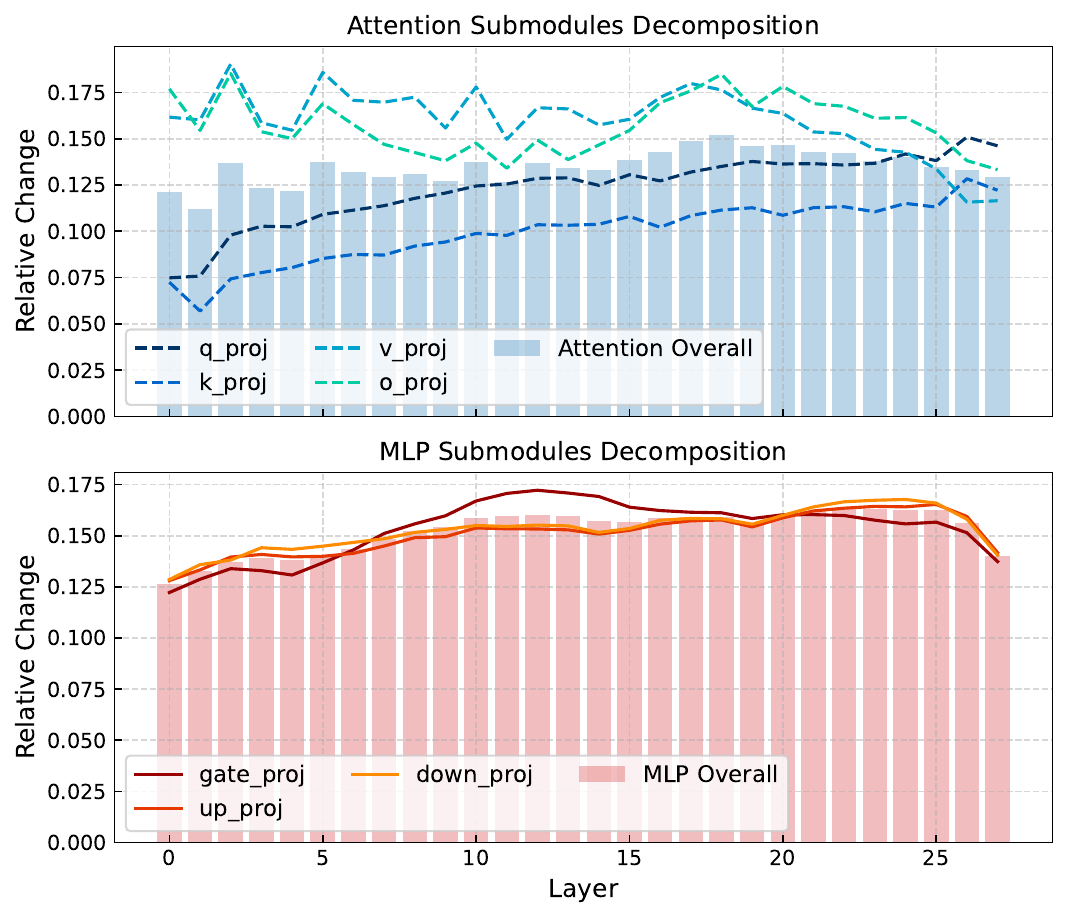}
  \vspace{-10pt}
    \caption{Layer-wise parameter changes induced by instruction tuning. Top: relative changes in attention submodules (q, k, v, o, and overall). Bottom: relative changes in MLP submodules (gate, up, down, and overall).}

  \label{fig:para-diff}
\end{figure}

Fig.~\ref{fig:para-diff} then provides a complementary static view from parameter space. Across layers, MLP modules exhibit larger average parameter changes than attention modules, with especially pronounced drift in gating-related components. While this result is not itself a causal localization test, it is consistent with the WP-based conclusion that post-training writes instruction-following behavior more strongly into sparse neuron-level parameter subsets than into attention parameters alone.

\begin{table}[h]
    \centering
    \vspace{-5pt}
    \caption{
    Comparison between exact Weight Patching and its gradient-based approximation, averaged over six tasks, including the overlap of Top-20 heads and Top-1000 neurons, total analysis time, and peak VRAM on three RTX 4090 GPUs.
    }
    \vspace{-8pt}

    \setlength{\aboverulesep}{0pt}
    \setlength{\belowrulesep}{0pt}
    \renewcommand{\arraystretch}{1.25}

    \resizebox{1\linewidth}{!}{%
    \begin{tabular}{l c c c c} 
        \specialrule{1.2pt}{0pt}{0pt}

        \rowcolor{mygray}
        & \textbf{Gradient} & \textbf{Intersection} & \textbf{Total Time} & \textbf{Peak VRAM} \\
        \rowcolor{mygray}
        \multirow{-2}{*}{\textbf{Model}} & {\textbf{Attribution}} & \textbf{(\%)} & \textbf{(Second)} & \textbf{(MiB)} \\
        \hline

        \multicolumn{5}{c}{\cellcolor{mygray!40}\textbf{Head-level}} \\
        \hline

        \multirow{2}{*}{LLama-3.2-3B} & \ding{56} & \multirow{2}{*}{$66.00$} & $384.85$ & $13150$ \\
        ~ & \ding{52}  &  & $4.03$ & $15760$ \\
        \hline 

        \multirow{2}{*}{LLama-3.1-8B} & \ding{56} & \multirow{2}{*}{$67.50$} & $1099.68$ & $34458 $ \\
        ~ & \ding{52}  &  & $4.58$ & $37747$ \\
        \hline 

        \multirow{2}{*}{LLama-2-13B} & \ding{56} & \multirow{2}{*}{$66.00$} & $3097.15$ & $51921$ \\
        ~ & \ding{52}  &  & $15.54$ & $55699$ \\
        \hline

        \multicolumn{5}{c}{\cellcolor{mygray!40}\textbf{Neuron-level}} \\
        \hline

        \multirow{2}{*}{LLama-3.2-3B} & \ding{56} & \multirow{2}{*}{$80.14$ } & $90829.54$ & $13304$ \\
        ~ & \ding{52}   &  & $5.89$ & $16096$ \\
        \hline

        \multirow{2}{*}{LLama-3.1-8B} & \ding{56} & \multirow{2}{*}{$53.33$ } & $484704.23$  & $35317$ \\
        ~ & \ding{52}   &  & $5.19$ & $40991$ \\
        \hline

        \multirow{2}{*}{LLama-2-13B} & \ding{56} & \multirow{2}{*}{$48.97$ } & $1766160.89$  & $52491$ \\
        ~ & \ding{52}   &  & $28.01$ & $67831$ \\
        
        \specialrule{1.2pt}{0pt}{0pt}
    \end{tabular}}
    \vspace{-5pt}
    \label{tab:intersection}
\end{table}

\subsection{Fidelity and Efficiency of Gradient Approximation}
\label{sec:exp_approximation}

Because the gradient-based score introduced in Sec.~\ref{sec:efficient_screening} is only a first-order surrogate to exact Weight Patching (WP), we evaluate it from two perspectives: fidelity to exact WP and computational efficiency. Table~\ref{tab:intersection} shows that the approximation preserves a substantial fraction of the most important components identified by exact WP at both granularities. At the head level, the overlap of top-ranked modules remains stable across model sizes, at roughly $66$--$68\%$, indicating that the approximation retains the dominant ranking structure needed for head screening. At the neuron level, where the table reports averages over all six tasks, the overlap is higher on the 3B model and decreases on larger models, but it still preserves a substantial portion of the exact ranking structure. Thus, while the approximation is not a substitute for exact WP, it retains sufficient ranking fidelity to support coarse-to-fine localization over both heads and neurons.

\begin{figure}[htbp]
  \centering
  \includegraphics[width=0.98\linewidth]{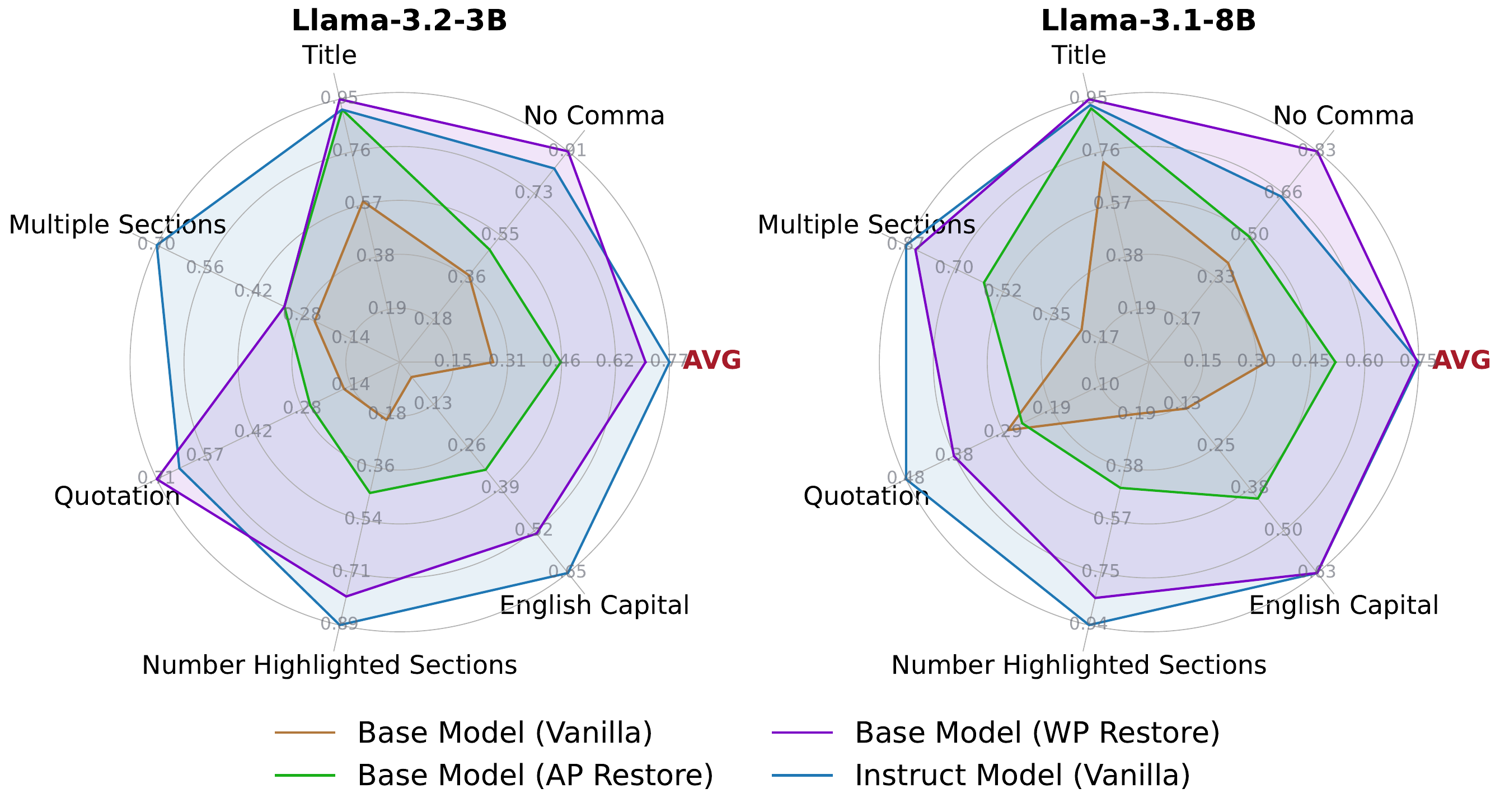}
  \vspace{-10pt}
  \caption{
  Instruction-following recovery via injection of localized critical components. Injecting the identified head and neuron subsets into base models substantially restores performance across model sizes, supporting their reuse as transferable capability carriers.}
  \label{fig:restore_transfer}
  \vspace{-5pt}
\end{figure}

Its practical value is evident at both levels, but it is especially important for neuron-scale analysis. At the head level, exact WP requires hundreds to thousands of seconds, whereas the approximation reduces screening to only a few seconds. At the neuron level, the gain is decisive: exact WP requires on the order of $10^5$ to $10^6$ seconds across the three model sizes, while the approximation remains in the range of only a few to a few tens of seconds. In this sense, the approximation makes head-level screening efficient and neuron-level screening feasible. Notably, this gain comes primarily from reduced wall-clock time rather than reduced memory footprint, since peak VRAM remains comparable and is sometimes slightly higher under the gradient-based implementation. Overall, the approximation should be viewed as a practical screening mechanism that makes fine-grained Weight Patching operational at model scale, while exact intervention remains the source of causal validation.

 \subsection{Application: Mechanism-Aware Expert Composition}
\label{sec:exp_application}

\definecolor{myblue}{HTML}{E0F0FA}
\newcommand{\up}[1]{\textcolor{green!60!black}{\scriptsize{(#1)}}}
\newcommand{\down}[1]{\textcolor{red!80!black}{\scriptsize{(#1)}}}

\begin{table*}[!th]
    \centering
    \caption{ 
    Performance comparison of the average baseline, prior merging methods, and precise fusion (ours) across four expert combinations.}
    \label{tab:model_fusion_comparison}

    \setlength{\aboverulesep}{0pt}
    \setlength{\belowrulesep}{0pt}
    \renewcommand{\arraystretch}{1.25}

    \resizebox{\linewidth}{!}{%
    \begin{tabular}{l l ccc ccc c}
        \specialrule{1.2pt}{0pt}{0pt}
        \rowcolor{mygray}
        \textbf{Model(s)} & \textbf{Method} & \textbf{HumanEval} & \textbf{MBPP} & \textbf{MMLU} & \textbf{MATH} & \textbf{GSM8K} & \textbf{IFEval} & \textbf{Avg.} \\
        \hline
        \multicolumn{9}{c}{\cellcolor{mygray!40}\textbf{Base Models}} \\
        \hline
        Base              & - & $18.90$ & $28.20$ & $52.34$ & $0.80$ & $3.00$ & $23.59$ & $21.14$ \\
        Code              & - & $19.51$ & $29.20$ & $52.96$ & $5.90$ & $24.60$ & $23.55$ & $25.95$ \\
        Instruction Tuned & - & $30.49$ & $32.20$ & $52.23$ & $7.30$ & $42.10$ & $35.51$ & $33.31$ \\
        Math              & - & $14.02$ & $23.40$ & $50.00$ & $12.90$ & $57.80$ & $21.80$ & $29.99$ \\
        \hline

        \multicolumn{9}{c}{\cellcolor{mygray!40}\textbf{Merged Models}} \\
        \hline

        \multirow{7}{*}{Code + Instruction}
        & Avg Baseline & $28.05$ & $31.60$ & $\mathbf{55.04}$ & $7.20$ & $42.10$ & \underline{$40.30$} & $34.05$ \\
        & DARE Task Arithmetic & $26.83$ \down{$-1.22$} & $34.40$ \up{$+2.80$} & $53.53$ \down{$-1.51$} & \underline{$8.40$} \up{$+1.20$} & $45.80$ \up{$+3.70$} & $33.42$ \down{$-6.88$} & $33.73$ \down{$-0.32$} \\
        & DARE Ties & $\mathbf{30.49}$ \up{$+2.44$} & \underline{$35.20$} \up{$+3.60$} & $53.40$ \down{$-1.64$} & $\mathbf{8.60}$ \up{$+1.40$} & \underline{$46.20$} \up{$+4.10$} & $33.28$ \down{$-7.02$} & \underline{$34.53$} \up{$+0.48$} \\
        & Task Arithmetic & \underline{$29.27$} \up{$+1.22$} & $33.80$ \up{$+2.20$} & $53.44$ \down{$-1.60$} & $\mathbf{8.60}$ \up{$+1.40$} & $\mathbf{47.10}$ \up{$+5.00$} & $31.60$ \down{$-8.70$} & $33.97$ \down{$-0.08$} \\
        & Ties Merging & $16.46$ \down{$-11.59$} & $23.60$ \down{$-8.00$} & $52.70$ \down{$-2.34$} & $2.70$ \down{$-4.50$} & $5.40$ \down{$-36.70$} & $24.48$ \down{$-15.82$} & $20.89$ \down{$-13.16$} \\
        & WIDEN & $26.22$ \down{$-1.83$} & $\mathbf{35.60}$ \up{$+4.00$} & \underline{$54.90$} \down{$-0.14$} & $8.30$ \up{$+1.10$} & $45.00$ \up{$+2.90$} & $30.42$ \down{$-9.88$} & $33.41$ \down{$-0.64$} \\
        & \cellcolor{myblue}\textbf{Ours} & \cellcolor{myblue}$\mathbf{30.49}$ \up{$+2.44$} & \cellcolor{myblue}\underline{$35.20$} \up{$+3.60$} & \cellcolor{myblue}$54.66$ \down{$-0.38$} & \cellcolor{myblue}$7.80$ \up{$+0.60$} & \cellcolor{myblue}$44.40$ \up{$+2.30$} & \cellcolor{myblue}$\mathbf{53.97}$ \up{$+13.67$} & \cellcolor{myblue}$\mathbf{37.75}$ \up{$+3.70$} \\
        \hline

        \multirow{7}{*}{Code + Math}
        & Avg Baseline & $15.85$ & \underline{$30.20$} & $52.22$ & $10.40$ & $49.00$ & $21.05$ & $29.79$ \\
        & DARE Task Arithmetic & $16.46$ \up{$+0.61$} & $28.60$ \down{$-1.60$} & $51.96$ \down{$-0.26$} & $\mathbf{15.10}$ \up{$+4.70$} & $\mathbf{64.70}$ \up{$+15.70$} & $22.02$ \up{$+0.97$} & $33.14$ \up{$+3.35$} \\
        & DARE Ties & $17.07$ \up{$+1.22$} & $27.40$ \down{$-2.80$} & $51.92$ \down{$-0.30$} & $14.90$ \up{$+4.50$} & $63.60$ \up{$+14.60$} & $22.53$ \up{$+1.48$} & $32.90$ \up{$+3.11$} \\
        & Task Arithmetic & $\mathbf{18.29}$ \up{$+2.44$} & $28.60$ \down{$-1.60$} & $52.10$ \down{$-0.12$} & \underline{$15.00$} \up{$+4.60$} & $\mathbf{64.70}$ \up{$+15.70$} & $21.92$ \up{$+0.87$} & \underline{$33.43$} \up{$+3.64$} \\
        & Ties Merging & $15.85$ \up{$+0.0$} & $26.80$ \down{$-3.40$} & $51.86$ \down{$-0.36$} & $14.30$ \up{$+3.90$} & $62.60$ \up{$+13.60$} & $21.63$ \up{$+0.58$} & $32.17$ \up{$+2.38$} \\
        & WIDEN & $17.07$ \up{$+1.22$} & $29.40$ \down{$-0.80$} & \underline{$53.35$} \up{$+1.13$} & $14.20$ \up{$+3.80$} & \underline{$64.40$} \up{$+15.40$} & \underline{$24.02$} \up{$+2.97$} & $\mathbf{33.74}$ \up{$+3.95$} \\
        & \cellcolor{myblue}\textbf{Ours} & \cellcolor{myblue}\underline{$17.68$} \up{$+1.83$} & \cellcolor{myblue}$\mathbf{30.60}$ \up{$+0.40$} & \cellcolor{myblue}$\mathbf{53.43}$ \up{$+1.21$} & \cellcolor{myblue}$13.00$ \up{$+2.60$} & \cellcolor{myblue}$59.90$ \up{$+19.90$} & \cellcolor{myblue}$\mathbf{24.35}$ \up{$+3.30$} & \cellcolor{myblue}$33.16$ \up{$+3.37$} \\
        \hline

        \multirow{7}{*}{Instruction + Math}
        & Avg Baseline & $20.73$ & $29.00$ & $54.33$ & $12.70$ & $61.80$ & $29.11$ & $34.61$ \\
        & DARE Task Arithmetic & $5.49$ \down{$-15.24$} & $19.00$ \down{$-10.00$} & $51.08$ \down{$-3.25$} & $9.80$ \down{$-2.90$} & $54.30$ \down{$-7.50$} & $32.35$ \up{$+3.24$} & $28.67$ \down{$-5.94$} \\
        & DARE Ties & $8.54$ \down{$-12.19$} & $23.80$ \down{$-5.20$} & $51.39$ \down{$-2.94$} & $9.20$ \down{$-3.50$} & $54.10$ \down{$-7.70$} & $33.89$ \up{$+4.78$} & $30.15$ \down{$-4.46$} \\
        & Task Arithmetic & $4.27$ \down{$-16.46$} & $20.20$ \down{$-8.80$} & $51.50$ \down{$-2.83$} & $10.00$ \down{$-2.70$} & $54.20$ \down{$-7.60$} & $31.31$ \up{$+2.20$} & $28.58$ \down{$-6.03$} \\
        & Ties Merging & $\mathbf{28.05}$ \up{$+7.32$} & $\mathbf{34.60}$ \up{$+5.60$} & \underline{$54.45$} \up{$+0.12$} & $8.70$ \down{$-4.00$} & $44.70$ \down{$-17.10$} & \underline{$34.04$} \up{$+4.93$} & $34.09$ \down{$-0.52$} \\
        & WIDEN & \underline{$24.39$} \up{$+3.66$} & $30.40$ \up{$+1.40$} & $54.20$ \down{$-0.13$} & $\mathbf{14.60}$ \up{$+1.90$} & $\mathbf{66.00}$ \up{$+4.20$} & $30.82$ \up{$+1.71$} & \underline{$36.74$} \up{$+2.13$} \\
        & \cellcolor{myblue}\textbf{Ours} & \cellcolor{myblue}$22.56$ \up{$+1.83$} & \cellcolor{myblue}\underline{$32.40$} \up{$+3.40$} & \cellcolor{myblue}$\mathbf{54.98}$ \up{$+0.65$} & \cellcolor{myblue}\underline{$14.00$} \up{$+1.30$} & \cellcolor{myblue}\underline{$64.10$} \up{$+2.30$} & \cellcolor{myblue}$\mathbf{37.87}$ \up{$+8.76$} & \cellcolor{myblue}$\mathbf{37.65}$ \up{$+3.04$} \\
        \hline

        \multirow{7}{*}{Code + Instruction + Math}
        & Avg Baseline & \underline{$23.17$} & $30.20$ & $54.19$ & $11.30$ & $54.20$ & $27.37$ & $33.41$ \\
        & DARE Task Arithmetic & $11.59$ \down{$-11.58$} & $19.60$ \down{$-10.60$} & $50.89$ \down{$-3.30$} & $9.10$ \down{$-2.20$} & $49.70$ \down{$-4.50$} & $33.20$ \up{$+5.83$} & $29.01$ \down{$-4.40$} \\
        & DARE Ties & $13.41$ \down{$-9.76$} & $21.20$ \down{$-9.00$} & $51.15$ \down{$-3.04$} & $8.70$ \down{$-2.60$} & $51.50$ \down{$-2.70$} & \underline{$35.75$} \up{$+8.38$} & $30.28$ \down{$-3.13$} \\
        & Task Arithmetic & $11.59$ \down{$-11.58$} & $19.60$ \down{$-10.60$} & $51.20$ \down{$-2.99$} & $9.00$ \down{$-2.30$} & $52.70$ \down{$-1.50$} & $32.87$ \up{$+5.50$} & $29.49$ \down{$-3.92$} \\
        & Ties Merging & $21.34$ \down{$-1.83$} & $29.20$ \down{$-1.00$} & $53.97$ \down{$-0.22$} & $6.30$ \down{$-5.00$} & $29.20$ \down{$-25.00$} & $26.95$ \down{$-0.42$} & $27.83$ \down{$-5.58$} \\
        & WIDEN & $\mathbf{25.00}$ \up{$+1.83$} & $\mathbf{33.20}$ \up{$+3.00$} & \underline{$54.58$} \up{$+0.39$} & $\mathbf{13.50}$ \up{$+2.20$} & $\mathbf{64.20}$ \up{$+10.00$} & $31.44$ \up{$+4.07$} & \underline{$36.99$} \up{$+3.58$} \\
        & \cellcolor{myblue}\textbf{Ours} & \cellcolor{myblue}$22.56$ \down{$-0.61$} & \cellcolor{myblue}\underline{$31.00$} \up{$+0.80$} & \cellcolor{myblue}$\mathbf{55.01}$ \up{$+0.82$} & \cellcolor{myblue}\underline{$13.10$} \up{$+1.80$} & \cellcolor{myblue}\underline{$62.90$} \up{$+8.70$} & \cellcolor{myblue}$\mathbf{42.14}$ \up{$+14.77$} & \cellcolor{myblue}$\mathbf{37.79}$ \up{$+4.38$} \\

        \hline
        \specialrule{1.2pt}{0pt}{0pt}
    \end{tabular}%
    }
\end{table*}

If the modules localized by Weight Patching indeed capture source-relevant parameter subsets, they should support not only post hoc explanation but also mechanism-aware reuse.
As an initial reuse test, Fig.~\ref{fig:restore_transfer} restores only the top 5\% AP- or WP-localized head and neuron parameters into the base model and compares the resulting performance with the vanilla base and instruct models. Even under this tight budget, the restored model recovers a substantial portion of instruction-following performance across the evaluated IFEval tasks, with WP-localized subsets consistently outperforming their AP-based counterparts. This suggests that the localized subsets capture compact and reusable capability-relevant structure, naturally motivating their reuse beyond post hoc analysis.

We then ask whether this reuse potential transfers to multi-expert fusion. Table~\ref{tab:model_fusion_comparison} shows that using localized source-level importance to allocate experts across heads and neurons improves expert composition over the average baseline and prior merging methods. Precise fusion achieves the best average result in three of the four expert combinations---\textit{Code + Instruction}, \textit{Instruction + Math}, and \textit{Code + Instruction + Math}---while remaining competitive on \textit{Code + Math}. The gains are largest when the fused experts are more behaviorally distinct, especially in mixtures involving the instruction expert and on IFEval, where preserving instruction-following-relevant components matters most. The smaller margin on \textit{Code + Math} is also informative: code and math experts share stronger, structured reasoning updates, so generic merging is already effective, and the room for additional gains from source-aware allocation is correspondingly smaller. Rather than favoring one expert globally, Precise Fusion assigns experts at the component level according to localized source-level importance, better preserving sparse capability-relevant components while reducing the interference introduced by uniform averaging or globally shared merge coefficients. 
Taken together, these results show that Weight Patching is useful not only for localizing capability-relevant parameter subsets, but also for reusing the resulting localized structure in mechanism-aware expert composition.
\section{Discussion}

\textbf{Mechanistic Implication for Instruction Tuning.}
The recovered hierarchy in Fig.~\ref{fig:if_hierarchy} suggests a staged division of labor, in which shallow neurons emerge as 
candidate source-side carriers of instruction-conditioned control features, mid-layer attention heads appear to aggregate and route these features into an instruction-following direction, and downstream instruction-specific modules help translate the recovered control state into response generation. Activation-space and parameter-space interventions provide complementary evidence for this picture: the former reveals where task-relevant signals become effective during inference, whereas the latter identifies which post-training parameter changes contribute most strongly to reinstating the target control state.

\textbf{Scope and Assumptions.}
The interpretation developed here rests on three explicit assumptions. First, Weight Patching is defined in a paired-model setting: a base model and a behavior-specialized counterpart share the same architecture, so the causal object is the post-training parameter difference rather than an architecture-independent decomposition of a single checkpoint. 
Second, ``source-level'' is defined relative to the intervention granularity used here—namely, head- and neuron-level parameter slices—so the recovered carriers should be interpreted as granularity-dependent rather than irreducible final origins.
Third, in the generative-control setting studied here, Weight Patching is instantiated through a vector-anchor interface because text-level counterfactuals are unstable and single-token readouts are incomplete. This interface is a practical criterion for the present analyses, not a requirement of Weight Patching in general. Under the extraction procedure used here, anchor-vector stability was not uniform across all IFEval tasks, so we focus on six representative tasks for which the analyses are reliable.

\textbf{Broader Potential.}
Although instruction following is our primary empirical setting, the framework may extend to other paired-model specializations for which a stable internal or behavioral interface can be defined; however, such extensions remain to be validated. In this paper, the resulting source-level component scores already prove useful beyond explanation, supporting mechanism-aware expert composition and achieving the best average performance among the compared methods in our evaluated fusion setting. More broadly, this source-level perspective may also aid training and post-training by identifying which parameterized components failed to acquire the intended specialization, introduced harmful interactions, or should be preferentially preserved, repaired, or reweighted. Developing such diagnostic and targeted repair mechanisms is a promising direction for future work.

\section{Related Work}
\label{sec:related_work}

\subsection{Mechanistic Localization and Causal Intervention in Language Models}

Mechanistic interpretability aims to localize model behavior to the internal components that causally realize it, rather than merely describing correlations in activations or outputs~\cite{olah2020zoom,anthropic2021mathematical,bereska2024mechanistic,rauker2023transparent,geiger2023causal_abstraction}. In language models, this perspective has motivated a broad line of work that treats attention heads, MLPs, neurons, residual-stream updates, and information paths as meaningful computational units, supported by causal abstraction and causal scrubbing as mechanism-level evaluation frameworks~\cite{induction_head,circuit_analysis,math_hanna2023gpt,math_zhang2024interpreting,neuron2_knowledge_acl2022,wang2022skill_neurons,geiger2021causal,geiger2023causal_abstraction,chan2022causal_scrubbing}. A central interventionist tool in this literature is activation patching~\cite{activation_patching}, later extended to path patching, attribution patching, and circuit-discovery methods~\cite{path_patching,attribution_patching,syed2024attribution_patching_outperforms,circuit_discovery_nips2023}; related causal tracing work has further shown that internal interventions can identify behavior-relevant modules and even support subsequent parameter editing~\cite{nips2022locating_knowledge}. However, patching results are known to depend on corruption design, readout choice, and interpretation protocol~\cite{heimersheim2024interpret_activation_patching,zhang2024best_practices_activation_patching}, and activation-space importance does not by itself establish that a capability is implemented in the patched module's own parameters~\cite{geiger2023causal_abstraction,heimersheim2024interpret_activation_patching,nips2022locating_knowledge}. Our work addresses this gap by moving the intervention from activations to weights, with the goal of identifying source-level parameter carriers rather than downstream convergence states.

\subsection{Interpreting Instruction Following and Behavioral Control}

Instruction following is one of the clearest capabilities induced by post-training and alignment~\cite{instructgpt_ouyang2022,flan_wei2022,flan_collection_longpre2023,bai2022rlhf,bai2022constitutional}, with improvements further supported by instruction collections, self-generated supervision, small high-quality alignment sets, and direct preference optimization~\cite{mishra2022natural_instructions,wang2022supernaturalinstructions,self_instruct_wang2023,zhou2023lima,rafailov2023dpo}. Recent interpretability work has begun to characterize this specialization in terms of internal control representations: instruction-sensitive probes, task and function vectors, activation steering, and representation engineering all suggest that instruction-conditioned behavior can often be monitored or manipulated through relatively structured directions in representation space~\cite{llm_know_instruction_iclr2024,task_vector_head_iclr2024,task_vector_iclr2025,steering_2024,repe_zou2023}. Other studies further show that instructions and demonstrations need not rely on identical task representations, and that instruction tuning can systematically alter routing patterns related to goals, actions, and constraints~\cite{task_vector_instr_demonstr_naacl2025,instr_behavior_change_naacl2024}. Yet this line of work remains largely focused on representations, activations, or inference-time steering, rather than on which parameters act as source-level carriers of instruction-conditioned control; this limitation is especially important for generative behaviors, where text-level counterfactuals are hard to construct, and single-token metrics are incomplete~\cite{IFEval_zhou2023instruction,heimersheim2024interpret_activation_patching,zhang2024best_practices_activation_patching}. Our work addresses this problem through a vector-anchor behavioral interface together with parameter-space intervention.

\subsection{Parameter-Space Editing, Specialized Models, and Model Merging}

Related studies on direct parameter manipulation, particularly in knowledge editing, demonstrate that localized or structured weight updates can alter model behavior without full retraining~\cite{nips2022locating_knowledge,meng2023memit,mitchell2021mend,mitchell2022serac}.
Work on task arithmetic and parameter-efficient finetuning further suggests that post-training specialization is often concentrated in structured and transferable parameter subspaces~\cite{task_arithmetic_ilharco2023,hu2021lora,dettmers2023qlora}. Beyond editing, weight-space composition has become an increasingly important paradigm for combining specialized checkpoints, including Model Soups, Fisher Merging, TIES-Merging, Model Ratatouille, Language Model Merging, and MergeKit~\cite{wortsman2022model_soups,matena2022fisher_merging,yadav2023ties,rame2023model_ratatouille,Yu2023LanguageMA,goddard2024mergekit}. These studies motivate our paired-model setting, where the difference between a base model and a specialized counterpart is treated as an analyzable object; unlike prior work on editing or merging, however, our primary goal is not to modify behavior as an end in itself, but to use parameter replacement as a causal probe for source-level mechanistic localization, while also showing that the recovered source signals can support mechanism-aware model merging.

\section{Conclusion}
\label{sec:conclusion}
 
This paper presents a framework for source-oriented mechanistic analysis in paired same-architecture large language models, centered on Weight Patching, a parameter-space intervention for identifying candidate source-side parameter carriers under a fixed input.
For generative-control tasks, we instantiate this analysis through a vector-anchor behavioral interface and a scalable first-order screening strategy, making source-oriented analysis practical for open-ended generation.
In the instruction following, the recovered evidence is consistent with a staged organization in which shallow components act as candidate source-side carriers, mid-layer modules serve as aggregation/routing interfaces, and downstream modules support execution. Beyond mechanistic explanation, the same recovered component scores also support practical reuse, including scalable fine-grained screening and WP-guided mechanism-aware expert merging.

\section*{Acknowledgment}
This work was supported by the Natural Science Foundation of China under Grant 62571507.

\bibliographystyle{IEEEtran}
\bibliography{ref}

@article{qwen25tech,
  title   = {Qwen2.5 Technical Report},
  author  = {{Qwen Team}},
  journal = {arXiv preprint arXiv:2412.15115},
  year    = {2024}
}

@misc{mistral7bv03,
  title        = {{Mistral-7B-v0.3} Model Card},
  author       = {{Mistral AI Team}},
  year         = {2024},
  howpublished = {Hugging Face model card},
  url          = {https://huggingface.co/mistralai/Mistral-7B-v0.3}
}

@article{gemma2tech,
  title   = {Gemma 2: Improving Open Language Models at a Practical Size},
  author  = {{Gemma Team}},
  journal = {arXiv preprint arXiv:2408.00118},
  year    = {2024}
}

@article{olah2020zoom,
  title={Zoom in: An introduction to circuits},
  author={Olah, Chris and Cammarata, Nick and Schubert, Ludwig and Goh, Gabriel and Petrov, Michael and Carter, Shan},
  journal={Distill},
  volume={5},
  number={3},
  pages={e00024--001},
  year={2020}
}

@inproceedings{rauker2023transparent,
  title={Toward transparent ai: A survey on interpreting the inner structures of deep neural networks},
  author={R{\"a}uker, Tilman and Ho, Anson and Casper, Stephen and Hadfield-Menell, Dylan},
  booktitle={2023 ieee conference on secure and trustworthy machine learning (satml)},
  pages={464--483},
  year={2023},
  organization={IEEE}
}

@article{geiger2021causal,
  title={Causal abstractions of neural networks},
  author={Geiger, Atticus and Lu, Hanson and Icard, Thomas and Potts, Christopher},
  journal={Advances in Neural Information Processing Systems},
  volume={34},
  pages={9574--9586},
  year={2021}
}

@article{anthropic2021mathematical,
  title={A mathematical framework for transformer circuits},
  author={Elhage, Nelson and Nanda, Neel and Olsson, Catherine and Henighan, Tom and Joseph, Nicholas and Mann, Ben and Askell, Amanda and Bai, Yuntao and Chen, Anna and Conerly, Tom and others},
  journal={Transformer Circuits Thread},
  volume={1},
  number={1},
  pages={12},
  year={2021}
}

@article{bereska2024mechanistic,
  title   = {Mechanistic Interpretability for {AI} Safety: A Review},
  author  = {Bereska, Leonard and Gavves, Efstratios},
  journal = {Transactions on Machine Learning Research (TMLR)},
  year    = {2024} 
}

@inproceedings{activation_patching,
  title     = {Interpretability in the Wild: A Circuit for Indirect Object Identification in {GPT-2} Small},
  author    = {Wang, Kevin Ro and Variengien, Alexandre and Conmy, Arthur and Shlegeris, Buck and Steinhardt, Jacob},
  booktitle = {International Conference on Learning Representations (ICLR)},
  year      = {2023}
}

@article{path_patching,
  title   = {Localizing Model Behavior with Path Patching},
  author  = {Goldowsky-Dill, Nicholas and MacLeod, Chris and Sato, Lucas and Arora, Aryaman},
  journal = {arXiv preprint arXiv:2304.05969},
  year    = {2023}
}

@article{attribution_patching,
  title={Attribution patching: Activation patching at industrial scale},
  author={Nanda, Neel},
  journal={URL: https://www. neelnanda. io/mechanistic-interpretability/attribution-patching},
  volume={15},
  pages={19},
  year={2023}
}

@inproceedings{nips2022locating_knowledge,
  title     = {Locating and Editing Factual Associations in {GPT}},
  author    = {Meng, Kevin and Bau, David and Andonian, Alex and Belinkov, Yonatan},
  booktitle = {Advances in Neural Information Processing Systems (NeurIPS)},
  volume    = {35},
  pages     = {17359--17372},
  year      = {2022}
}

@inproceedings{circuit_discovery_nips2023,
  title     = {Towards Automated Circuit Discovery for Mechanistic Interpretability},
  author    = {Conmy, Arthur and Mavor-Parker, Augustine and Lynch, Aengus and Heimersheim, Stefan and Garriga-Alonso, Adri{\`a}},
  booktitle = {Advances in Neural Information Processing Systems (NeurIPS)},
  volume    = {36},
  pages     = {16318--16352},
  year      = {2023}
}

@article{circuit_analysis,
  title   = {Does Circuit Analysis Interpretability Scale? Evidence from Multiple Choice Capabilities in Chinchilla},
  author  = {Lieberum, Tom and Rahtz, Matthew and Kram{\'a}r, J{\'a}nos and Irving, Geoffrey and Shah, Rohin and Mikulik, Vladimir},
  journal = {arXiv preprint arXiv:2307.09458},
  year    = {2023}
}

@inproceedings{math_zhang2024interpreting,
  title     = {Interpreting and Improving {Large Language Models} in Arithmetic Calculation},
  author    = {Zhang, Wei and Wan, Chaoqun and Zhang, Yonggang and Cheung, Yiu-ming and Tian, Xinmei and Shen, Xu and Ye, Jieping},
  booktitle = {International Conference on Machine Learning (ICML)},
  year      = {2024}
}

@inproceedings{math_hanna2023gpt,
  title={How does GPT-2 compute greater-than?: Interpreting mathematical abilities in a pre-trained language model},
  author={Hanna, Michael and Liu, Ollie and Variengien, Alexandre},
  booktitle={Advances in Neural Information Processing Systems},
  volume={36},
  pages={76033--76060},
  year={2023}
}

@article{induction_head,
  title   = {In-Context Learning and Induction Heads},
  author  = {Olsson, Catherine and Elhage, Nelson and Nanda, Neel and Joseph, Nicholas and DasSarma, Nova and Henighan, Tom and others},
  journal = {arXiv preprint arXiv:2209.11895},
  year    = {2022}
}

@inproceedings{zhang2024best_practices_activation_patching,
  title     = {Towards Best Practices of Activation Patching in Language Models: Metrics and Methods},
  author    = {Zhang, Fred and Nanda, Neel},
  booktitle = {International Conference on Learning Representations (ICLR)},
  year      = {2024}
}

@article{heimersheim2024interpret_activation_patching,
  title   = {How to Use and Interpret Activation Patching},
  author  = {Heimersheim, Stefan and Nanda, Neel},
  journal = {arXiv preprint arXiv:2404.15255},
  year    = {2024}
}

@article{geiger2023causal_abstraction,
  title={Causal abstraction: A theoretical foundation for mechanistic interpretability},
  author={Geiger, Atticus and Ibeling, Duligur and Zur, Amir and Chaudhary, Maheep and Chauhan, Sonakshi and Huang, Jing and Arora, Aryaman and Wu, Zhengxuan and Goodman, Noah and Potts, Christopher and others},
  journal={Journal of Machine Learning Research},
  volume={26},
  number={83},
  pages={1--64},
  year={2025}
}

@article{chan2022causal_scrubbing,
  title={Causal scrubbing: A method for rigorously testing interpretability hypotheses},
  author={Chan, Lawrence and Garriga-Alonso, Adria and Goldowsky-Dill, Nicholas and Greenblatt, Ryan and Nitishinskaya, Jenny and Radhakrishnan, Ansh and Shlegeris, Buck and Thomas, Nate},
  booktitle={AI Alignment Forum},
  volume={2},
  pages={19},
  year={2022}
}

@inproceedings{syed2024attribution_patching_outperforms,
  title={Attribution patching outperforms automated circuit discovery},
  author={Syed, Aaquib and Rager, Can and Conmy, Arthur},
  booktitle={Proceedings of the 7th BlackboxNLP Workshop: Analyzing and Interpreting Neural Networks for NLP},
  pages={407--416},
  year={2024}
}

@inproceedings{neuron2_knowledge_acl2022,
  title     = {Knowledge Neurons in Pretrained {Transformers}},
  author    = {Dai, Damai and Dong, Li and Hao, Yaru and Sui, Zhifang and Chang, Baobao and Wei, Furu},
  booktitle = {Proceedings of the 60th Annual Meeting of the Association for Computational Linguistics (ACL)},
  pages     = {8493--8502},
  year      = {2022}
}

@inproceedings{meng2023memit,
  title = {Mass-Editing Memory in a Transformer},
  author={Kevin Meng and Arnab Sen Sharma and Alex Andonian and Yonatan Belinkov and David Bau},
  booktitle = {International Conference on Learning Representations (ICLR)},
  year = {2023}
}

@inproceedings{mitchell2022serac,
  title = {Memory-Based Model Editing at Scale},
  author = {Mitchell, Eric and Lin, Charles and Bosselut, Antoine and Manning, Christopher D. and Finn, Chelsea},
  booktitle = {Proceedings of the 39th International Conference on Machine Learning (ICML)},
  year = {2022},
  pages = {15817--15831},
  publisher = {PMLR}
}

@misc{mitchell2021mend,
  title = {Fast Model Editing at Scale},
  author = {Mitchell, Eric and Lin, Charles and Bosselut, Antoine and Finn, Chelsea and Manning, Christopher D.},
  year = {2021},
  eprint = {2110.11309},
  archivePrefix = {arXiv},
  primaryClass = {cs.CL}
}

@inproceedings{llm_know_instruction_iclr2024,
  author    = {Juyeon Heo and Christina Heinze{-}Deml and Oussama Elachqar and Kwan Ho Ryan Chan and Shirley You Ren and Andrew C. Miller and Udhyakumar Nallasamy and Jaya Narain},
  title     = {Do {LLMs} "Know" Internally When They Follow Instructions?},
  booktitle = {The Thirteenth International Conference on Learning Representations (ICLR)},
  year      = {2025}
}

@inproceedings{task_vector_head_iclr2024,
  author    = {Eric Todd and Millicent L. Li and Arnab Sen Sharma and Aaron Mueller and Byron C. Wallace and David Bau},
  title     = {Function Vectors in {Large Language Models}},
  booktitle = {The Twelfth International Conference on Learning Representations (ICLR)},
  year      = {2024}
}

@inproceedings{task_vector_iclr2025,
  author    = {Alessandro Stolfo and Vidhisha Balachandran and Safoora Yousefi and Eric Horvitz and Besmira Nushi},
  title     = {Improving Instruction-Following in Language Models Through Activation Steering},
  booktitle = {The Thirteenth International Conference on Learning Representations (ICLR)},
  year      = {2025}
}

@inproceedings{steering_2024,
  title     = {Steering {Llama 2} via Contrastive Activation Addition},
  author    = {Nina Rimsky and Nick Gabrieli and Julian Schulz and Meg Tong and Evan Hubinger and Alexander Turner},
  booktitle = {Proceedings of the 62nd Annual Meeting of the Association for Computational Linguistics (ACL)},
  pages     = {15504--15522},
  year      = {2024}
}

@article{task_vector_instr_demonstr_naacl2025,
  title   = {Do Different Prompting Methods Yield a Common Task Representation in Language Models?},
  author  = {Davidson, Guy and Gureckis, Todd M. and Lake, Brenden M. and Williams, Adina},
  journal = {arXiv preprint arXiv:2505.12075},
  year    = {2025}
}

@inproceedings{instr_behavior_change_naacl2024,
  author    = {Xuansheng Wu and Wenlin Yao and Jianshu Chen and Xiaoman Pan and Xiaoyang Wang and Ninghao Liu and Dong Yu},
  title     = {From Language Modeling to Instruction Following: Understanding the Behavior Shift in {LLMs} After Instruction Tuning},
  booktitle = {Proceedings of the 2024 Conference of the North American Chapter of the Association for Computational Linguistics (NAACL)},
  pages     = {2341--2369},
  year      = {2024}
}

@article{instructgpt_ouyang2022,
  title={Training language models to follow instructions with human feedback},
  author={Ouyang, Long and Wu, Jeffrey and Jiang, Xu and Almeida, Diogo and Wainwright, Carroll and Mishkin, Pamela and Zhang, Chong and Agarwal, Sandhini and Slama, Katarina and Ray, Alex and others},
  journal={Advances in neural information processing systems},
  volume={35},
  pages={27730--27744},
  year={2022}
}

@article{flan_wei2022,
  title        = {Finetuned Language Models Are Zero-Shot Learners},
  author       = {Wei, Jason and Bosma, Maarten and Zhao, Vincent Y. and Guu, Kelvin and Yu, Adams Wei and Lester, Brian and Du, Nan and Dai, Andrew M. and Le, Quoc V.},
  journal      = {arXiv preprint arXiv:2109.01652},
  year         = {2021}
}

@inproceedings{flan_collection_longpre2023,
  title={The flan collection: Designing data and methods for effective instruction tuning},
  author={Longpre, Shayne and Hou, Le and Vu, Tu and Webson, Albert and Chung, Hyung Won and Tay, Yi and Zhou, Denny and Le, Quoc V and Zoph, Barret and Wei, Jason and others},
  booktitle={International conference on machine learning},
  pages={22631--22648},
  year={2023},
  organization={PMLR}
}

@inproceedings{self_instruct_wang2023,
  title={Self-instruct: Aligning language models with self-generated instructions},
  author={Wang, Yizhong and Kordi, Yeganeh and Mishra, Swaroop and Liu, Alisa and Smith, Noah A and Khashabi, Daniel and Hajishirzi, Hannaneh},
  booktitle={Proceedings of the 61st annual meeting of the association for computational linguistics (volume 1: long papers)},
  pages={13484--13508},
  year={2023}
}

@article{repe_zou2023,
  title        = {Representation Engineering: A Top-Down Approach to AI Transparency},
  author       = {Zou, Andy and Phan, Long and Chen, Sarah and Campbell, James and Guo, Phillip and Ren, Richard and Pan, Alexander and Yin, Xuwang and Mazeika, Mantas and Dombrowski, Ann-Kathrin and Goel, Shashwat and Li, Nathaniel and Byun, Michael J. and Wang, Zifan and Mallen, Steven and Basart, Sanmi Koyejo and Song, Dawn and Fredrikson, Matt and Kolter, J. Zico and Hendrycks, Dan},
  journal      = {arXiv preprint arXiv:2310.01405},
  year         = {2023}
}

@article{task_arithmetic_ilharco2023,
  title        = {Editing Models with Task Arithmetic},
  author       = {Ilharco, Gabriel and Ribeiro, Marco Tulio and Wortsman, Mitchell and Gururangan, Suchin and Schmidt, Ludwig and Hajishirzi, Hannaneh and Farhadi, Ali},
  journal      = {arXiv preprint arXiv:2212.04089},
  year         = {2022}
}

@article{bai2022rlhf,
  title={Training a helpful and harmless assistant with reinforcement learning from human feedback},
  author={Bai, Yuntao and Jones, Andy and Ndousse, Kamal and Askell, Amanda and Chen, Anna and DasSarma, Nova and Drain, Dawn and Fort, Stanislav and Ganguli, Deep and Henighan, Tom and others},
  journal={arXiv preprint arXiv:2204.05862},
  year={2022}
}

@article{bai2022constitutional,
  title={Constitutional ai: Harmlessness from ai feedback},
  author={Bai, Yuntao and Kadavath, Saurav and Kundu, Sandipan and Askell, Amanda and Kernion, Jackson and Jones, Andy and Chen, Anna and Goldie, Anna and Mirhoseini, Azalia and McKinnon, Cameron and others},
  journal={arXiv preprint arXiv:2212.08073},
  year={2022}
}

@article{rafailov2023dpo,
  title={Direct preference optimization: Your language model is secretly a reward model},
  author={Rafailov, Rafael and Sharma, Archit and Mitchell, Eric and Manning, Christopher D and Ermon, Stefano and Finn, Chelsea},
  journal={Advances in neural information processing systems},
  volume={36},
  pages={53728--53741},
  year={2023}
}

@article{zhou2023lima,
  title   = {LIMA: Less Is More for Alignment},
  author  = {Zhou, Chunting and others},
  journal = {arXiv preprint arXiv:2305.11206},
  year    = {2023}
}

@article{hu2021lora,
  title={Lora: Low-rank adaptation of large language models.},
  author={Hu, Edward J and Shen, Yelong and Wallis, Phillip and Allen-Zhu, Zeyuan and Li, Yuanzhi and Wang, Shean and Wang, Liang and Chen, Weizhu and others},
  journal={Iclr},
  volume={1},
  number={2},
  pages={3},
  year={2022}
}

@article{dettmers2023qlora,
  title={Qlora: Efficient finetuning of quantized llms},
  author={Dettmers, Tim and Pagnoni, Artidoro and Holtzman, Ari and Zettlemoyer, Luke},
  journal={Advances in neural information processing systems},
  volume={36},
  pages={10088--10115},
  year={2023}
}

@article{touvron2023llama2,
  title   = {{Llama 2}: Open Foundation and Fine-Tuned Chat Models},
  author  = {Touvron, Hugo and others},
  journal = {arXiv preprint arXiv:2307.09288},
  year    = {2023}
}

@article{llama3,
  title={The llama 3 herd of models},
  author={Grattafiori, Aaron and Dubey, Abhimanyu and Jauhri, Abhinav and Pandey, Abhinav and Kadian, Abhishek and Al-Dahle, Ahmad and Letman, Aiesha and Mathur, Akhil and Schelten, Alan and Vaughan, Alex and others},
  journal={arXiv preprint arXiv:2407.21783},
  year={2024}
}

@inproceedings{mmlu,
  title     = {Measuring Massive Multitask Language Understanding},
  author    = {Hendrycks, Dan and Burns, Collin and Basart, Steven and Zou, Andy and Mazeika, Mantas and Song, Dawn and Steinhardt, Jacob},
  booktitle = {International Conference on Learning Representations (ICLR)},
  year      = {2021}
}

@article{IFEval_zhou2023instruction,
  title   = {Instruction-Following Evaluation for {Large Language Models}},
  author  = {Zhou, Jeffrey and Lu, Tianjian and Mishra, Swaroop and Brahma, Siddhartha and Basu, Sujoy and Luan, Yi and Zhou, Denny and Hou, Le},
  journal = {arXiv preprint arXiv:2311.07911},
  year    = {2023}
}

@Article{Yu2023LanguageMA,
 author = {Le Yu and Yu Bowen and Haiyang Yu and Fei Huang and Yongbin Li},
 booktitle = {arXiv.org},
 journal = {ArXiv},
 title = {Language Models are Super Mario: Absorbing Abilities from Homologous Models as a Free Lunch},
 volume = {abs/2311.03099},
 year = {2023}
}

@inproceedings{matena2022fisher_merging,
  author    = {Michael S. Matena and Colin A. Raffel},
  title     = {Merging Models with Fisher-Weighted Averaging},
  booktitle = {Advances in Neural Information Processing Systems},
  year      = {2022}
}

@inproceedings{yadav2023ties,
  author    = {Prateek Yadav and Derek Tam and Leshem Choshen and Colin Raffel and Mohit Bansal},
  title     = {TIES-Merging: Resolving Interference When Merging Models},
  booktitle = {Advances in Neural Information Processing Systems},
  year      = {2023}
}

@inproceedings{goddard2024mergekit,
  author    = {Charles Goddard and Shamane Siriwardhana and Malikeh Ehghaghi and Luke Meyers and Vladimir Karpukhin and Brian Benedict and Mark McQuade and Jacob Solawetz},
  title     = {Arcee's MergeKit: A Toolkit for Merging Large Language Models},
  booktitle = {Proceedings of the 2024 Conference on Empirical Methods in Natural Language Processing: Industry Track},
  pages     = {477--485},
  year      = {2024},
  address   = {Miami, Florida, US},
  publisher = {Association for Computational Linguistics},
  doi       = {10.18653/v1/2024.emnlp-industry.36}
}

@inproceedings{wortsman2022model_soups,
  title     = {Model Soups: Averaging Weights of Multiple Fine-Tuned Models Improves Accuracy without Increasing Inference Time},
  author    = {Wortsman, Mitchell and Ilharco, Gabriel and Gadre, Samir Ya and Roelofs, Rebecca and Gontijo-Lopes, Raphael and Morcos, Ari S. and Namkoong, Hongseok and Farhadi, Ali and Carmon, Yair and Kornblith, Simon and Schmidt, Ludwig},
  booktitle = {Proceedings of the 39th International Conference on Machine Learning},
  series    = {Proceedings of Machine Learning Research},
  volume    = {162},
  pages     = {23965--23998},
  year      = {2022},
  publisher = {PMLR}
}

@article{yu2024widen,
  author  = {Le Yu and Bowen Yu and Haiyang Yu and Fei Huang and Yongbin Li},
  title   = {Extend Model Merging from Fine-Tuned to Pre-Trained Large Language Models via Weight Disentanglement},
  journal = {arXiv preprint arXiv:2408.03092},
  year    = {2024}
}

@inproceedings{nobari2025aim,
  title     = {Activation-Informed Merging of Large Language Models},
  author    = {Nobari, Amin Heyrani and Alimohammadi, Kaveh and ArjomandBigdeli, Ali and Srivastava, Akash and Ahmed, Faez and Azizan, Navid},
  booktitle = {Advances in Neural Information Processing Systems},
  year      = {2025},
  note      = {NeurIPS 2025},
  url       = {https://arxiv.org/abs/2502.02421}
}

@inproceedings{mishra2022natural_instructions,
  title     = {Cross-Task Generalization via Natural Language Crowdsourcing Instructions},
  author    = {Mishra, Swaroop and Khashabi, Daniel and Baral, Chitta and Hajishirzi, Hannaneh},
  booktitle = {Proceedings of the 60th Annual Meeting of the Association for Computational Linguistics (Volume 1: Long Papers)},
  pages     = {3470--3487},
  year      = {2022},
  address   = {Dublin, Ireland},
  publisher = {Association for Computational Linguistics},
  doi       = {10.18653/v1/2022.acl-long.244}
}

@inproceedings{wang2022supernaturalinstructions,
  title     = {Super-NaturalInstructions: Generalization via Declarative Instructions on 1600+ NLP Tasks},
  author    = {Wang, Yizhong and Mishra, Swaroop and Alipoormolabashi, Pegah and Kordi, Yeganeh and Mirzaei, Amirreza and Naik, Atharva and Ashok, Arjun and Dhanasekaran, Arut Selvan and Arunkumar, Anjana and Stap, David and Pathak, Eshaan and Karamanolakis, Giannis and Lai, Haizhi and Purohit, Ishan and Mondal, Ishani and Anderson, Jacob and Kuznia, Kirby and Doshi, Krima and Pal, Kuntal Kumar and Patel, Maitreya and Moradshahi, Mehrad and Parmar, Mihir and Purohit, Mirali and Varshney, Neeraj and Kaza, Phani Rohitha and Verma, Pulkit and Puri, Ravsehaj Singh and Karia, Rushang and Doshi, Savan and Sampat, Shailaja Keyur and Mishra, Siddhartha and Reddy A, Sujan and Patro, Sumanta and Dixit, Tanay and Shen, Xudong},
  booktitle = {Proceedings of the 2022 Conference on Empirical Methods in Natural Language Processing},
  pages     = {5085--5109},
  year      = {2022},
  address   = {Abu Dhabi, United Arab Emirates},
  publisher = {Association for Computational Linguistics},
  doi       = {10.18653/v1/2022.emnlp-main.340}
}

@inproceedings{wang2022skill_neurons,
  title     = {Finding Skill Neurons in Pre-trained Transformer-based Language Models},
  author    = {Wang, Xiaozhi and Wen, Kaiyue and Zhang, Zhengyan and Hou, Lei and Liu, Zhiyuan and Li, Juanzi},
  booktitle = {Proceedings of the 2022 Conference on Empirical Methods in Natural Language Processing},
  pages     = {11132--11152},
  year      = {2022},
  address   = {Abu Dhabi, United Arab Emirates},
  publisher = {Association for Computational Linguistics},
  doi       = {10.18653/v1/2022.emnlp-main.765}
}

@inproceedings{rame2023model_ratatouille,
  title     = {Recycling Diverse Models for Out-of-Distribution Generalization},
  author    = {Ram{\'e}, Alexandre and Ahuja, Kartik and Zhang, Jianyu and Cord, Matthieu and Bottou, L{\'e}on and Lopez-Paz, David},
  booktitle = {Proceedings of the 40th International Conference on Machine Learning},
  series    = {Proceedings of Machine Learning Research},
  volume    = {202},
  pages     = {28656--28679},
  year      = {2023},
  publisher = {PMLR}
}

@article{chen2021codex,
  title   = {Evaluating Large Language Models Trained on Code},
  author  = {Chen, Mark and Tworek, Jerry and Jun, Heewoo and Yuan, Qiming and de Oliveira Pinto, Henrique Ponde and Kaplan, Jared and Edwards, Harri and Burda, Yuri and Joseph, Nicholas and Brockman, Greg and Ray, Alex and Puri, Raul and Krueger, Gretchen and Petrov, Michael and Khlaaf, Heidy and Sastry, Girish and Mishkin, Pamela and Chan, Brooke and Gray, Scott and Ryder, Nick and Pavlov, Mikhail and Power, Alethea and Kaiser, {\L}ukasz and Bavarian, Mohammad and Winter, Clemens and Tillet, Philippe and Petroski Such, Felipe and Cummings, Dave and Plappert, Matthias and Chantzis, Fotios and Barnes, Elizabeth and Herbert-Voss, Ariel and Guss, William Hebgen and Nichol, Alex and Paino, Alex and Tezak, Nikolas and Tang, Jie and Babuschkin, Igor and Balaji, Suchir and Jain, Shantanu and Saunders, William and Hesse, Christopher and Carr, Andrew N. and Leike, Jan and Achiam, Josh and Misra, Vedant and Morikawa, Evan and Radford, Alec and Knight, Matthew and Brundage, Miles and Murati, Mira and Mayer, Katie and Welinder, Peter and McGrew, Bob and Amodei, Dario and McCandlish, Sam and Sutskever, Ilya and Zaremba, Wojciech},
  journal = {arXiv preprint arXiv:2107.03374},
  year    = {2021}
}

@article{austin2021program_synthesis,
  title   = {Program Synthesis with Large Language Models},
  author  = {Austin, Jacob and Odena, Augustus and Nye, Maxwell and Bosma, Maarten and Michalewski, Henryk and Dohan, David and Jiang, Ellen and Cai, Carrie and Terry, Michael and Le, Quoc and Sutton, Charles},
  journal = {arXiv preprint arXiv:2108.07732},
  year    = {2021}
}

@article{cobbe2021gsm8k,
  title   = {Training Verifiers to Solve Math Word Problems},
  author  = {Cobbe, Karl and Kosaraju, Vineet and Bavarian, Mohammad and Chen, Mark and Jun, Heewoo and Kaiser, {\L}ukasz and Plappert, Matthias and Tworek, Jerry and Hilton, Jacob and Nakano, Reiichiro and Hesse, Christopher and Schulman, John},
  journal = {arXiv preprint arXiv:2110.14168},
  year    = {2021}
}

@inproceedings{hendrycks2021math,
  title     = {Measuring Mathematical Problem Solving With the {MATH} Dataset},
  author    = {Hendrycks, Dan and Burns, Collin and Kadavath, Saurav and Arora, Akul and Basart, Steven and Tang, Eric and Song, Dawn and Steinhardt, Jacob},
  booktitle = {Thirty-Fifth Conference on Neural Information Processing Systems Datasets and Benchmarks Track},
  year      = {2021}
}

@misc{wizardlm,
  title        = {WizardLM: Empowering Large Language Models to Follow Complex Instructions},
  author       = {Xu, Can and Sun, Qingfeng and Zheng, Kai and Geng, Xiubo and Zhao, Pengcheng and Feng, Jiashuo and Tao, Chongyang and Jiang, Daxin},
  year         = {2023},
  eprint       = {2304.12244},
  archivePrefix= {arXiv},
  primaryClass = {cs.CL}
}

@misc{wizardmath,
  title        = {WizardMath: Empowering Mathematical Reasoning for Large Language Models via Reinforced Evol-Instruct},
  author       = {Luo, Haoran and Sun, Qingfeng and Xu, Can and Zhao, Pu and Lou, Jianguang and Tao, Chongyang and Geng, Xiubo and Lin, Qingwei and Chen, Siheng and Zhang, Dongmei},
  year         = {2023},
  eprint       = {2308.09583},
  archivePrefix= {arXiv},
  primaryClass = {cs.CL}
}

@misc{codealpaca,
  author       = {Chaudhary, Sahil},
  title        = {Code Alpaca: An Instruction-Following LLaMA Model for Code Generation},
  year         = {2023},
  howpublished = {GitHub repository},
  note         = {Accessed 2026-03-19}
}

@inproceedings{lan2024shared_circuits,
  title     = {Towards Interpretable Sequence Continuation: Analyzing Shared Circuits in Large Language Models},
  author    = {Lan, Michael and Torr, Philip and Barez, Fazl},
  booktitle = {Proceedings of the 2024 Conference on Empirical Methods in Natural Language Processing (EMNLP)},
  pages     = {12576--12601},
  year      = {2024}
}

@article{kramar2024atpstar,
  title   = {{AtP*}: An Efficient and Scalable Method for Localizing LLM Behaviour to Components},
  author  = {Kram{\'a}r, J{\'a}nos and Lieberum, Tom and Shah, Rohin and Nanda, Neel},
  journal = {arXiv preprint arXiv:2403.00745},
  year    = {2024},
  doi     = {10.48550/arXiv.2403.00745}
}

\clearpage 

\appendices 

\twocolumn[\centerline{\Large \bfseries Supplementary Material}\vspace{4ex}]

\subsection*{Appendix Overview}
This appendix provides comprehensive supplementary materials that expand on the methods, experiments, and findings presented in the main paper. The appendix is organized into detailed methodological formulations, rigorous experimental setups, and extended empirical evaluations. In addition, an accompanying codebase is provided in the supplementary materials.

\subsubsection*{A.1 \quad Textual Sections and Visualizations}
The textual appendix contains three main domains, structured as follows:

\begin{itemize}[leftmargin=1.2em, itemsep=3pt, topsep=3pt]
    \item \textbf{Detailed Methodological Formulations (Appendix A):} Provides the complete mathematical and operational formulations underlying our framework. This includes architecture-specific parameterizations for attention and MLP components (App.~\ref{app:component_interface}), the full construction of the vector-anchor interface (App.~\ref{app:anchor_construction}), definitions for exact Activation Patching and Weight Patching operators (App.~\ref{app:activation_intervention}, \ref{app:weight_patching_details}), first-order approximation derivations for efficient full-model screening (App.~\ref{app:first_order_screening}), the staged hierarchy recovery procedure (App.~\ref{app:hierarchy_recovery}), and the exact implementation rules for WP-guided model fusion (App.~\ref{app:fusion_details}).
    
    \item \textbf{Reproducibility and Experimental Setup (Appendix B):} Details the precise configurations required for reproducibility. This section covers the architectural statistics of the evaluated Llama models (App.~\ref{app:repro_models}), task descriptions and evaluation protocols from IFEval (App.~\ref{app:repro_tasks}), data splitting and vector-extraction strategies (App.~\ref{app:repro_split}), granular hyperparameters for steering and patching interventions (App.~\ref{app:repro_hparams}), component selection rules (App.~\ref{app:repro_rules}), and the explicit hardware/precision settings for model merging (App.~\ref{app:repro_merging_settings}).
    
    \item \textbf{Extended Results and Visualizations (Appendix C):} Introduces supplementary empirical evidence verifying the robustness of our core findings. This section provides rich visual and quantitative materials, including \textbf{component localization heatmaps} that demonstrate cross-scale source--aggregation separation (Llama-3.1-8B and 13B) and layer-wise parameter shifts (App.~\ref{app:extended_heatmaps}, \ref{app:distribution_scales}, \ref{app:para_diff}). It also features \textbf{causal validation radar charts} detailing behavioral recovery trajectories across diverse LLM families (Gemma, Mistral, Qwen) (App.~\ref{app:causal_validation}), vocabulary-space projections (App.~\ref{app:vocab_proj}, \ref{app:wp_downstream_proj}), selective attention analysis (App.~\ref{app:attn-instruct}), and qualitative \textbf{text generation case studies} showcasing explicit behavioral changes under targeted module ablation and restoration.
\end{itemize}

\subsubsection*{A.2 \quad Supplementary Code Implementation}
As part of the supplementary materials, we provide the code implementation for our mechanistic probes and downstream applications. The repository highlights our core contributions in parameter-space localization:

\begin{itemize}[leftmargin=1.2em, itemsep=3pt, topsep=3pt]
    \item \textbf{Weight Patching (WP) and Gradient Approximations:} Features the core implementation of our proposed exact Weight Patching operator for source-level capability localization (\texttt{./code/src/main\_wp.py}). To enable efficient full-model screening without exhaustive enumeration, we provide its first-order gradient attribution variant, WP-Attr (\texttt{./code/src/main\_wp\_attr.py}). Both projection and KL-divergence metrics are supported. For comparative analysis, we also include a cross-model Activation Patching (AP) baseline. Notably, as formalized in App.~\ref{app:activation_intervention}, unlike standard AP (which contrasts two inputs within a single model), our formulation performs a \textit{single-input, dual-model} intervention by patching instruction-conditioned activations into the base model under a fixed prompt.
    \item \textbf{WP-Guided Model Merging:} The fine-grained model fusion (\texttt{./code/src/core/model\_merging.py}), which leverages the derived head-level and neuron-level WP attribution scores to dynamically interpolate parameters across multiple SFT models. This implementation inherently supports Grouped-Query Attention (GQA) architectures.
    \item \textbf{Evaluation and Analysis Pipeline:} Example shell scripts orchestrating the workflow (e.g., \texttt{./code/sh/wp.sh} and \texttt{./code/sh/wp\_attr.sh})—from contrastive steering vector extraction and component-wise patching, to WP attribution scoring and multi-model merging.
\end{itemize}
\clearpage
\section{Detailed Methodological Formulations}
\subsection{Detailed Component Interface and Architecture-Specific Parameterization}
\label{app:component_interface}

This subsection provides the full component-level interface used throughout the method. While the main text retains only the minimal residual-stream view needed to define intervention units, the detailed formulation is important for implementation and reproducibility. In particular, we specify how attention heads and SwiGLU neurons are represented as write-back units in the shared residual stream, and how architecture-specific details such as grouped-query attention affect the definition of replaceable parameter slices. These details justify the component granularity adopted by both Weight Patching and its first-order approximation.

\paragraph{Residual-stream coordinate system.}
Let the input token sequence be $x=(x_1,\dots,x_T)$, and let
\begin{equation}
\mathbf Z_0=\mathrm{Embed}(x)\in\mathbb R^{T\times d_{\mathrm{model}}}
\label{eq:app_embed}
\end{equation}
denote the input embedding matrix. For layer $l\in\{1,\dots,L\}$, let $\mathbf Z_l\in\mathbb R^{T\times d_{\mathrm{model}}}$ denote the residual-state matrix after layer $l$, and let $\mathbf z_l^{(t)}\in\mathbb R^{d_{\mathrm{model}}}$ denote its row at position $t$. Under a pre-norm decoder, the residual stream is updated as
\begin{align}
\mathbf Z'_l
&=
\mathbf Z_{l-1}
+
\mathrm{Attn}_l\!\bigl(\mathrm{Norm}(\mathbf Z_{l-1})\bigr),
\label{eq:app_residual_attn}
\\
\mathbf Z_l
&=
\mathbf Z'_l
+
\mathrm{MLP}_l\!\bigl(\mathrm{Norm}(\mathbf Z'_l)\bigr).
\label{eq:app_residual_mlp}
\end{align}
This shared residual-stream space serves as the common coordinate system in which all later activation-space and parameter-space interventions are defined.

\paragraph{Attention heads as write-back units.}
Let $\mathbf X=\mathrm{Norm}(\mathbf Z_{l-1})$ denote the normalized input to layer $l$. For query head $h$, the head-level query slice is
\begin{equation}
\mathbf Q^{(l,h)}
=
\mathbf X W_Q^{(l,h)}
\in\mathbb R^{T\times d_k}.
\label{eq:app_head_q}
\end{equation}
In standard multi-head attention, the key and value slices can be written as
\begin{equation}
\mathbf K^{(l,h)}=\mathbf X W_K^{(l,h)},
\qquad
\mathbf V^{(l,h)}=\mathbf X W_V^{(l,h)}.
\label{eq:app_head_kv_mha}
\end{equation}
The resulting head output is
\begin{equation}
\mathbf O^{(l,h)}
=
\mathrm{Softmax}\!\left(
\frac{\mathbf Q^{(l,h)}(\mathbf K^{(l,h)})^\top}{\sqrt{d_k}}
+\mathbf M
\right)\mathbf V^{(l,h)},
\label{eq:app_head_output}
\end{equation}
where $\mathbf M$ is the causal mask. The head writes back to the residual stream through its output projection:
\begin{equation}
\Delta\mathbf Z_{\mathrm{attn}}^{(l,h)}
=
\mathbf O^{(l,h)}W_O^{(l,h)}
\in\mathbb R^{T\times d_{\mathrm{model}}}.
\label{eq:app_head_writeback}
\end{equation}
This write-back form makes the attention head a natural component-level intervention unit.

\paragraph{Grouped-query attention and head granularity.}
In architectures with grouped-query attention, multiple query heads may share the same key and value projections. In that setting, replacing $W_K$ or $W_V$ locally would affect an entire group of query heads simultaneously, rather than an individual head. This breaks the intended component granularity and introduces causal ambiguity in head-level parameter interventions. For this reason, the parameter-space intervention in the main text and Appendix~\ref{app:weight_patching_details} restricts the head-level replaceable slice to the head-specific query and output parameters, while treating shared key/value parameters as outside the scope of a clean single-head replacement.

\paragraph{SwiGLU MLPs and neuron write-back.}
For the SwiGLU MLP used in modern decoder-only LLMs, the feed-forward block at layer $l$ is
\begin{equation}
\mathrm{MLP}_l(\mathbf X)
=
\Big(
\mathrm{SiLU}(\mathbf X W_{\mathrm{gate}}^{(l)})
\odot
(\mathbf X W_{\mathrm{up}}^{(l)})
\Big)W_{\mathrm{down}}^{(l)}.
\label{eq:app_swiglu}
\end{equation}
Let $N^{(l,j)}$ denote the $j$-th hidden unit. Its contribution can be written as
\begin{equation}
\Delta\mathbf Z_{\mathrm{mlp}}^{(l,j)}
=
\mathbf a^{(l,j)}W_{\mathrm{down}}^{(l)}[j,:]
\in\mathbb R^{T\times d_{\mathrm{model}}},
\label{eq:app_neuron_writeback}
\end{equation}
where $\mathbf a^{(l,j)}$ denotes the corresponding hidden activation induced jointly by the gate and up-projection columns. Thus, an MLP neuron naturally forms a parameter-grounded feature-detection, gating, and write-back unit.

\paragraph{Candidate component set.}
Taken together, both attention heads and MLP neurons ultimately appear as additive component-level updates in the same residual-stream space. We therefore define the candidate component set as
\begin{equation}
\mathcal C
=
\left\{
H^{(l,h)}
\right\}
\cup
\left\{
N^{(l,j)}
\right\},
\label{eq:app_component_set}
\end{equation}
where the first subset ranges over attention heads and the second over MLP neurons. All exact and first-order interventions in the paper are defined with respect to this common component interface.

\subsection{Anchor Construction, Layer Selection, and Steering Calibration}
\label{app:anchor_construction}

This subsection gives the full construction of the vector-anchor behavioral interface used in the main text. The key idea is to replace unstable text-level counterfactuals with an internal control representation extracted from contrastive instruction-present and instruction-removed pairs. We detail how layer-wise task directions are computed, how the anchor layer and anchor position are selected, and how steering magnitude is calibrated in an input-adaptive way so that the resulting interface can be used consistently across steering, localization, restoration, and ablation.

\paragraph{Paired data construction.}
Let $\mathcal D_{\mathrm{inst}}=\{(x_r^{(i)},x_{cf}^{(i)})\}_{i=1}^{N}$ denote a paired dataset, where $x_r^{(i)}$ is an instruction-present input and $x_{cf}^{(i)}$ is its instruction-removed counterpart with matched context. In all analyses, the extraction split used to construct task directions is kept disjoint from the evaluation split used for steering, localization, ablation, and restoration, so that the vector-anchor interface is not defined and tested on the same examples.

\paragraph{Layer-wise task directions.}
For each candidate layer $l\in\mathcal L$, we extract an instruction direction by averaging residual-stream differences at a designated anchor position $t_a$:
\begin{equation}
\mathbf v^{(l)}
=
\frac{1}{N}
\sum_{i=1}^{N}
\left[
\mathbf z_l^{(t_a)}(x_r^{(i)}\mid M_{\mathrm{sft}})
-
\mathbf z_l^{(t_a)}(x_{cf}^{(i)}\mid M_{\mathrm{sft}})
\right].
\label{eq:app_layer_direction}
\end{equation}
This construction suppresses shared semantic content while preserving the control-relevant variation induced by the instruction itself.

\paragraph{Anchor position and anchor layer selection.}
The anchor position $t_a$ is the designated readout position used to evaluate whether the control representation has been formed. Given the layer-wise directions in Eq.~\eqref{eq:app_layer_direction}, the anchor layer is selected according to the behavioral recovery achieved by steering along each candidate direction:
\begin{equation}
l_a
=
\arg\max_{l\in\mathcal L}
\mathrm{Recover}\!\left(\mathbf v^{(l)}\right),
\label{eq:app_anchor_layer_selection}
\end{equation}
where $\mathrm{Recover}(\cdot)$ denotes the steering-based behavioral recovery score evaluated on a held-out split. The final task vector is then set to
\begin{equation}
\mathbf v
=
\mathbf v^{(l_a)}.
\label{eq:app_task_vector}
\end{equation}
In practice, the selected anchor layers tend to lie in middle-to-late layers, supporting the interpretation that the extracted vector behaves as an internal control representation rather than a shallow lexical residue.

\paragraph{Anchor utility and anchor gap.}
Based on the selected anchor layer and task vector, the anchor utility is defined as
\begin{equation}
\mathcal F_a(M,x)
=
\operatorname{sim}\!\bigl(\mathbf z_a(x\mid M),\mathbf v\bigr),
\qquad
\mathbf z_a(x\mid M)\equiv \mathbf z_{l_a}^{(t_a)}(x\mid M),
\label{eq:app_anchor_utility}
\end{equation}
and the specialized-to-base anchor gap is
\begin{equation}
G(x)
=
\mathcal F_a(M_{\mathrm{sft}},x)-\mathcal F_a(M_{\mathrm{base}},x).
\label{eq:app_anchor_gap_a2}
\end{equation}
This gap serves as the normalization term used throughout exact and first-order restoration analysis.

\paragraph{Input-adaptive steering calibration.}
To steer the base model using the extracted task vector, we first normalize the direction:
\begin{equation}
\hat{\mathbf v}
=
\frac{\mathbf v}{\|\mathbf v\|_2}.
\label{eq:app_normalized_task_vector}
\end{equation}
Next, we estimate the average projection strength of instructed examples along this direction:
\begin{equation}
\bar\mu
=
\frac{1}{N}
\sum_{i=1}^{N}
\left\langle
\mathbf z_a(x_r^{(i)}\mid M_{\mathrm{sft}}),
\hat{\mathbf v}
\right\rangle.
\label{eq:app_avg_projection}
\end{equation}
Given an input $x$, the input-adaptive steering coefficient is then defined as
\begin{equation}
\alpha(x)
=
\bar\mu
-
\left\langle
\mathbf z_a(x\mid M_{\mathrm{base}}),
\hat{\mathbf v}
\right\rangle.
\label{eq:app_alpha}
\end{equation}
The corresponding steering intervention at the anchor is
\begin{equation}
\tilde{\mathbf z}_a(x\mid M_{\mathrm{base}})
\leftarrow
\mathbf z_a(x\mid M_{\mathrm{base}})
+
\alpha(x)\hat{\mathbf v}.
\label{eq:app_anchor_steering}
\end{equation}
This calibration ensures that the intervention magnitude adapts to the current base-model state rather than using a fixed global scale across all examples.

\paragraph{Role of the interface.}
If the steered base model reliably recovers the target behavior, the extracted direction is treated as a valid functional anchor for subsequent mechanistic localization. The vector-anchor interface therefore serves as a shared internal behavioral criterion under which steering, activation-space restoration, parameter-space restoration, and downstream hierarchy analysis can all be compared consistently.

\paragraph{Summary.}
Taken together, Eqs.~\eqref{eq:app_layer_direction}--\eqref{eq:app_anchor_steering} define the full vector-anchor construction used in the paper. The paired instruction/no-instruction contrast yields a task direction, anchor-layer selection turns this direction into a stable readout interface, and input-adaptive calibration makes the resulting intervention usable across examples without introducing an arbitrary global steering scale.

\subsection{Exact Activation-Space Intervention and Cross-Model Activation Patching}
\label{app:activation_intervention}

This subsection provides the exact activation-space intervention protocol used as auxiliary evidence in the main text. Activation patching is not the primary source-level method of this paper, but it remains useful for identifying where instruction-conditioned signals become causally effective during inference. We therefore distinguish the standard clean/corrupted activation-patching template from the cross-model activation-patching setting used here, in which activations from the specialized model are patched into the base-model run under a fixed input.

\paragraph{Standard activation patching.}
In the conventional setting, activation patching is defined on a single model with a clean input $x_{\mathrm{clean}}$ and a corrupted input $x_{\mathrm{corr}}$. For a component $c\in\mathcal C$, the patched run is
\begin{equation}
M\!\left(
x_{\mathrm{corr}}
\mid
do(\mathbf z_c \leftarrow \mathbf z_c(x_{\mathrm{clean}}))
\right).
\label{eq:app_standard_ap}
\end{equation}
Given a generic utility function $\mathcal F$, the corresponding restoration effect is
\begin{equation}
\begin{split}
\Delta\mathcal F_{\mathrm{patch}}(c)
&= \mathcal F\!\bigl(
M(x_{\mathrm{corr}}\mid do(\mathbf z_c \leftarrow \mathbf z_c(x_{\mathrm{clean}}))),
x_{\mathrm{corr}}
\bigr) \\
&\quad - \mathcal F\!\bigl(M(x_{\mathrm{corr}}),x_{\mathrm{corr}}\bigr).
\end{split}
\label{eq:app_standard_ap_effect}
\end{equation}
This form measures whether a component's clean-run activation is sufficient to restore behavior in the corrupted run.

\paragraph{Cross-model activation patching under fixed input.}
In the present paired-model setting, the goal is different. Rather than contrasting two inputs within one model, we compare two models under the same instruction-present input $x$. For any component $c\in\mathcal C$, let
\begin{equation}
\mathbf z_c(x\mid M_{\mathrm{base}})
\qquad\text{and}\qquad
\mathbf z_c(x\mid M_{\mathrm{sft}})
\label{eq:app_cross_model_states}
\end{equation}
denote the corresponding component activations in the base and specialized models. The cross-model activation-patched run is then defined as
\begin{equation}
\widetilde M_{\mathrm{base}}^{(c)}(x)
:=
M_{\mathrm{base}}\!\left(
x \mid do\!\bigl(
\mathbf z_c \leftarrow \mathbf z_c(x\mid M_{\mathrm{sft}})
\bigr)
\right).
\label{eq:app_cross_model_patch}
\end{equation}
This intervention asks whether the specialized activation of a local component is sufficient to restore the anchor-side control representation when transplanted into the base-model run under the same input.

\paragraph{Anchor-based utility for activation-space restoration.}
The paper evaluates activation-space restoration through the same vector-anchor interface used in Weight Patching. The anchor utility is
\begin{equation}
\mathcal F_a(M,x)
=
\operatorname{sim}\!\left(
\mathbf z_{l_a}^{(t_a)}(x\mid M),
\mathbf v
\right),
\label{eq:app_anchor_utility_a3}
\end{equation}
and the corresponding specialized-to-base gap is
\begin{equation}
G(x)
=
\mathcal F_a(M_{\mathrm{sft}},x)-\mathcal F_a(M_{\mathrm{base}},x).
\label{eq:app_anchor_gap_a3}
\end{equation}
Using this shared interface, the exact activation-side restoration effect of component $c$ is defined as
\begin{equation}
E_a(c)
=
\mathbb E_{x\sim\mathcal D_{\mathrm{inst}}}
\left[
\frac{
\mathcal F_a(\widetilde M_{\mathrm{base}}^{(c)},x)
-
\mathcal F_a(M_{\mathrm{base}},x)
}{
G(x)
}
\right].
\label{eq:app_exact_activation_effect}
\end{equation}
Large $E_a(c)$ indicates that the component plays an important role in forming, routing, or restoring the task-relevant control signal during inference.

\paragraph{Interpretation.}
The role of cross-model activation patching differs from that of Weight Patching. Activation patching identifies where behavior-relevant information becomes causally effective during inference, such as restoration points, routing modules, or aggregation bottlenecks. It does not, by itself, establish that the corresponding capability is implemented in the patched component's own parameters. This is precisely why activation-space evidence is used in the paper as a complementary tool for hierarchy recovery, whereas source-level parameter causality is assigned only through Weight Patching.

\paragraph{Use in hierarchy analysis.}
In the main text and Appendix~\ref{app:hierarchy_recovery}, exact activation-side restoration is used in two ways. First, components with large $E_a(c)$ are treated as candidates for aggregation and routing modules. Second, the specialized activation difference induced by these components provides the supply-side signal used in the later supply--need matching procedure. In this sense, activation-space intervention is not a separate competing method, but an auxiliary mechanism for identifying where the control signal flows once it has been written.

\paragraph{Summary.}
Taken together, Eqs.~\eqref{eq:app_standard_ap}--\eqref{eq:app_exact_activation_effect} define the exact activation-space intervention protocol used in the paper. The standard clean/corrupted formulation provides the general causal template, and the cross-model fixed-input version adapts that template to the paired-model setting studied here. The resulting activation-side effect serves as evidence for information flow, restoration, and aggregation, complementing the source-level parameter evidence provided by Weight Patching.

\subsection{Detailed Weight Patching Operator and Multi-Component Restoration}
\label{app:weight_patching_details}

This subsection provides the full parameter-space intervention operator underlying Weight Patching. The main text focuses on the single-component case because source localization is defined at the component level, but the same replacement rule extends naturally to multi-component restoration and ablation. We therefore formalize both the single-component and set-wise parameter-patched models, specify the exact parameter slices used for attention heads and MLP neurons, and clarify why the chosen slices align with the functional granularity of the mechanistic units studied in this paper.

\paragraph{Single-component and set-wise parameter replacement.}
Let $M_{\mathrm{base}}$ and $M_{\mathrm{sft}}$ denote the base model and the specialized counterpart, with parameter sets $\theta_{\mathrm{base}}$ and $\theta_{\mathrm{sft}}$, respectively. For any component $c\in\mathcal C$, let $\Theta^{(c)}$ denote its replaceable parameter slice. More generally, for a component set $\mathcal S\subseteq\mathcal C$, define the union slice
\begin{equation}
\Theta^{(\mathcal S)}
=
\bigcup_{c\in\mathcal S}\Theta^{(c)}.
\label{eq:app_union_slice}
\end{equation}
The corresponding set-wise replaced parameter tensor is defined elementwise as
\begin{equation}
\theta_{\mathrm{base}\leftarrow\mathrm{sft}}^{(\mathcal S)}[p]
=
\begin{cases}
\theta_{\mathrm{sft}}[p], & p\in \Theta^{(\mathcal S)},\\
\theta_{\mathrm{base}}[p], & p\notin \Theta^{(\mathcal S)}.
\end{cases}
\label{eq:app_replaced_params}
\end{equation}
This induces the set-wise parameter-patched model
\begin{equation}
M_{\mathrm{base}}^{(\mathcal S\leftarrow \mathrm{sft})}
:=
\operatorname{Replace}\!\left(
M_{\mathrm{base}},M_{\mathrm{sft}};\Theta^{(\mathcal S)}
\right).
\label{eq:app_set_patch_model}
\end{equation}
The single-component case is recovered by setting $\mathcal S=\{c\}$:
\begin{equation}
M_{\mathrm{base}}^{(c\leftarrow \mathrm{sft})}
:=
\operatorname{Replace}\!\left(
M_{\mathrm{base}},M_{\mathrm{sft}};\Theta^{(c)}
\right).
\label{eq:app_component_patch_model}
\end{equation}

The distinction between Eqs.~\eqref{eq:app_set_patch_model} and \eqref{eq:app_component_patch_model} is operationally important. Source localization is defined at the single-component level, because the goal is to test whether the specialized parameters of an individual head or neuron are sufficient to reintroduce the target control signal. By contrast, Eq.~\eqref{eq:app_set_patch_model} is used only when evaluating multi-component restoration or ablation, such as top-$K$ recovery experiments, where the objective is to test whether a sparse set of already-ranked components can collectively close a substantial fraction of the anchor gap.

\paragraph{Attention-head parameter slices.}
For an attention head $H^{(l,h)}$, the replaceable slice is defined as
\begin{equation}
\Theta^{(H^{(l,h)})}
=
\left\{
W_Q^{(l,h)},\,W_O^{(l,h)}
\right\}.
\label{eq:app_head_slice}
\end{equation}
This design deliberately excludes $W_K$ and $W_V$. In standard multi-head attention, one could in principle define a larger head-level slice that also includes key and value projections. However, in architectures with grouped-query attention, key and value projections are shared across multiple query heads. A local replacement of $W_K$ or $W_V$ would therefore affect a group of query heads simultaneously, breaking the intended component granularity and introducing causal ambiguity. Restricting the head-level slice to the head-specific query and output projections preserves a clean intervention unit while remaining faithful to the head's role in reading from and writing back to the residual stream.

When needed for implementation, let $\mathcal I_h$ denote the index slice corresponding to query head $h$. Then Eq.~\eqref{eq:app_head_slice} can be understood as replacing the columns of $W_Q^{(l)}$ indexed by $\mathcal I_h$ together with the rows of $W_O^{(l)}$ indexed by $\mathcal I_h$. This index-level view is also used in Appendix~\ref{app:first_order_screening} to derive head-level first-order aggregation formulas.

\paragraph{MLP-neuron parameter slices.}
For an MLP neuron $N^{(l,j)}$ in a SwiGLU block, the replaceable slice is defined as
\begin{equation}
\Theta^{(N^{(l,j)})}
=
\left\{
W_{\mathrm{gate}}^{(l)}[:,j],\,
W_{\mathrm{up}}^{(l)}[:,j],\,
W_{\mathrm{down}}^{(l)}[j,:]
\right\}.
\label{eq:app_neuron_slice}
\end{equation}
This slice preserves the full functional interface of the neuron. The gate and up-projection columns determine which input patterns activate the neuron and how strongly they are gated, while the down-projection row determines the direction in residual-stream space to which the activated feature is written back. Replacing only one of these three parts would no longer correspond to a coherent neuron-level intervention. Equation~\eqref{eq:app_neuron_slice} therefore treats the neuron as a complete feature-detection, gating, and write-back unit.

\paragraph{Source-level exact restoration effect.}
Under the vector-anchor behavioral interface defined in the main text, the source-level effect of component $c$ is measured by how much of the anchor gap is restored when only that component's specialized parameters are transplanted into the base model:
\begin{equation}
E_w(c)
=
\mathbb E_{x\sim\mathcal D_{\mathrm{inst}}}
\left[
\frac{
\mathcal F_a\!\left(M_{\mathrm{base}}^{(c\leftarrow \mathrm{sft})},x\right)
-
\mathcal F_a\!\left(M_{\mathrm{base}},x\right)
}{
G(x)
}
\right],
\label{eq:app_wp_effect}
\end{equation}
where
\begin{equation}
G(x)
=
\mathcal F_a(M_{\mathrm{sft}},x)-\mathcal F_a(M_{\mathrm{base}},x)
\label{eq:app_anchor_gap}
\end{equation}
is the specialized-to-base anchor gap. A large value of $E_w(c)$ indicates that the specialized parameters of component $c$ are sufficient to recover a substantial fraction of the control representation missing from the base model under the same input. This is the central source-level causal criterion used throughout the paper.

For a component set $\mathcal S\subseteq\mathcal C$, the corresponding set-wise restoration effect is defined analogously:
\begin{equation}
E_w(\mathcal S)
=
\mathbb E_{x\sim\mathcal D_{\mathrm{inst}}}
\left[
\frac{
\mathcal F_a\!\left(M_{\mathrm{base}}^{(\mathcal S\leftarrow \mathrm{sft})},x\right)
-
\mathcal F_a\!\left(M_{\mathrm{base}},x\right)
}{
G(x)
}
\right].
\label{eq:app_wp_effect_set}
\end{equation}
Equation~\eqref{eq:app_wp_effect_set} is not used to define the basic localization unit, but it is useful in two situations. First, it supports multi-component validation, where one tests whether a sparse set of top-ranked components can jointly recover the anchor-side control signal. Second, it supports targeted ablations or restoration studies that probe whether the recovered source carriers behave collectively as a reusable mechanism rather than as isolated components.

\paragraph{Relation to the sparse source-localization objective.}
The single-component and set-wise formulations above also connect to the sparse restoration objective in a more explicit optimization form. At a high level, source localization seeks a sparse component set whose replacement substantially restores the anchor representation in the base model:
\begin{equation}
\begin{aligned}
\min_{\mathcal S\subseteq\mathcal C}\quad & |\mathcal S|\\
\text{s.t.}\quad
& \operatorname{sim}\!\left(
\mathbf z_{l_a}^{(t_a)}\!\left(x\mid M_{\mathrm{base}}^{(\mathcal S\leftarrow \mathrm{sft})}\right),
\mathbf v
\right)
\ge \tau .
\end{aligned}
\label{eq:app_sparse_source_objective}
\end{equation}
In practice, this objective is not solved directly. Instead, the paper proceeds by first ranking components using first-order screening, then validating top candidates with the exact single-component effect in Eq.~\eqref{eq:app_wp_effect}, and finally using Eq.~\eqref{eq:app_wp_effect_set} only for sparse set restoration or ablation.

\paragraph{Summary.}
Taken together, Eqs.~\eqref{eq:app_set_patch_model}--\eqref{eq:app_wp_effect_set} define Weight Patching as a parameter-space intervention family centered on component-aligned replacement. The single-component case provides the source-level causal primitive, the set-wise case supports sparse restoration and ablation, and the head/neuron slice definitions ensure that the intervention granularity remains consistent with the mechanistic units analyzed throughout the paper.

\subsection{First-Order Approximation, Head/Neuron Aggregation, and Full-Model Screening}
\label{app:first_order_screening}

This subsection develops the first-order approximation used to scale source localization to all candidate heads and neurons. The exact activation-space and parameter-space interventions defined in the main text provide the most direct causal evidence, but applying them exhaustively to every component in a large language model is computationally expensive, especially at neuron scale. To make full-model screening practical, we approximate each intervention by the local first-order response of the anchor utility. This yields efficient activation-side and weight-side attribution scores that preserve the distinction between dynamic signal flow and parameter-side responsibility, while allowing exact AP/WP validation to be restricted to a small top-ranked subset.

\paragraph{A unified first-order view.}
Both activation patching and Weight Patching can be interpreted as perturbations applied around the base model. Let $\mathcal F_a(M_{\mathrm{base}},x)$ denote the anchor utility defined in the main text. For a small perturbation $\delta u$ applied to an intermediate state or a parameter slice, the corresponding utility change admits the first-order approximation
\begin{equation}
\Delta \mathcal F_a
\approx
\left\langle
\nabla_u \mathcal F_a(M_{\mathrm{base}},x),
\delta u
\right\rangle.
\label{eq:app_first_order_template}
\end{equation}
This form makes explicit that the intervention effect is approximated by the alignment between a local sensitivity direction and the perturbation induced by the specialized model. The approximation is used only for screening: exact source-level conclusions still rely on the exact Weight Patching effect in the main text.

\paragraph{Activation-side first-order approximation.}
For an input $x$ and component $c\in\mathcal C$, define the cross-model activation difference
\begin{equation}
\Delta \mathbf z_c(x)
=
\mathbf z_c(x\mid M_{\mathrm{sft}})
-
\mathbf z_c(x\mid M_{\mathrm{base}}).
\label{eq:app_delta_z}
\end{equation}
A first-order expansion of the anchor utility around the base-model activation gives
\begin{equation}
\mathcal F_a(\mathbf z_c+\Delta \mathbf z_c,x)
-
\mathcal F_a(\mathbf z_c,x)
\approx
\left\langle
\nabla_{\mathbf z_c}\mathcal F_a(M_{\mathrm{base}},x),
\Delta \mathbf z_c(x)
\right\rangle.
\label{eq:app_aap_taylor}
\end{equation}
Accordingly, the activation-side attribution score is defined as
\begin{equation}
\mathrm{Attr}_{\mathrm{act}}(c)
=
\mathbb E_{x\sim\mathcal D_{\mathrm{inst}}}
\left[
\frac{
\left\langle
\nabla_{\mathbf z_c}\mathcal F_a(M_{\mathrm{base}},x),
\Delta \mathbf z_c(x)
\right\rangle
}{
G(x)
}
\right].
\label{eq:app_aap_score}
\end{equation}
Large $\mathrm{Attr}_{\mathrm{act}}(c)$ indicates that the specialized activation shift of component $c$ is well aligned with the local direction that increases the anchor utility, making the component a strong candidate for activation-side restoration, aggregation, or routing analysis.

\paragraph{Weight-side first-order approximation.}
Let $\Delta\theta=\theta_{\mathrm{sft}}-\theta_{\mathrm{base}}$ denote the post-training parameter change. For a component $c$ with parameter slice $\Theta^{(c)}$, the first-order approximation to the weight-side effect is
\begin{equation}
\mathrm{Attr}_{\mathrm{wt}}(c)
=
\mathbb E_{x\sim\mathcal D_{\mathrm{inst}}}
\left[
\frac{
\sum_{p\in\Theta^{(c)}}
\Delta\theta_p
\frac{\partial \mathcal F_a(M_{\mathrm{base}},x)}{\partial \theta_p}
}{
G(x)
}
\right].
\label{eq:app_wap_score}
\end{equation}
Equation~\eqref{eq:app_wap_score} measures whether the post-training parameter update on slice $\Theta^{(c)}$ points in a direction that increases the anchor utility at the base-model parameter point. Unlike Eq.~\eqref{eq:app_aap_score}, which approximates activation-side restoration, Eq.~\eqref{eq:app_wap_score} is directly aligned with the source-level question of where the target control signal is written in parameters.

\paragraph{Head-level aggregation.}
For an attention head $H^{(l,h)}$, the replaceable slice contains the head-specific query and output parameters only. Let $\mathcal I_h$ denote the index slice corresponding to query head $h$. The head-level first-order score is aggregated from the query and output parts:
\begin{equation}
\mathrm{Attr}_{\mathrm{wt}}(H^{(l,h)})
=
\mathrm{Attr}_{Q}(l,h)
+
\mathrm{Attr}_{O}(l,h),
\label{eq:app_head_attr_sum}
\end{equation}
where
\begin{gather}
\mathrm{Attr}_{Q}(l,h) = \nonumber \\
\mathbb E_{x\sim\mathcal D_{\mathrm{inst}}}
\left[
\frac{
\sum
\Bigl(
\bigl(
\Delta W_{Q}^{(l)}
\odot
\nabla_{W_{Q}^{(l)}}\mathcal F_a(M_{\mathrm{base}},x)
\bigr)
[:,\mathcal I_h]
\Bigr)
}{
G(x)
}
\right],
\label{eq:app_head_attr_q}
\\
\mathrm{Attr}_{O}(l,h) = \nonumber \\
\mathbb E_{x\sim\mathcal D_{\mathrm{inst}}}
\left[
\frac{
\sum
\Bigl(
\bigl(
\Delta W_{O}^{(l)}
\odot
\nabla_{W_{O}^{(l)}}\mathcal F_a(M_{\mathrm{base}},x)
\bigr)
[\mathcal I_h,:]
\Bigr)
}{
G(x)
}
\right].
\label{eq:app_head_attr_o}
\end{gather}
Here, $\odot$ denotes elementwise multiplication, and $\sum(\cdot)$ sums all entries in the selected slice. This aggregation preserves the component granularity established in Appendix~\ref{app:weight_patching_details}: only the head-specific query and output interfaces contribute to the head-level WAP score.

If activation-side head-level screening is also needed, an analogous aggregation can be defined by treating the head write-back activation $\Delta \mathbf Z_{\mathrm{attn}}^{(l,h)}$ as the intervention object and applying Eq.~\eqref{eq:app_aap_score} at the head level. In practice, however, the main screening emphasis in this paper is on the weight-side score in Eq.~\eqref{eq:app_head_attr_sum}.

\paragraph{Neuron-level aggregation.}
For an MLP neuron $N^{(l,j)}$, the parameter slice contains the gate column, up-projection column, and down-projection row. The neuron-level first-order score is therefore aggregated over these three parts:
\begin{equation}
\mathrm{Attr}_{\mathrm{wt}}(N^{(l,j)})
=
\mathrm{Attr}_{\mathrm{gate}}(l,j)
+
\mathrm{Attr}_{\mathrm{up}}(l,j)
+
\mathrm{Attr}_{\mathrm{down}}(l,j),
\label{eq:app_neuron_attr_sum}
\end{equation}
with
\begin{gather}
\mathrm{Attr}_{\mathrm{gate}}(l,j) = \nonumber \\
\mathbb E_{x\sim\mathcal D_{\mathrm{inst}}}
\left[
\frac{
\sum
\Bigl(
\bigl(
\Delta W_{\mathrm{gate}}^{(l)}
\odot
\nabla_{W_{\mathrm{gate}}^{(l)}}\mathcal F_a(M_{\mathrm{base}},x)
\bigr)
[:,j]
\Bigr)
}{
G(x)
}
\right],
\label{eq:app_neuron_attr_gate}
\\
\mathrm{Attr}_{\mathrm{up}}(l,j) = \nonumber \\
\mathbb E_{x\sim\mathcal D_{\mathrm{inst}}}
\left[
\frac{
\sum
\Bigl(
\bigl(
\Delta W_{\mathrm{up}}^{(l)}
\odot
\nabla_{W_{\mathrm{up}}^{(l)}}\mathcal F_a(M_{\mathrm{base}},x)
\bigr)
[:,j]
\Bigr)
}{
G(x)
}
\right],
\label{eq:app_neuron_attr_up}
\\
\mathrm{Attr}_{\mathrm{down}}(l,j) = \nonumber \\
\mathbb E_{x\sim\mathcal D_{\mathrm{inst}}}
\left[
\frac{
\sum
\Bigl(
\bigl(
\Delta W_{\mathrm{down}}^{(l)}
\odot
\nabla_{W_{\mathrm{down}}^{(l)}}\mathcal F_a(M_{\mathrm{base}},x)
\bigr)
[j,:]
\Bigr)
}{
G(x)
}
\right].
\label{eq:app_neuron_attr_down}
\end{gather}
This aggregation reflects the fact that a SwiGLU neuron is not a single scalar parameter but a structured functional unit whose feature detection, gating, and write-back direction are jointly responsible for its contribution to the control representation.

\paragraph{Full-model screening and exact validation.}
The first-order scores above are used to rank all candidate heads and neurons before exact intervention. Let
\begin{equation}
\mathcal C_{\mathrm{top}}^{\mathrm{wt}}
=
\operatorname{TopK}(\mathcal C,\mathrm{Attr}_{\mathrm{wt}},K_{\mathrm{wt}})
\label{eq:app_topk_wt}
\end{equation}
denote the top-ranked components under weight-side screening. If activation-side screening is also performed, define
\begin{equation}
\mathcal C_{\mathrm{top}}^{\mathrm{act}}
=
\operatorname{TopK}(\mathcal C,\mathrm{Attr}_{\mathrm{act}},K_{\mathrm{act}})
\label{eq:app_topk_act}
\end{equation}
Exact Weight Patching is then applied only to components in $\mathcal C_{\mathrm{top}}^{\mathrm{wt}}$, and exact activation-side intervention is applied only to components in $\mathcal C_{\mathrm{top}}^{\mathrm{act}}$ when needed for hierarchy recovery. Thus, the overall pipeline separates screening from validation:
\begin{align}
\mathrm{Attr}_{\mathrm{wt}}(c)
&\Longrightarrow
\text{ranking over all } c\in\mathcal C,
\\
E_w(c)
&\Longrightarrow
\text{exact source-level validation} \nonumber \\
&\qquad \text{on top-ranked candidates.}
\end{align}
This two-stage design is especially important for neuron-scale localization, where exhaustive exact Weight Patching is prohibitively expensive.

\paragraph{Role of the approximation.}
The first-order approximation is not intended to replace exact intervention as the source of causal evidence. Its role is methodological: it preserves the distinction between activation-side and parameter-side scoring while making full-model screening computationally feasible. In the empirical pipeline, the approximation is therefore judged by two criteria only: whether it retains the components most likely to matter under exact validation, and whether it reduces runtime enough to make fine-grained head- and neuron-level localization practical.

\paragraph{Summary.}
Taken together, Eqs.~\eqref{eq:app_aap_score}--\eqref{eq:app_topk_act} define a unified first-order screening layer for the proposed framework. Activation-side approximation highlights components whose cross-model state differences most strongly support anchor restoration, whereas weight-side approximation highlights components whose post-training parameter changes most strongly support source-level recovery. Head-level and neuron-level aggregation ensure that the approximation remains aligned with the same mechanistic units used by exact Weight Patching in the main text.

 \subsection{Hierarchy Recovery, Supplier--Target Tracing, and Source Validation}
\label{app:hierarchy_recovery}

This subsection gives the full hierarchy-recovery procedure used to connect source carriers to aggregation, routing, and downstream execution modules. The main text states the high-level functional roles and introduces the tracing intuition; here we provide the corresponding staged utilities, exact activation-side recovery criterion, supplier--target tracing rule, and source-validation condition in a unified form. Importantly, this stage does not impose a fixed Activation-Patching-to-Weight-Patching pipeline. Rather, activation-side and parameter-side analyses remain standalone views under the same anchor criterion, while their joint use provides a more structured mechanistic account.

\paragraph{Why hierarchy recovery is needed.}
Weight Patching identifies components whose own specialized parameters are sufficient to restore the target control representation, but this source-level evidence alone does not reconstruct the full mechanism by which instruction-conditioned control is formed, routed, and ultimately executed. In particular, a component can be highly important in activation space because it acts as a aggregation bottleneck or routing interface, even when the capability is not primarily written in that component's own parameters. We therefore separate three functional roles:
\begin{itemize}
    \item \textbf{source carriers}, i.e., components whose specialized parameters causally reintroduce the target control signal;
    \item \textbf{aggregation or routing modules}, i.e., components whose activations strongly restore or relay the control signal during inference;
    \item \textbf{downstream execution modules}, i.e., components whose intervention most directly restores output behavior relative to the specialized model.
\end{itemize}
This separation turns a flat importance ranking into a staged mechanistic account.

\paragraph{Staged utilities for control formation and execution.}
To distinguish control formation from output realization, two complementary utilities are used. For the control-formation stage, the utility is the anchor utility introduced in the main text:
\begin{equation}
\mathcal F_{\mathrm{up}}(M,x)
=
\mathcal F_a(M,x).
\label{eq:app_fup}
\end{equation}
This quantity measures whether the instruction-conditioned control representation has been formed at the anchor. For the execution stage, a behavioral utility is defined in output space by comparing the model under evaluation against the specialized model:
\begin{equation}
\mathcal F_{\mathrm{down}}(M,x,y)
=
- D_{\mathrm{KL}}\!\left(
p_{\mathrm{sft}}(\cdot\mid x,y_{<t})
\,\|\,
p_M(\cdot\mid x,y_{<t})
\right).
\label{eq:app_fdown}
\end{equation}
This readout is used in the main text because downstream
execution modules are most directly characterized by how
an intervention changes the distribution of the immediately
next generated token.

For completeness, the execution stage can also be evaluated
with a projection-based readout that mirrors the upstream
anchor utility, except that the readout is taken at the final
layer rather than the anchor layer. Let
\begin{equation}
z_{\mathrm{out}}(M, x, y_{<t})
\equiv
z^{(t)}_{L}(x, y_{<t} \mid M)
\end{equation}
denote the final-layer residual state at the position used to
predict the immediately next token, and let $v_{\mathrm{out}}$
denote the task direction extracted by the same contrastive
construction as in Eq.~(11), but at layer $L$. The
projection-based downstream utility is then
\begin{equation}
F^{\mathrm{proj}}_{\mathrm{down}}(M, x, y)
=
\mathrm{sim}
\!\left(
z_{\mathrm{out}}(M, x, y_{<t}),
v_{\mathrm{out}}
\right).
\label{eq:downstream_proj_utility}
\end{equation}
This variant reuses the same vector-projection logic as the
upstream metric, but moves the readout to the final layer.
In practice, the projection-based and KL-based readouts
yield qualitatively similar localization patterns. Equation
\eqref{eq:downstream_utility} is used only for downstream
execution analysis; it is not the source-level objective of
the method. Whenever the projection-based variant is used,
the downstream restoration effect in Eq.~(64) is defined
analogously by substituting
$F^{\mathrm{proj}}_{\mathrm{down}}$ for $F_{\mathrm{down}}$.

\paragraph{Activation-side recovery of aggregation and routing modules.}
Given the exact activation-side restoration effect
\begin{equation}
E_a(c)
=
\mathbb E_{x\sim\mathcal D_{\mathrm{inst}}}
\left[
\frac{
\mathcal F_a(\widetilde M_{\mathrm{base}}^{(c)},x)
-
\mathcal F_a(M_{\mathrm{base}},x)
}{
G(x)
}
\right],
\label{eq:app_hierarchy_ea}
\end{equation}
components with large $E_a(c)$, or large activation-side first-order scores $\mathrm{Attr}_{\mathrm{act}}(c)$, are treated as candidates for aggregation and routing modules. These modules are not assumed to be source carriers. Rather, they mark locations where instruction-conditioned information becomes functionally important, is integrated from earlier suppliers, or is redirected toward later execution stages. Operationally, let
\begin{equation}
\mathcal S_{\mathrm{conv}}
=
\left\{
c\in\mathcal C:
E_a(c)>\tau_a
\right\}
\label{eq:app_conv_set}
\end{equation}
denote the set of activation-side aggregation candidates, where $\tau_a$ is a threshold chosen from exact or screened activation-side validation. In practice, top-ranked components under $E_a(c)$ or $\mathrm{Attr}_{\mathrm{act}}(c)$ can also be used when a fixed-cardinality selection is preferred.

\paragraph{Target-side need and supplier-side support.}
To avoid ambiguity with the global anchor-relative upstream/downstream partition used in the main text, local component-to-component tracing is written here in supplier--target form. For a chosen target component $c_{\mathrm{tar}}$, define
\begin{equation}
\mathbf g_{\mathrm{need}}(c_{\mathrm{tar}})
=
\nabla_{\Psi_{c_{\mathrm{tar}}}(\mathbf Z)}
\mathcal F_{\mathrm{up}}(M_{\mathrm{base}},x)
\label{eq:app_gneed}
\end{equation}
as the target-side need direction, where $\Psi_{c_{\mathrm{tar}}}(\mathbf Z)$ extracts the residual-stream input to $c_{\mathrm{tar}}$. This direction describes which perturbation at the target input would most effectively increase the anchor-side control utility.

For a candidate supplier component $c_{\mathrm{sup}}$, define its weight-side support to $c_{\mathrm{tar}}$ by
\begin{equation}
s_{\mathrm{wt}}(c_{\mathrm{tar}},c_{\mathrm{sup}})
=
\sum_{p\in\Theta^{(c_{\mathrm{sup}})}}
\Delta\theta_p\,
\frac{
\partial \mathcal F_{\mathrm{need}}^{(c_{\mathrm{tar}})}(M_{\mathrm{base}},x)
}{
\partial \theta_p
},
\label{eq:app_vsupply}
\end{equation}
where
\begin{equation}
\mathcal F_{\mathrm{need}}^{(c_{\mathrm{tar}})}(M,x)
=
\left\langle
\Psi_{c_{\mathrm{tar}}}(\mathbf Z(M,x)),
\mathbf g_{\mathrm{need}}(c_{\mathrm{tar}})
\right\rangle.
\label{eq:app_fneed}
\end{equation}
This quantity measures how strongly the post-training parameter change on $c_{\mathrm{sup}}$ supports the target-side need of $c_{\mathrm{tar}}$.

\paragraph{Functional link strength.}
The functional link between a target component and a candidate supplier is measured by
\begin{equation}
E_{\mathrm{link}}(c_{\mathrm{tar}},c_{\mathrm{sup}})
=
\mathbb E_{x\sim\mathcal D_{\mathrm{inst}}}
\left[
\frac{
s_{\mathrm{wt}}(c_{\mathrm{tar}},c_{\mathrm{sup}})
}{
G(x)
}
\right].
\label{eq:app_elink}
\end{equation}
A large value of $E_{\mathrm{link}}(c_{\mathrm{tar}},c_{\mathrm{sup}})$ indicates that the post-training parameter change on $c_{\mathrm{sup}}$ is well aligned with increasing the need utility of the target component $c_{\mathrm{tar}}$. This criterion is used to distinguish functionally relevant suppliers from unrelated components whose parameters changed during post-training but do not support the specific target under study.

Given a target component $c_{\mathrm{tar}}$, one can therefore define a candidate supplier set
\begin{equation}
\mathcal S_{\mathrm{sup}}(c_{\mathrm{tar}})
=
\left\{
c\in\mathcal C:
E_{\mathrm{link}}(c_{\mathrm{tar}},c)>\tau_{\mathrm{link}}
\right\},
\label{eq:app_upstream_candidates}
\end{equation}
where $\tau_{\mathrm{link}}$ is a link-strength threshold, or equivalently use the top-ranked suppliers under $E_{\mathrm{link}}$.

\paragraph{Source validation.}
A strong supplier--target link does not by itself imply that the supplier component is a true source carrier. Some components may simply relay information that was formed even earlier, rather than storing the relevant post-training update in their own parameters. To distinguish genuine source carriers from relay nodes, candidate supplier components are further validated using exact Weight Patching. Specifically, a component is identified as a source carrier only if it satisfies both strong parameter-side restoration and strong functional linkage:
\begin{equation}
\mathrm{IsSource}(c_{\mathrm{sup}})
\iff
E_w(c_{\mathrm{sup}})>\tau_w
\;\land\;
E_{\mathrm{link}}(c_{\mathrm{tar}},c_{\mathrm{sup}})>\tau_{\mathrm{link}}.
\label{eq:app_is_source}
\end{equation}
The first condition asks whether the specialized parameters of $c_{\mathrm{sup}}$ are themselves sufficient to restore the missing control representation. The second condition asks whether the component participates in the current supplier--target chain. Only when both conditions hold is the component treated as a source-level carrier in the recovered hierarchy. Accordingly, for a given target component $c_{\mathrm{tar}}$, the validated source set is
\begin{equation}
\mathcal S_{\mathrm{src}}(c_{\mathrm{tar}})
=
\left\{
c\in\mathcal C:
E_w(c)>\tau_w
\;\land\;
E_{\mathrm{link}}(c_{\mathrm{tar}},c)>\tau_{\mathrm{link}}
\right\}.
\label{eq:app_source_set}
\end{equation}

\paragraph{Downstream execution modules.}
The final stage of the hierarchy concerns components that most directly implement behavior at the output level. These are identified by evaluating interventions under the downstream utility in Eq.~\eqref{eq:app_fdown}. For notational compactness, let $\mathcal I_c$ denote the intervention applied to component $c$ in the chosen evaluation protocol, which may be an exact activation-side intervention or an exact parameter-side intervention depending on the analysis. The corresponding downstream restoration effect is
\begin{gather}
E_{\mathrm{down}}(c) = \nonumber \\
\mathbb E_{(x,y)\sim\mathcal D_{\mathrm{eval}}}
\left[
\frac{
\mathcal F_{\mathrm{down}}(\mathcal I_c(M_{\mathrm{base}}),x,y)
-
\mathcal F_{\mathrm{down}}(M_{\mathrm{base}},x,y)
}{
\mathcal F_{\mathrm{down}}(M_{\mathrm{sft}},x,y)
-
\mathcal F_{\mathrm{down}}(M_{\mathrm{base}},x,y)
}
\right].
\label{eq:app_edown}
\end{gather}
Downstream execution modules are then selected as components with strong output-space restoration, for example
\begin{equation}
\mathcal S_{\mathrm{exe}}
=
\left\{
c\in\mathcal C:
E_{\mathrm{down}}(c)>\tau_{\mathrm{down}}
\right\}.
\label{eq:app_execution_set}
\end{equation}
This stage is used to characterize the final realization of behavior rather than to define the core source-localization objective.

\paragraph{Recovered hierarchical circuit.}
Combining the three stages above yields a directed hierarchical circuit representation rather than a flat component list. Let
\begin{equation}
\mathcal S_{\mathrm{src}}
=
\bigcup_{c_{\mathrm{tar}}\in\mathcal S_{\mathrm{conv}}}
\mathcal S_{\mathrm{src}}(c_{\mathrm{tar}})
\label{eq:app_all_sources}
\end{equation}
collect all validated source carriers linked to at least one traced target component. The recovered circuit set is then
\begin{equation}
\mathcal S_{\mathrm{circuit}}
=
\mathcal S_{\mathrm{src}}
\cup
\mathcal S_{\mathrm{conv}}
\cup
\mathcal S_{\mathrm{exe}}.
\label{eq:app_circuit_union}
\end{equation}
The resulting structure corresponds to a directional functional chain that begins at source-level carriers, passes through aggregation and routing interfaces, and culminates in downstream execution modules.

\paragraph{Practical recovery procedure.}
The full hierarchy-recovery pipeline can be summarized as follows.
\begin{enumerate}
    \item Select target modules for tracing, either from activation-side evidence, from Weight Patching evidence, or from prior mechanistic hypotheses.
    \item For each target component $c_{\mathrm{tar}}$, compute the target-side need vector $\mathbf g_{\mathrm{need}}(c_{\mathrm{tar}})$.
    \item For each candidate supplier component $c_{\mathrm{sup}}$, compute its weight-side support $s_{\mathrm{wt}}(c_{\mathrm{tar}},c_{\mathrm{sup}})$ and the corresponding link strength $E_{\mathrm{link}}(c_{\mathrm{tar}},c_{\mathrm{sup}})$.
    \item Retain only those supplier candidates that also satisfy the parameter-side source criterion in Eq.~\eqref{eq:app_is_source}.
    \item Use downstream behavioral recovery under $\mathcal F_{\mathrm{down}}$ to characterize the execution-stage modules $\mathcal S_{\mathrm{exe}}$ when needed.
\end{enumerate}
This procedure yields a hierarchy rather than a single global ranking: source modules are supported by exact Weight Patching, aggregation and routing modules are supported by activation-side restoration, and downstream execution modules are supported by output-level behavioral recovery.

\paragraph{Summary.}
Taken together, Eqs.~\eqref{eq:app_fup}--\eqref{eq:app_circuit_union} define a staged hierarchy-recovery procedure that integrates source-level parameter evidence, activation-side aggregation analysis, supplier--target tracing, and output-level execution analysis into a unified mechanistic account. The resulting representation supports both standalone comparison between activation-side and parameter-side localization and their joint use in a directional circuit hypothesis.

\subsection{Detailed WP-Guided Fusion Formulation and Implementation}
\label{app:fusion_details}

This subsection provides the full formulation of the exploratory WP-guided fusion extension. The main text gives the core component-wise weighting rule, but the implementation of the fused model is most transparent when written at the slice level and linked explicitly to the same head and neuron units used in Weight Patching. We therefore detail how exact or first-order component scores are converted into nonnegative expert weights, how those weights are normalized across experts at each component, and how the resulting coefficients are applied consistently to full head and neuron slices.

\paragraph{Shared-base multi-expert setting.}
Let $M_{\mathrm{base}}$ denote a shared pretrained base model, and let $\{M^{(k)}\}_{k=1}^{K}$ denote $K$ expert models obtained by finetuning the same base toward different capabilities. Their parameters are denoted by $\{\theta^{(k)}\}_{k=1}^{K}$, and the expert-specific parameter deltas relative to the common base are
\begin{equation}
\Delta\theta^{(k)}
=
\theta^{(k)}-\theta_{\mathrm{base}}.
\label{eq:app_fusion_delta}
\end{equation}
Because all experts are derived from the same base checkpoint, the differences among them can be expressed uniformly in terms of these component-wise parameter deltas.

\paragraph{Component-level expert scoring.}
For each expert model $M^{(k)}$ and each component $c\in\mathcal C$, we compute a component score $s^{(k)}(c)$ that reflects how strongly expert $k$ is supported on component $c$ as a carrier of the target capability. In the exact setting, this score can be instantiated by the Weight Patching effect:
\begin{equation}
s^{(k)}(c)
=
E_w^{(k)}(c),
\label{eq:app_fusion_exact_score}
\end{equation}
where $E_w^{(k)}(c)$ denotes the exact source-level restoration effect obtained by patching the slice of expert $k$ into the shared base model. In the efficient setting, the score can instead be approximated by the first-order weight attribution:
\begin{equation}
s^{(k)}(c)
=
\mathrm{Attr}_{\mathrm{wt}}^{(k)}(c).
\label{eq:app_fusion_fo_score}
\end{equation}
The formulation does not require the two choices to be used simultaneously; rather, either exact or first-order component scores can serve as the basis for the fusion rule depending on computational budget and granularity.

\paragraph{Nonnegative truncation.}
Since component scores can be negative, directly using them as fusion weights would allow suppressive or interfering updates to influence the merged model. To avoid this, only positive evidence is retained:
\begin{equation}
\bar s^{(k)}(c)
=
\max\!\bigl(s^{(k)}(c),0\bigr).
\label{eq:app_fusion_positive}
\end{equation}
This step preserves components for which expert $k$ receives positive support while assigning zero weight to experts whose updates on component $c$ are estimated to be harmful or misaligned with the target behavior.

\paragraph{Component-wise cross-expert normalization.}
For each component $c$, the retained nonnegative scores are normalized across experts to obtain the fusion weight of expert $k$ on that component:
\begin{equation}
\alpha^{(k)}(c)
=
\begin{cases}
\dfrac{\bar s^{(k)}(c)}{\sum_{m=1}^{K}\bar s^{(m)}(c)},
&
\sum_{m=1}^{K}\bar s^{(m)}(c)>0, \\[8pt]
\dfrac{1}{K},
&
\sum_{m=1}^{K}\bar s^{(m)}(c)=0.
\end{cases}
\label{eq:app_fusion_alpha}
\end{equation}
By construction, this yields
\begin{equation}
\sum_{k=1}^{K}\alpha^{(k)}(c)=1,
\qquad
\forall\, c\in\mathcal C.
\label{eq:app_fusion_simplex}
\end{equation}
Equation~\eqref{eq:app_fusion_alpha} has three useful limiting behaviors. When a component is strongly supported by a single expert, the weights approach a near one-hot allocation. When multiple experts have positive support on the same component, the rule reduces to local importance-weighted averaging. When no expert exhibits reliable positive evidence on that component, the rule falls back to uniform averaging.

\paragraph{Component-wise fused model.}
Let $\theta_{\mathrm{fuse}}$ denote the parameters of the fused model $M_{\mathrm{fuse}}$. For each component $c$, its fused parameter slice is defined by
\begin{equation}
\theta_{\mathrm{fuse}}^{(c)}
=
\theta_{\mathrm{base}}^{(c)}
+
\sum_{k=1}^{K}
\alpha^{(k)}(c)\,\Delta\theta^{(k,c)},
\label{eq:app_fusion_slice_delta}
\end{equation}
where
\begin{equation}
\Delta\theta^{(k,c)}
=
\theta^{(k,c)}-\theta_{\mathrm{base}}^{(c)}.
\label{eq:app_fusion_component_delta}
\end{equation}
Using Eq.~\eqref{eq:app_fusion_simplex}, Eq.~\eqref{eq:app_fusion_slice_delta} is equivalently written as
\begin{equation}
\theta_{\mathrm{fuse}}^{(c)}
=
\sum_{k=1}^{K}
\alpha^{(k)}(c)\,\theta^{(k,c)}.
\label{eq:app_fusion_slice_avg}
\end{equation}
Equation~\eqref{eq:app_fusion_slice_delta} emphasizes the interpretation of fusion as a component-wise combination of expert deltas around a common base, whereas Eq.~\eqref{eq:app_fusion_slice_avg} emphasizes its interpretation as a convex combination of expert slices.

\paragraph{Slice-level implementation.}
The fusion granularity is kept strictly consistent with the component interfaces used throughout Weight Patching. For an attention head $H^{(l,h)}$, the same coefficient $\alpha^{(k)}(H^{(l,h)})$ is applied to the full parameter slice defined in Eq.~\eqref{eq:app_head_slice}. For an MLP neuron $N^{(l,j)}$, the same coefficient $\alpha^{(k)}(N^{(l,j)})$ is applied to the full slice defined in Eq.~\eqref{eq:app_neuron_slice}. Thus, the merged model is constructed at the same head-level and neuron-level structural granularity used by the mechanistic analysis, rather than by unconstrained scalar-wise mixing across the entire parameter space.

\paragraph{Relation to uniform averaging.}
A useful reference case is obtained by setting
\begin{equation}
\alpha^{(k)}(c)=\frac{1}{K},
\qquad
\forall\, c\in\mathcal C,\; k=1,\dots,K.
\label{eq:app_fusion_uniform}
\end{equation}
Under this choice, the method degenerates to component-independent uniform averaging, which no longer distinguishes among experts according to their source-level support on different functional modules. The essential difference of the proposed fusion rule is therefore not merely that it averages parameters, but that it performs a component-aligned causal allocation of expert responsibility.

\paragraph{Interpretation.}
The component-wise coefficients in Eq.~\eqref{eq:app_fusion_alpha} make explicit the question of which expert should dominate which mechanistic unit. This differs from global averaging or globally shared task-vector combinations, where the same mixing rule is applied uniformly across the full parameter space. In the present formulation, a head or neuron is preserved from an expert model precisely when that expert receives stronger support on the corresponding localized capability carrier. As a result, the fused model is designed to better preserve sparse source-level interfaces and to reduce destructive interference among experts at functionally distinct components.

\paragraph{Practical construction.}
The full fusion procedure can be summarized as follows.
\begin{enumerate}
    \item For each expert model $M^{(k)}$ and each component $c\in\mathcal C$, compute a component score $s^{(k)}(c)$ using either exact Weight Patching or first-order weight attribution.
    \item Apply nonnegative truncation via Eq.~\eqref{eq:app_fusion_positive}.
    \item Normalize the retained scores across experts for each component using Eq.~\eqref{eq:app_fusion_alpha}.
    \item Construct each fused component slice using Eq.~\eqref{eq:app_fusion_slice_delta} or, equivalently, Eq.~\eqref{eq:app_fusion_slice_avg}.
    \item Assemble all fused slices into the final fused model $M_{\mathrm{fuse}}$.
\end{enumerate}

\paragraph{Summary.}
Taken together, Eqs.~\eqref{eq:app_fusion_delta}--\eqref{eq:app_fusion_uniform} define a mechanism-aware multi-expert fusion rule aligned with the same interpretable units used throughout Weight Patching. The resulting fused model is not formed by a globally uniform parameter average, but by a structured component-wise allocation rule that reuses source-level importance signals as practical units for low-cost and interpretable multi-capability composition.

\section{Reproducibility and Experimental Setup}
\label{app:repro}

\subsection{Models, Checkpoints, and Architectures}
\label{app:repro_models}

Our mechanistic analysis is conducted on paired base/instruct models from the Llama family, including Llama-3.2-3B, Llama-3.1-8B, and Llama-2-13B. In each case, we analyze a base model together with its instruction-tuned counterpart under the same architecture, so that behavioral differences can be attributed to post-training parameter changes rather than architectural variation. We include 3B, 8B, and 13B scales to test whether the main observations of our framework remain consistent across model sizes and model generations. For the model-fusion experiments, we follow the common setup in recent merging work and use Llama-2-13B-based expert models as the shared backbone, which enables direct comparison to standard instruction, math, and code specialists~\cite{wizardlm,wizardmath,codealpaca,nobari2025aim}.

These models also differ in attention parameterization, which affects the implementation of head-level parameter interventions. Llama-2-13B uses the same number of KV heads as attention heads, making head-level slicing more direct. By contrast, Llama-3.1-8B and Llama-3.2-3B adopt grouped-query attention, where KV heads are shared across multiple query heads. Accordingly, our head-level Weight Patching implementation follows the underlying architecture and uses architecture-consistent parameter slicing for each model family. Table~\ref{tab:appendix_model_arch} summarizes the key architectural statistics used throughout the experiments, including the number of layers, hidden size, attention heads, KV heads, MLP width, and vocabulary size.

\begin{table}[t]
\centering
\caption{Architectural statistics of the model families used in our experiments.}
\label{tab:appendix_model_arch}
\setlength{\tabcolsep}{3.5pt}
\renewcommand{\arraystretch}{1.1}
\resizebox{\columnwidth}{!}{%
\begin{tabular}{lcccccc}
\toprule
Model & Layers & Hidden Dim & Attention & KV Heads & MLP Width & Vocab Size \\
 & ($L$) & ($d_{\text{model}}$) & Heads &  &  &  \\
\midrule
Llama-2-13B  & 40 & 5120 & 40 & 40 & 13824 & 32000 \\
Llama-3.1-8B & 32 & 4096 & 32 & 8  & 14336 & 128256 \\
Llama-3.2-3B & 28 & 3072 & 24 & 8  & 8192  & 128256 \\
\bottomrule
\end{tabular}%
}
\end{table}
 
\subsection{Tasks, Data Sources, and Evaluation Protocol}
\label{app:repro_tasks}

Our primary interpretability experiments are built on six instruction-following tasks drawn from IFEval~\cite{IFEval_zhou2023instruction}. Rather than aiming to exhaust the full benchmark, we focus on a subset of tasks for which stable intermediate steering vectors can be reliably obtained under the current vector-extraction framework. In this sense, the selected tasks serve as an operational subset for mechanistic analysis: they provide sufficiently stable behavioral interfaces for localization, ablation, and restoration, while still covering diverse forms of procedural constraints, including punctuation control, title formatting, structural organization, quotation boundaries, highlighted numbered sections, and global capitalization. As discussed in the main text, our goal is not to design a stronger vector-extraction method, but to use the strongest available method as a stable interface for generative-behavior analysis~\cite{task_vector_iclr2025}.

For vector extraction, data usage, and preprocessing, we follow \cite{task_vector_iclr2025}. For final task evaluation on IFEval, we also follow the corresponding official evaluation protocol, and therefore do not repeat the low-level implementation details here. In addition, for the model-fusion experiments, we follow the task-specific evaluation protocols used in prior model-merging work, especially AIM and its referenced baselines~\cite{nobari2025aim}. Throughout the paper, this yields a consistent division between benchmark-defined end-task evaluation and our own mechanistic localization pipeline.

\subsection{Data Usage, Split Strategy, and Vector-Extraction Protocol}
\label{app:repro_split}

For interpretability analysis, the data used to extract steering vectors and the data used for subsequent steering and localization are not shared. In the default setup, we split the available data into two halves: one half is used for steering-vector extraction, and the other half is used for downstream interpretability analysis. The forward-pass data used to produce the heatmaps in the main text follows this split strategy. We also ran an alternative setting in which vector extraction and subsequent interpretability analysis use the same data, and found that the resulting localization patterns are highly similar. This indicates that the main conclusions do not depend sensitively on the particular split choice.

The task-wise sample counts used in vector extraction are 434 for \textit{english\_capital}, 479 for \textit{multiple\_sections}, 464 for \textit{number\_highlighted\_sections}, 483 for \textit{title}, 406 for \textit{no\_comma}, and 426 for \textit{quotation}. In practice, we further observe that the localization heatmaps stabilize quickly: for the main interpretability analyses, using roughly 50 samples is already sufficient for the heatmap pattern to stop changing in any meaningful way. Increasing the sample size beyond this point has little effect on the qualitative conclusions. This empirical observation supports both the efficiency and the stability of the proposed localization framework. Since our focus is not on improving vector extraction itself, all vector-related procedures are kept fully aligned with \cite{task_vector_iclr2025}, and we leave the detailed discussion of vector-construction methodology to that work.

\begin{figure*}[!ht]
  \centering
  \includegraphics[width=0.95\linewidth]{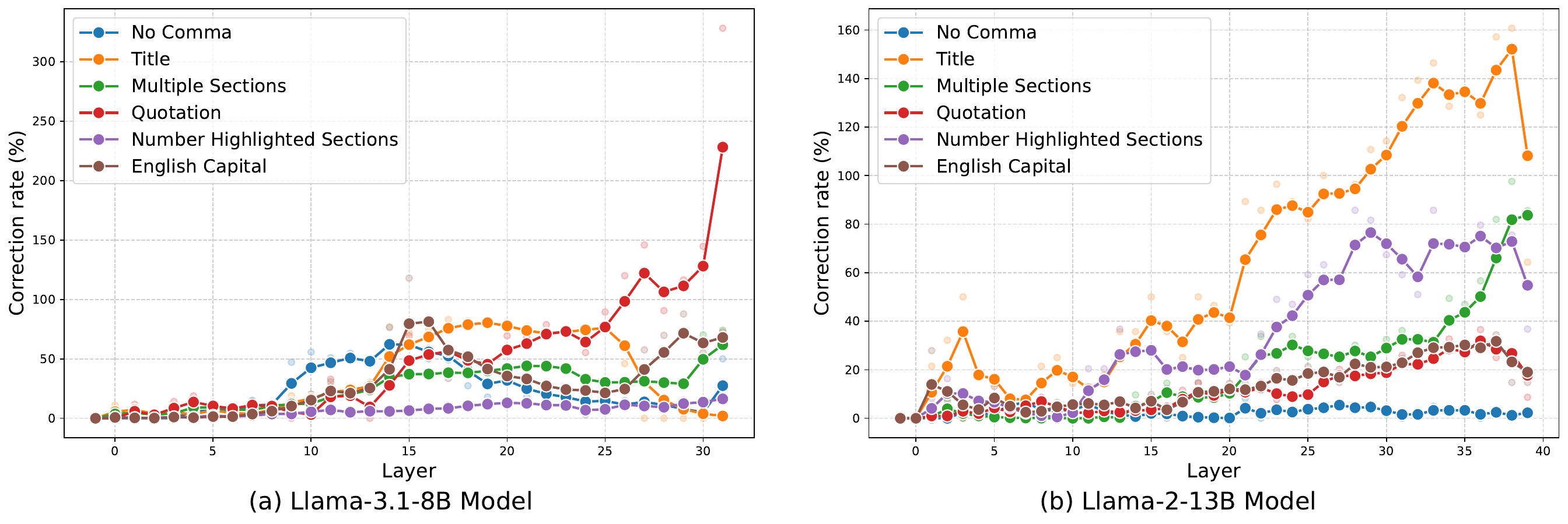}
  \vspace{-10pt}
  \caption{\textbf{Layer-wise steering correction rates for Llama-3.1-8B and Llama-2-13B.} This figure complements Fig.~\ref{fig:steering_plot} in the main text by showing the steering effects of task vectors across all layers for the 8B and 13B models on six representative IFEval tasks. Similar to the 3B model, optimal instruction-following recovery consistently peaks in the mid-to-late layers across both larger models. This demonstrates the cross-model stability of the vector-anchor interface and confirms that instruction-conditioned control representations are reliably formed after early token-level processing.}
  \label{sup:steering_plot}
\end{figure*}

\begin{figure}[!ht]
    \centering
    \includegraphics[width=0.95\linewidth]{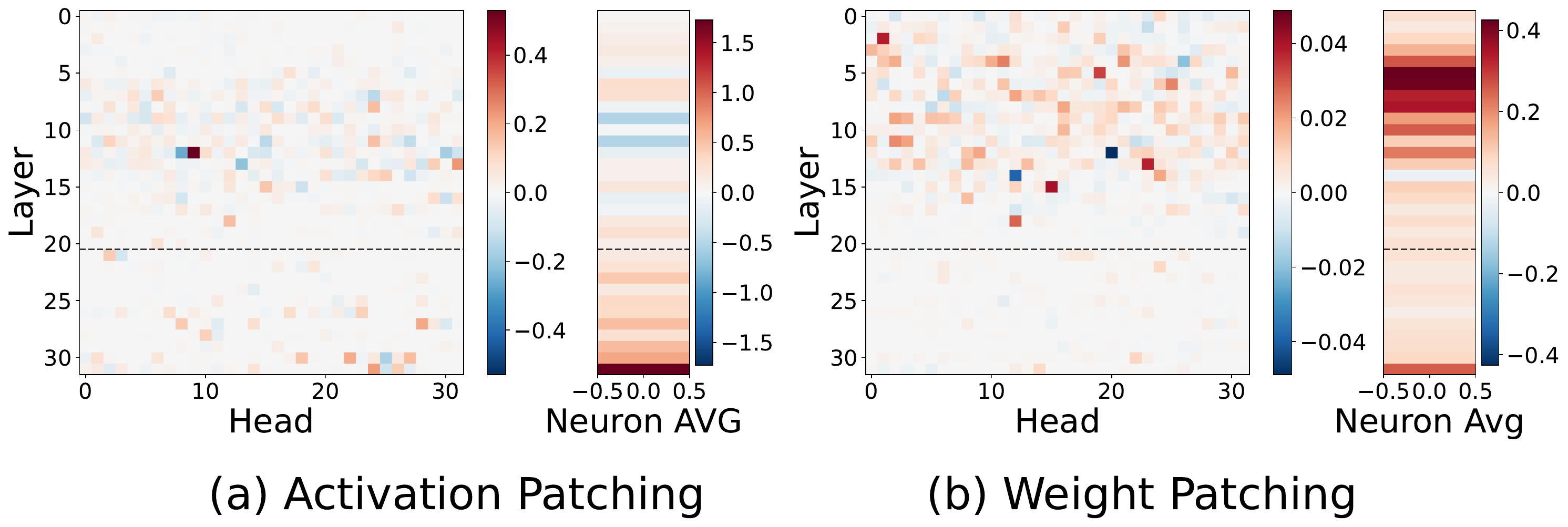}
    \vspace{-10pt}
  
    \caption{Heatmaps of component importance under activation patching and Weight Patching on the English Capital task on \textbf{Llama-3.1-8B}. }
    \label{sup:heatmap-8b}
\end{figure}

\begin{figure}[!th]
    \centering
    \includegraphics[width=0.95\linewidth]{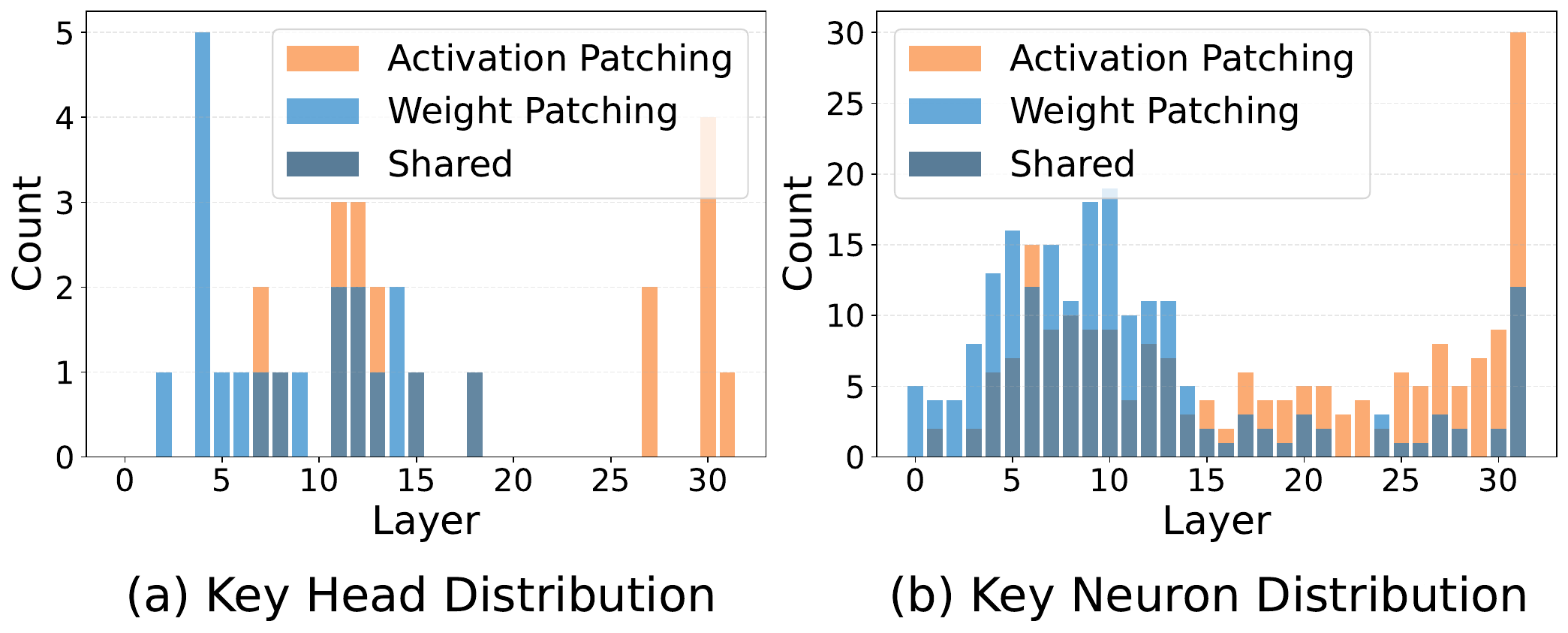}
    \vspace{-10pt}
    
    \caption{ Layer distributions and overlaps of top-ranked components identified by activation patching and weight patching on \textbf{Llama-3.1-8B} .}
    \label{sup:distribution-8b}
\end{figure}

\subsection{Hyperparameters and Implementation Details}
\label{app:repro_hparams}
These settings are used consistently across the main interpretability experiments unless otherwise noted.
Unless otherwise specified, vector extraction follows \cite{task_vector_iclr2025} with a batch size of 10. For steering, we use projection-based intervention, with batch sizes of 50 for the 3B model, 20 for the 8B model, and 20 for the 13B model. The default analysis layer is selected according to steering performance; in our main setup, the default layer is 17 for 3B, 20 for 8B, and 27 for 13B. For patching, the batch size is 10 for 3B, 10 for 8B, and 1 for 13B. The implementation of parameter slicing is kept consistent with the underlying attention architecture of each model family.

\begin{figure}[!ht]
    \centering
    \includegraphics[width=0.95\linewidth]{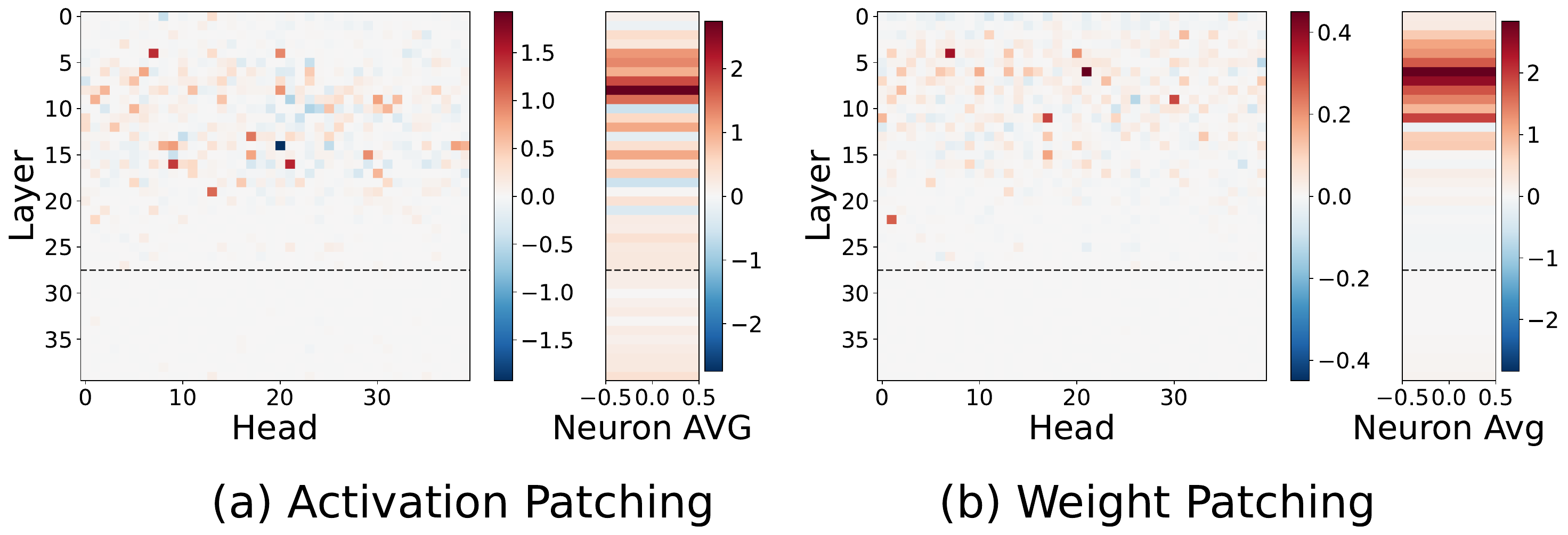}
    \vspace{-10pt}
  
    \caption{ Heatmaps of component importance under activation patching and Weight Patching on the English Capital task on \textbf{Llama-2-13B}. }
    \label{sup:heatmap-13b}
\end{figure}

\begin{figure}[!th]
    \centering
    \includegraphics[width=0.95\linewidth]{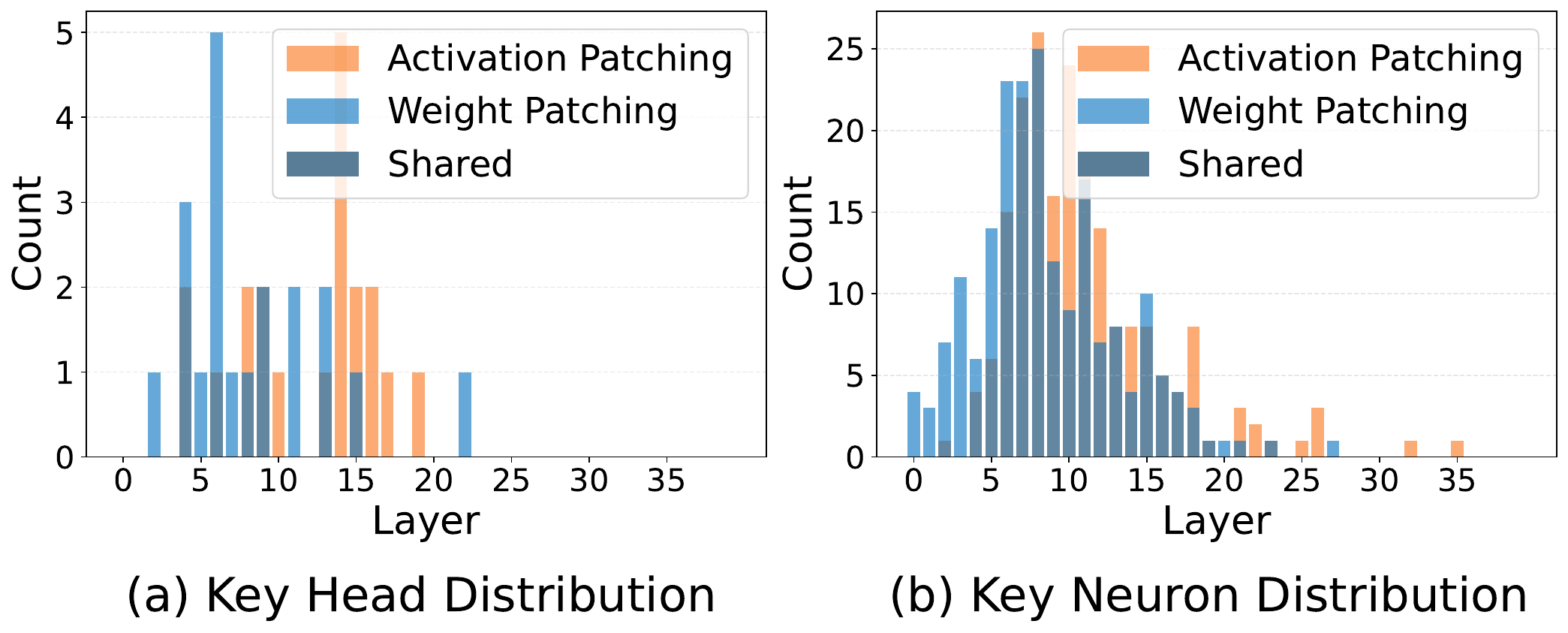}
    \vspace{-10pt}
    
    \caption{ Layer distributions and overlaps of top-ranked components identified by activation patching and weight patching on \textbf{Llama-2-13B}. }
    \label{sup:distribution-13b}
\end{figure}

For knockout and restore experiments, we use a batch size of 10. In the restore radar plots, the restored fraction for heads and neurons increases from 1\% to 5\% with a step size of 1\%. In the joint AP+WP restoration setting, the restored fraction increases from 2\% to 10\% with a step size of 2\%. In the attention-ratio analysis, we use the top-10 heads when comparing the average attention mass assigned to instruction and non-instruction regions. In the WP--WtP efficiency comparison, intersections are computed using Top-20 heads and Top-10 MLP layers. In the head-to-neuron analysis, we first select the Top-20 heads and then retrieve the Top-200 upstream neurons for each head, where neuron rankings are computed by absolute value.

\begin{table*}[t]
    \centering
    \caption{Vocabulary projection results of task vectors from shallow to deep layers across six tasks in the \textbf{Llama-3.2-3B} model.}
    \vspace{-8pt}
    \resizebox{1\textwidth}{!}{%
    \begin{tabular}{l c l} 
        \specialrule{1.2pt}{0pt}{0pt}
        \rowcolor{mygray}
        \textbf{Task} & \textbf{Layer} & \textbf{Projection (Top-10 vocabulary)} \\
        \hline
        \multirow{5}{*}{No Comma} & 3 & <\text{wear}>, <\text{adam}>, <\text{Dipl}>, <\text{cca}>, <\text{McCorm}>, <\text{<U+FFFD>}>, <\text{Synd}>, <\text{least}>, <\text{dex}>, <\text{ook}> \\
        & 10 & <\text{-horizontal}>, <\text{ippi}>, <\text{Artem}>, <\text{<U+FFFD>}>, <\text{Cap}>, <\text{naked}>, <\text{urs}>, <\text{<U+FFFD>}>, <\text{imuth}>, <\text{eni}> \\
        & 15 & <\text{gov}>, <\text{tv}>, <\text{repeat}>, <\text{nge}>, <\text{\begin{CJK*}{UTF8}{gbsn}尖\end{CJK*}}>, <\text{alien}>, <\text{enez}>, <\text{U+0626}>, <\text{dont}>, <\text{\foreignlanguage{russian}{око}}> \\
        & 20 & <\text{moreover}>, <\text{$\tau\acute{\iota}$}>, <\text{('.')}>, <\text{hàng}>, <\text{doesnt}>, <\text{furthermore}>, <\text{dont}>, <\text{Dont}>, <\text{auen}>, <\text{fine}> \\
        & 25 & <\text{youre}>, <\text{doesnt}>, <\text{its}>, <\text{theres}>, <\text{its}>, <\text{Dont}>, <\text{Lets}>, <\text{Its}>, <\text{dont}>, <\text{hey}> \\
        \specialrule{0em}{2pt}{2pt} \\
        \multirow{5}{*}{Title} & 3 & <\text{ofi}>, <\text{Arch}>, <\text{\begin{CJK*}{UTF8}{gbsn}卒\end{CJK*}}>, <\text{\begin{CJK*}{UTF8}{min}ルト\end{CJK*}}>, <\text{onna}>, <\text{phin}>, <\text{positioning}>, <\text{\begin{CJK*}{UTF8}{mj}탈\end{CJK*}}>, <\text{Princip}>, <\text{ghan}> \\
        & 10 & <\text{Frem}>, <\text{norm}>, <\text{aket}>, <\text{cken}>, <\text{Seek}>, <\text{Matcher}>, <\text{addons}>, <\text{rhe}>, <\text{onen}>, <\text{orch}> \\
        & 15 & <\text{cent}>, <\text{ardu}>, <\text{getString}>, <\text{\begin{CJK*}{UTF8}{gbsn}『\end{CJK*}}>, <\text{Cent}>, <\text{ulla}>, <\text{<U+FFFD>}>, <\text{installations}>, <\text{mos}>, <\text{Tops}> \\
        & 20 & <\text{either}>, <\text{<<}>, <\text{Hockey}>, <\text{also}>, <\text{either}>, <\text{<U+FFFD>}>, <\text{also}>, <\text{cartel}>, <\text{<U+FFFD>}>, <\text{Either}> \\
        & 25 & <\text{<<}>, <\text{<<}>, <\text{)<<}>, <\text{]<<}>, <\text{"<<}>, <\text{"<<}>, <\text{<}>, <\text{<<-}>, <\text{<}>, <\text{)<}> \\
        \specialrule{0em}{2pt}{2pt} \\
        \multirow{5}{*}{Multiple Sections} & 3 & <\text{ofi}>, <\text{importantly}>, <\text{|x}>, <\text{adam}>, <\text{oleans}>, <\text{overall}>, <\text{Overall}>, <\text{gems}>, <\text{Heavy}>, <\text{arken}> \\
        & 10 & <\text{sher}>, <\text{thus}>, <\text{thus}>, <\text{Matcher}>, <\text{imper}>, <\text{\begin{CJK*}{UTF8}{gbsn}空\end{CJK*}}>, <\text{DAY}>, <\text{anik}>, <\text{Hell}>, <\text{darkness}> \\
        & 15 & <\text{ensuing}>, <\text{BC}>, <\text{ulla}>, <\text{unker}>, <\text{Sel}>, <\text{ASS}>, <\text{trap}>, <\text{<U+FFFD>}>, <\text{<U+0E40U+0E02>}>, <\text{allon}> \\
        & 20 & <\text{section}>, <\text{**}>, <\text{sections}>, <\text{Section}>, <\text{\#\#\#}>, <\text{Section}>, <\text{section}>, <\text{**}>, <\text{\#\#\#}>, <\text{-section}> \\
        & 25 & <\text{SECTION}>, <\text{section}>, <\text{SECTION}>, <\text{[section}>, <\text{-section}>, <\text{(section}>, <\text{sect}>, <\text{section}>, <\text{.section}>, <\text{sectional}> \\
        \specialrule{0em}{2pt}{2pt} \\
        \multirow{5}{*}{Quotation} & 3 & <\text{Nope}>, <\text{cca}>, <\text{Arrange}>, <\text{positioned}>, <\text{rch}>, <\text{igned}>, <\text{ach}>, <\text{arken}>, <\text{Dipl}>, <\text{ernaut}> \\
        & 10 & <\text{Bye}>, <\text{rug}>, <\text{Matcher}>, <\text{wa}>, <\text{kir}>, <\text{cken}>, <\text{Pil}>, <\text{DISCLAIMER}>, <\text{oba}>, <\text{rud}> \\
        & 15 & <\text{Brace}>, <\text{midd}>, <\text{enclosing}>, <\text{surrounding}>, <\text{dipped}>, <\text{interiors}>, <\text{ins}>, <\text{yst}>, <\text{Bracket}>, <\text{surround}> \\
        & 20 & <\text{quote}>, <\text{quotes}>, <\text{Quote}>, <\text{quotation}>, <\text{quotations}>, <\text{quote}>, <\text{Quote}>, <\text{quotes}>, <\text{quoting}>, <\text{quoted}> \\
        & 25 & <\text{quotation}>, <\text{quotes}>, <\text{quotations}>, <\text{quote}>, <\text{quoted}>, <\text{Quote}>, <\text{quoting}>, <\text{"\textbackslash "}>, <\text{Quote}>, <\text{Quotes}> \\
        \specialrule{0em}{2pt}{2pt} \\
        \multirow{5}{*}{Number Highlighted Sections} & 3 & <\text{\begin{CJK*}{UTF8}{gbsn}卒\end{CJK*}}>, <\text{adam}>, <\text{rch}>, <\text{doit}>, <\text{plet}>, <\text{Americ}>, <\text{wear}>, <\text{<U+FFFD>}>, <\text{holm}>, <\text{importantly}> \\
        & 10 & <\text{seriousness}>, <\text{unter}>, <\text{Holy}>, <\text{aket}>, <\text{Matcher}>, <\text{Holy}>, <\text{ersh}>, <\text{orra}>, <\text{eller}>, <\text{unker}> \\
        & 15 & <\text{Guides}>, <\text{,}>, <\text{Introduction}>, <\text{cord}>, <\text{section}>, <\text{Pil}>, <\text{chal}>, <\text{Ot}>, <\text{ulla}>, <\text{Technologies}> \\
        & 20 & <\text{\_}>, <\text{**}>, <\text{\#\#\#}>, <\text{\#\#}>, <\text{**}>, <\text{\#\#\#}>, <\text{*\_}>, <\text{Introduction}>, <\text{Welcome}>, <\text{*}> \\
        & 25 & <\text{*}>, <\text{*}>, <\text{*\_}>, <\text{*\_}>, <\text{\_*}>, <\text{,*}>, <\text{*,}>, <\text{.*}>, <\text{(*}>, <\text{-*}> \\
        \specialrule{0em}{2pt}{2pt} \\
        \multirow{5}{*}{English Capital} & 3 & <\text{adam}>, <\text{Dipl}>, <\text{|x}>, <\text{Coal}>, <\text{\foreignlanguage{russian}{стек}}>, <\text{igne}>, <\text{ofi}>, <\text{importantly}>, <\text{akhir}>, <\text{utut}> \\
        & 10 & <\text{orp}>, <\text{eus}>, <\text{-horizontal}>, <\text{vé}>, <\text{readcrumb}>, <\text{Bord}>, <\text{\foreignlanguage{russian}{ург}}>, <\text{porr}>, <\text{ussia}>, <\text{oka}> \\
        & 15 & <\text{-Type}>, <\text{scription}>, <\text{ourse}>, <\text{supremacist}>, <\text{<U+FFFD>}>, <\text{sost}>, <\text{irut}>, <\text{alary}>, <\text{-height}>, <\text{Absolutely}> \\
        & 20 & <\text{UN}>, <\text{AL}>, <\text{IN}>, <\text{AN}>, <\text{IT}>, <\text{ST}>, <\text{DE}>, <\text{AU}>, <\text{ME}>, <\text{YOU}> \\
        & 25 & <\text{IT}>, <\text{HE}>, <\text{IN}>, <\text{WE}>, <\text{IN}>, <\text{YOU}>, <\text{DE}>, <\text{DE}>, <\text{THE}>, <\text{OF}> \\
        \bottomrule
    \end{tabular}
    }
    \label{sup:task_proj_3b}
\end{table*}

\begin{figure*}[!ht]
  \centering
  \includegraphics[width=0.95\linewidth]{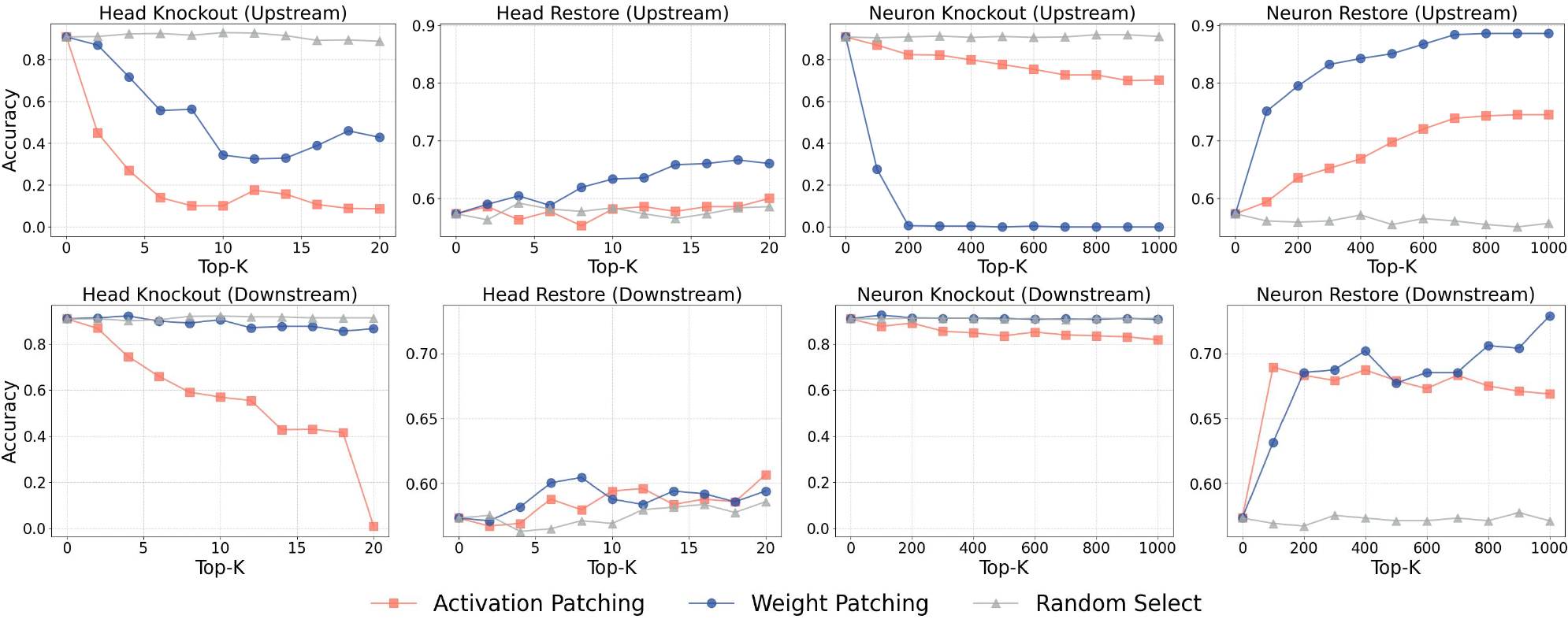}
  \vspace{-10pt}
  \caption{Causal validation of \textbf{Llama-3.2-3B} upstream and downstream modules on the \textbf{Title} task. }
  \label{sup:validation-upstream}
\end{figure*}

All generations used in the interpretability experiments are decoded with \texttt{do\_sample=False} and \texttt{max\_new\_tokens=512}. For model-fusion evaluation, inference uses the default automatic batch scheduling provided by vLLM. All experiments are conducted on three NVIDIA GeForce RTX 4090 GPUs.

\subsection{Ranking, Validation, and Component Selection Rules}
\label{app:repro_rules}

AAP and WAP are used directly for interpretability analysis and module localization, rather than as a preliminary screening step followed by manual top-$k$ curation. In contrast, exact Weight Patching is substantially more expensive, and is therefore applied mainly at the attention-head level. Exact neuron-level WP is only run on a limited set of cases for efficiency comparison, as reported in the main-text efficiency table. This computational gap is one of the main reasons why attribution-based approximations are necessary for fine-grained large-scale analysis.

Unless otherwise noted, the rankings used in heatmaps, layer distributions, and bar-chart summaries are all based on normalized scores. For restore and knockout experiments, the component sets are constructed by directly taking the top-ranked heads or neurons under the corresponding scoring rule, without additional filtering or reranking. In the joint AP+WP restoration experiments, the restored component set is defined as the union of the top-ranked modules identified by the two methods. These rules ensure that all reported restoration and ablation results are derived from a fixed and reproducible protocol rather than post hoc module selection.

\begin{table*}[t]
    \centering
    \caption{Vocabulary projection results of task vectors from shallow to deep layers across six tasks in the \textbf{Llama-3.1-8B} model.}
    \vspace{-8pt}
    \resizebox{1\textwidth}{!}{%
    \begin{tabular}{l c l} 
        \specialrule{1.2pt}{0pt}{0pt}
        
        \rowcolor{mygray}
        \textbf{Task} & \textbf{Layer} & \textbf{Projection (Top-10 vocabulary)} \\
        \hline
        \multirow{6}{*}{No Comma} & 3 & <\text{eca}>, <\text{ientes}>, <\text{holm}>, <\text{airport}>, <\text{ency}>, <\text{\begin{CJK*}{UTF8}{mj}독\end{CJK*}}>, <\text{elter}>, <\text{harma}>, <\text{hsi}>, <\text{Integral}> \\
        & 10 & <\text{enen}>, <\text{nge}>, <\text{bero}>, <\text{sic}>, <\text{458}>, <\text{.usermodel}>, <\text{vester}>, <\text{ome}>, <\text{alah}>, <\text{Wr}> \\
        & 15 & <\text{;amp}>, <\text{129}>, <\text{<U+2510>}>, <\text{Hammer}>, <\text{ober}>, <\text{stra}>, <\text{geme}>, <\text{<U+0E46>}>, <\text{aram}>, <\textgreek{ρίζ}> \\
        & 20 & <\text{engu}>, <\text{some}>, <\text{alternate}>, <\text{first}>, <\text{rita}>, <\text{oko}>, <\text{Lomb}>, <\text{olut}>, <\text{lam}>, <\text{engo}> \\
        & 25 & <\text{arto}>, <\text{folk}>, <\text{undi}>, <\text{oko}>, <\text{U+0627U+0644U+0646U+0627U+0633}>, <\text{stat}>, <\text{KEN}>, <\text{itemprop}>, <\text{ifications}>, <\textgreek{νω}> \\
        & 30 & <\text{we}>, <\text{oh}>, <\text{okay}>, <\text{one}>, <\text{Nicholson}>, <\text{first}>, <\text{\foreignlanguage{russian}{народ}}>, <\text{the}>, <\text{okay}>, <\text{an}> \\
        \specialrule{0em}{2pt}{2pt} \\
        \multirow{6}{*}{Title} & 3 & <\text{\begin{CJK*}{UTF8}{min}ープ\end{CJK*}}>, <\text{reater}>, <\text{foon}>, <\text{bane}>, <\text{levance}>, <\text{itsu}>, <\text{imest}>, <\text{fle}>, <\text{ingleton}>, <\text{rame}> \\
        & 10 & <\text{\foreignlanguage{russian}{сут}}>, <\text{iffe}>, <\text{ancel}>, <\text{ỳ}>, <\text{andin}>, <\text{ieran}>, <\text{zd}>, <\text{adr}>, <\text{ubbo}>, <\text{ADR}> \\
        & 15 & <\text{ěž}>, <\text{MAND}>, <\text{'gc}>, <\text{ANTA}>, <\text{Disposed}>, <\text{/Branch}>, <\text{)prepare}>, <\text{EdgeInsets}>, <\text{/crypto}>, <\text{anja}> \\
        & 20 & <\text{\foreignlanguage{russian}{луги}}>, <\text{addCriterion}>, <\text{\begin{CJK*}{UTF8}{gbsn}《\end{CJK*}}>, <\text{ait}>, <\text{ello}>, <\text{\begin{CJK*}{UTF8}{gbsn}『\end{CJK*}}>, <\text{loser}>, <\text{olare}>, <\text{;}>, <\text{DCALL}> \\
        & 25 & <\text{<<}>, <\text{<<}>, <\text{>>}>, <\text{adel}>, <\text{\foreignlanguage{russian}{надлеж}}>, <\text{stick}>, <\text{sett}>, <\text{addCriterion}>, <\text{>>}>, <\text{\begin{CJK*}{UTF8}{gbsn}芝\end{CJK*}}> \\
        & 30 & <\text{<<}>, <\text{<<}>, <\text{)<<}>, <\text{]<<}>, <\text{«}>, <\text{"<<}>, <\text{()<<}>, <\text{>>,}>, <\text{"<<}>, <\text{<<}> \\
        \specialrule{0em}{2pt}{2pt} \\
        \multirow{6}{*}{Multiple Sections} & 3 & <\text{imest}>, <\text{Mahon}>, <\text{\begin{CJK*}{UTF8}{min}ンフ\end{CJK*}}>, <\text{airport}>, <\text{Anonymous}>, <\text{edImage}>, <\text{ihan}>, <\text{uhl}>, <\text{foon}>, <\text{okens}> \\
        & 10 & <\text{ala}>, <\text{agate}>, <\text{iffe}>, <\text{esser}>, <\text{resse}>, <\text{ipe}>, <\text{Rooney}>, <\text{U+0641U+0634}>, <\text{pope}>, <\text{onte}> \\
        & 15 & <\text{yne}>, <\text{-valu}>, <\text{OTAL}>, <\text{ště}>, <\text{stice}>, <\text{ippi}>, <\text{esco}>, <\text{lak}>, <\text{adesh}>, <\text{\begin{CJK*}{UTF8}{min}くん\end{CJK*}}> \\
        & 20 & <\text{ymoon}>, <\text{olare}>, <\text{;element}>, <\text{incerely}>, <\text{miles}>, <\text{unday}>, <\text{onResponse}>, <\text{EncodingException}>, <\text{continua}>, <\text{\_exempt}> \\
        & 25 & <\text{section}>, <\text{Section}>, <\text{sections}>, <\text{section}>, <\text{\_section}>, <\text{Section}>, <\text{-section}>, <\text{SECTION}>, <\text{(section}>, <\text{.section}> \\
        & 30 & <\text{SECTION}>, <\text{section}>, <\text{SECTION}>, <\text{)section}>, <\text{\_SECTION}>, <\text{section}>, <\text{sections}>, <\text{.section}>, <\text{-section}>, <\text{(section}> \\
        \specialrule{0em}{2pt}{2pt} \\
        \multirow{6}{*}{Quotation} & 3 & <\text{elter}>, <\text{ientes}>, <\text{ovan}>, <\text{holm}>, <\text{\begin{CJK*}{UTF8}{min}ーマ\end{CJK*}}>, <\text{\begin{CJK*}{UTF8}{gbsn}呆\end{CJK*}}>, <\text{Beit}>, <\text{uegos}>, <\text{lv}>, <\text{eca}> \\
        & 10 & <\text{ahoo}>, <\text{enen}>, <\text{<U+0E1EU+0E23>}>, <\text{aal}>, <\text{zung}>, <\text{ehler}>, <\text{\foreignlanguage{russian}{сут}}>, <\text{emailer}>, <\text{hani}>, <\text{\begin{CJK*}{UTF8}{gbsn}午\end{CJK*}}> \\
        & 15 & <\text{ingers}>, <\text{><![}>, <\text{pod}>, <\text{npos}>, <\text{Beard}>, <\text{;amp}>, <\text{<U+0924U+0930>}>, <\text{\foreignlanguage{russian}{імі}}>, <\text{<U+094DU+0930U+0935>}>, <\text{azer}> \\
        & 20 & <\text{quotes}>, <\text{quote}>, <\text{quote}>, <\text{Quotes}>, <\text{quotation}>, <\text{QUOTE}>, <\text{Quotes}>, <\text{iten}>, <\text{-quote}>, <\text{quotes}> \\
        & 25 & <\text{quotes}>, <\text{quotation}>, <\text{quote}>, <\text{Quotes}>, <\text{Quotes}>, <\text{quote}>, <\text{quot}>, <\text{quotations}>, <\text{quoted}>, <\text{quoted}> \\
        & 30 & <\text{quotes}>, <\text{quotation}>, <\text{Quotes}>, <\text{quotations}>, <\text{"\textbackslash "}>, <\text{Quotes}>, <\text{<quote}>, <\text{'"}>, <\text{quote}>, <\text{'"'}> \\
        \specialrule{0em}{2pt}{2pt} \\
        \multirow{6}{*}{Number Highlighted Sections} & 3 & <\text{azen}>, <\text{lie}>, <\text{ottes}>, <\text{Underground}>, <\text{lies}>, <\text{ekil}>, <\text{hsi}>, <\text{imest}>, <\text{rez}>, <\text{indi}> \\
        & 10 & <\text{iffe}>, <\text{azu}>, <\text{resse}>, <\text{U+0641U+0634}>, <\text{hir}>, <\text{esser}>, <\text{\foreignlanguage{russian}{сут}}>, <\text{ikel}>, <\text{\begin{CJK*}{UTF8}{min}ャ\end{CJK*}}>, <\text{erde}> \\
        & 15 & <\text{jev}>, <\text{/Dk}>, <\text{\_mD}>, <\text{\_mC}>, <\text{iesel}>, <\text{Watkins}>, <\text{semiclass}>, <\text{Ģ}>, <\text{didFinish}>, <\text{.updateDynamic}> \\
        & 20 & <\text{ulta}>, <\text{\begin{CJK*}{UTF8}{gbsn}《\end{CJK*}}>, <\textgreek{ρη}>, <\text{\_mD}>, <\text{onda}>, <\text{ither}>, <\text{lotte}>, <\text{ipeg}>, <\text{\_mB}>, <\text{\_}> \\
        & 25 & <\text{\#\#\#}>, <\text{ullo}>, <\text{*}>, <\text{\#\#}>, <\text{\#\#\#}>, <\text{**}>, <\text{natural}>, <\text{ulta}>, <\text{\#\#\#\#}>, <\text{star}> \\
        & 30 & <\text{\#\#\#}>, <\text{*\_}>, <\text{*\_}>, <\text{\_*}>, <\text{\#\#\#}>, <\text{*}>, <\text{>*}>, <\text{*}>, <\text{!*}>, <\text{*",}> \\
        \specialrule{0em}{2pt}{2pt} \\
        \multirow{6}{*}{English Capital} & 3 & <\text{airport}>, <\text{ckt}>, <\text{ottes}>, <\text{\begin{CJK*}{UTF8}{min}ーマ\end{CJK*}}>, <\text{rex}>, <\text{Mahon}>, <\text{audit}>, <\text{foon}>, <\text{\begin{CJK*}{UTF8}{gbsn}秋\end{CJK*}}>, <\text{holm}> \\
        & 10 & <\text{adoo}>, <\text{enen}>, <\text{tingham}>, <\text{<U+FFFD>}>, <\text{\foreignlanguage{russian}{сут}}>, <\text{Tato}>, <\text{stro}>, <\text{<U+0E1EU+0E23>}>, <\text{angen}>, <\text{aida}> \\
        & 15 & <\text{edic}>, <\text{eldorf}>, <\text{\_construct}>, <\text{charge}>, <\text{<U+0E37U+0E2DU+0E02>}>, <\text{riba}>, <\text{tallest}>, <\text{TestMethod}>, <\text{andan}>, <\text{Charge}> \\
        & 20 & <\text{fon}>, <\text{F}>, <\text{,}>, <\text{RE}>, <\text{onic}>, <\text{edin}>, <\text{and}>, <\text{for}>, <\text{surf}> \\
        & 25 & <\text{EX}>, <\text{DE}>, <\text{PO}>, <\text{F}>, <\text{hear}>, <\text{,}>, <\text{isters}>, <\text{PR}>, <\text{.}> \\
        & 30 & <\text{iT}>, <\text{IT}>, <\text{HERE}>, <\text{HERE}>, <\text{HE}>, <\text{IT}>, <\text{OF}>, <\text{CO}>, <\text{HE}>, <\text{OK}> \\
        \bottomrule
    \end{tabular}}
    \label{tab:task_proj_8b}
\end{table*}

\begin{figure*}[!th]
  \centering
  \includegraphics[width=0.95\linewidth]{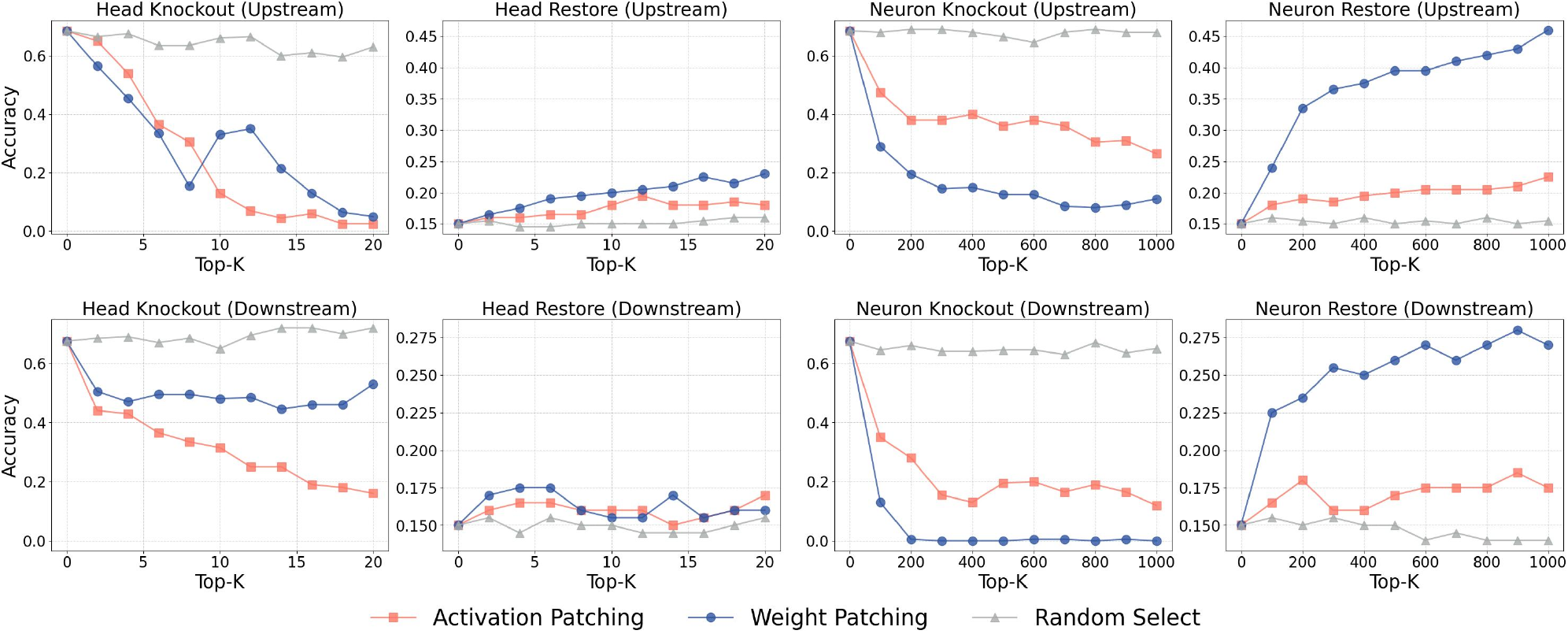}
  \vspace{-10pt}
  \caption{Causal validation of \textbf{Llama-3.1-8B} upstream and downstream modules on the \textbf{Title} task. }
  \label{sup:validation-upstream-8b}
\end{figure*}

\subsection{Implementation Settings for WP-Guided Model Merging}
\label{app:repro_merging_settings}

In the model-fusion experiments, we apply the proposed WP-guided component-wise merging to a set of Llama-2-13B-based expert models (e.g., instruction, math, and code specialists) using the pretrained Llama-2-13B as the shared base model. The detailed implementation settings and hyperparameter choices for the merging pipeline are outlined below.

\begin{table*}[t]
    \centering
    \caption{Vocabulary projection results of task vectors from shallow to deep layers across six tasks in the \textbf{Llama-2-13B} model.}
    \vspace{-8pt}
    \resizebox{1\textwidth}{!}{%
    \begin{tabular}{l c l} 
        \specialrule{1.2pt}{0pt}{0pt}
        \rowcolor{mygray}
        \textbf{Task} & \textbf{Layer} & \textbf{Projection (Top-10 vocabulary)} \\
        \hline
        
        \multirow{7}{*}{No Comma} & 3 & <\text{prob}>, <\text{str}>, <\text{bon}>, <\text{mang}>, <\text{Bon}>, <\text{ling}>, <\text{aud}>, <\text{ieu}>, <\text{Kra}>, <\text{istan}> \\
        & 10 & <\text{egy}>, <\text{Gram}>, <\text{firm}>, <\text{clam}>, <\text{haft}>, <\text{rails}>, <\text{Gal}>, <\text{WF}>, <\text{Wars}>, <\text{Sequ}> \\
        & 15 & <\text{comb}>, <\text{zero}>, <\text{emb}>, <\text{connected}>, <\text{cord}>, <\text{succession}>, <\text{port}>, <\text{God}>, <\text{pair}>, <\text{esch}> \\
        & 20 & <\text{zero}>, <\text{neg}>, <\text{without}>, <\text{Jordan}>, <\text{no}>, <\text{ash}>, <\text{sans}>, <\text{Without}>, <\text{single}> \\
        & 25 & <\text{single}>, <\text{without}>, <\text{un}>, <\text{neg}>, <\text{:}>, <\text{,}>, <\text{negative}>, <\text{Single}> \\
        & 30 & <\text{none}>, <\text{without}>, <\text{single}>, <\text{,}>, <\text{neg}>, <\text{None}>, <\text{acc}>, <\text{ats}>, <\text{tall}>, <\text{no}> \\
        & 35 & <\text{single}>, <\text{none}>, <\text{None}>, <\text{Single}>, <\text{plain}>, <\text{none}>, <\text{FT}>, <\text{lack}>, <\text{absence}>, <\text{single}> \\
        \specialrule{0em}{2pt}{2pt} \\
        \multirow{7}{*}{Title} & 3 & <\text{emot}>, <\text{following}>, <\text{gia}>, <\text{str}>, <\text{prob}>, <\text{fir}>, <\text{mate}>, <\text{s}>, <\text{Kal}>, <\text{LP}> \\
        & 10 & <\text{lam}>, <\text{dup}>, <\text{Riv}>, <\text{<U+028A>}>, <\text{ahu}>, <\text{omo}>, <\text{jam}>, <\text{Tol}>, <\text{fahrt}>, <\text{abase}> \\
        & 15 & <\text{myth}>, <\text{educated}>, <\text{ugel}>, <\text{contradiction}>, <\text{differential}>, <\text{oru}>, <\text{viation}>, <\text{SBN}>, <\text{Mey}>, <\text{fatal}> \\
        & 20 & <\text{double}>, <\text{double}>, <\text{Double}>, <\text{Double}>, <\text{twice}>, <\text{<'}>, <\text{Install}>, <\text{dispos}>, <\text{oul}>, <\text{ux}> \\
        & 25 & <\text{<<}>, <\text{<<}>, <\text{[[}>, <\text{<}>, <\text{[[}>, <\text{\{\{}>, <\text{>>}>, <\text{<}>, <\text{<-}>, <\text{\textasciitilde [}> \\
        & 30 & <\text{<<}>, <\text{<<}>, <\text{double}>, <\text{\{\{}>, <\text{[[}>, <\text{Double}>, <\text{double}>, <\text{Double}>, <\text{\textasciitilde [}>, <\text{<}> \\
        & 35 & <\text{<<}>, <\text{<<}>, <\text{[[}>, <\text{<}>, <\text{[[}>, <\text{<}>, <\text{>>}>, <\text{\{\{}>, <\text{>>}>, <\text{double}> \\
        \specialrule{0em}{2pt}{2pt} \\
        \multirow{7}{*}{Multiple Sections} & 3 & <\text{fol}>, <\text{mang}>, <\text{reli}>, <\text{Ald}>, <\text{ét}>, <\text{chter}>, <\text{ret}>, <\text{respons}>, <\text{respons}>, <\text{atel}> \\
        & 10 & <\text{Sitz}>, <\text{angol}>, <\text{Agu}>, <\text{gram}>, <\text{Gram}>, <\text{\foreignlanguage{russian}{ната}}>, <\text{Kant}>, <\text{Tol}>, <\text{ram}>, <\text{ális}> \\
        & 15 & <\text{outline}>, <\text{adt}>, <\text{dual}>, <\text{sections}>, <\text{CHAP}>, <\text{ált}>, <\text{structor}>, <\text{alse}>, <\text{hren}>, <\text{solem}> \\
        & 20 & <\text{sections}>, <\text{section}>, <\text{Section}>, <\text{section}>, <\text{zero}>, <\text{Section}>, <\text{sections}>, <\text{marked}>, <\text{Zero}>, <\text{ige}> \\
        & 25 & <\text{Section}>, <\text{section}>, <\text{sections}>, <\text{Section}>, <\text{section}>, <\text{sections}>, <\text{SE}>, <\text{SE}>, <\text{sec}>, <\text{§}> \\
        & 30 & <\text{Section}>, <\text{section}>, <\text{Section}>, <\text{section}>, <\text{sections}>, <\text{sections}>, <\text{§}>, <\text{SE}>, <\text{sect}>, <\text{SE}> \\
        & 35 & <\text{SE}>, <\text{section}>, <\text{SE}>, <\text{sections}>, <\text{Section}>, <\text{section}>, <\text{Section}>, <\text{sections}>, <\text{sect}>, <\text{\foreignlanguage{russian}{се}}> \\
        \specialrule{0em}{2pt}{2pt} \\
        \multirow{7}{*}{Quotation} & 3 & <\text{ret}>, <\text{emas}>, <\text{pr}>, <\text{Wa}>, <\text{lass}>, <\text{idos}>, <\text{Roth}>, <\text{Par}>, <\text{dos}>, <\text{marg}> \\
        & 10 & <\text{err}>, <\text{ailable}>, <\text{canton}>, <\text{Europ}>, <\text{TT}>, <\text{WA}>, <\text{yst}>, <\text{tip}>, <\text{ama}>, <\text{alin}> \\
        & 15 & <\text{ba}>, <\text{ere}>, <\text{cord}>, <\text{War}>, <\text{bomb}>, <\text{Bund}>, <\text{interrupted}>, <\text{za}>, <\text{ema}>, <\text{formed}> \\
        & 20 & <\text{double}>, <\text{Double}>, <\text{double}>, <\text{Double}>, <\text{doubles}>, <\text{dou}>, <\text{twice}>, <\text{doub}>, <\text{quot}>, <\text{quotes}> \\
        & 25 & <\text{quot}>, <\text{quotes}>, <\text{quote}>, <\text{quoted}>, <\text{quot}>, <\text{quote}>, <\text{double}>, <\text{double}>, <\text{'"}>, <\text{Double}> \\
        & 30 & <\text{quot}>, <\text{quote}>, <\text{quotes}>, <\text{double}>, <\text{quote}>, <\text{quot}>, <\text{quoted}>, <\text{double}>, <\text{Double}>, <\text{Double}> \\
        & 35 & <\text{quot}>, <\text{""}>, <\text{double}>, <\text{quotes}>, <\text{"'}>, <\text{"""}>, <\text{double}>, <\text{'"}>, <\text{quote}>, <\text{\begin{CJK*}{UTF8}{gbsn}″\end{CJK*}}> \\
        \specialrule{0em}{2pt}{2pt} \\
        \multirow{7}{*}{Number Highlighted Sections} & 3 & <\text{str}>, <\text{ogle}>, <\text{ieu}>, <\text{gia}>, <\text{mate}>, <\text{Constantin}>, <\text{emot}>, <\text{Kal}>, <\text{amma}>, <\text{anc}> \\
        & 10 & <\text{push}>, <\text{ama}>, <\text{Kant}>, <\text{trom}>, <\text{raham}>, <\text{SBN}>, <\text{angol}>, <\text{ci}>, <\text{\foreignlanguage{russian}{ната}}>, <\text{<U+028A>}> \\
        & 15 & <\text{myth}>, <\text{esch}>, <\text{inand}>, <\text{sens}>, <\text{East}>, <\text{ál}>, <\text{ografi}>, <\text{ba}>, <\text{za}>, <\text{organ}> \\
        & 20 & <\text{marked}>, <\text{mark}>, <\text{pill}>, <\text{plain}>, <\text{surrounded}>, <\text{inline}>, <\text{normal}>, <\text{Wolf}>, <\text{\begin{CJK*}{UTF8}{min}ュ\end{CJK*}}>, <\text{normal}> \\
        & 25 & <\text{*}>, <\text{*}>, <\text{(*}>, <\text{(*}>, <\text{*,}>, <\text{"*}>, <\text{*\textbackslash }>, <\text{.*}>, <\text{*.}>, <\text{*.}> \\
        & 30 & <\text{**}>, <\text{**}>, <\text{highlight}>, <\text{*}>, <\text{ital}>, <\text{*}>, <\text{Ital}>, <\text{bold}>, <\text{Ital}>, <\text{ital}> \\
        & 35 & <\text{*}>, <\text{\#\#}>, <\text{**}>, <\text{\_\_}>, <\text{\_\_}>, <\text{**}>, <\text{*}>, <\text{\_}>, <\text{\#\#}>, <\text{\#}> \\
        \specialrule{0em}{2pt}{2pt} \\
        \multirow{7}{*}{English Capital} & 3 & <\text{prob}>, <\text{et}>, <\text{mathemat}>, <\text{pr}>, <\text{dece}>, <\text{convers}>, <\text{ral}>, <\text{idos}>, <\text{prop}>, <\text{mang}> \\
        & 10 & <\text{Gram}>, <\text{uv}>, <\text{alus}>, <\text{WA}>, <\text{anu}>, <\text{<U+1ED9>}>, <\text{trom}>, <\text{álva}>, <\text{ailable}>, <\text{gram}> \\
        & 15 & <\text{JOIN}>, <\text{cord}>, <\text{tall}>, <\text{wide}>, <\text{\begin{CJK*}{UTF8}{gbsn}望\end{CJK*}}>, <\text{pat}>, <\text{Mey}>, <\text{white}>, <\text{AJAX}>, <\text{ük}> \\
        & 20 & <\text{capital}>, <\text{Capital}>, <\text{capit}>, <\text{THE}>, <\text{letters}>, <\text{upper}>, <\text{Capit}>, <\text{TH}>, <\text{CHAPTER}>, <\text{letter}> \\
        & 25 & <\text{capital}>, <\text{capit}>, <\text{Capit}>, <\text{Capital}>, <\text{THE}>, <\text{caps}>, <\text{THE}>, <\text{IT}>, <\text{TH}>, <\text{capitale}> \\
        & 30 & <\text{capital}>, <\text{Capit}>, <\text{capit}>, <\text{Capital}>, <\text{caps}>, <\text{upper}>, <\text{THE}>, <\text{TH}>, <\text{ALL}>, <\text{THE}> \\
        & 35 & <\text{THE}>, <\text{capital}>, <\text{THE}>, <\text{TH}>, <\text{Capital}>, <\text{Capit}>, <\text{IT}>, <\text{capit}>, <\text{UN}>, <\text{AS}> \\
        \bottomrule
    \end{tabular}}
    \label{tab:task_proj_13b}
\end{table*}

\begin{figure*}[!th]
  \centering
  \includegraphics[width=0.95\linewidth]{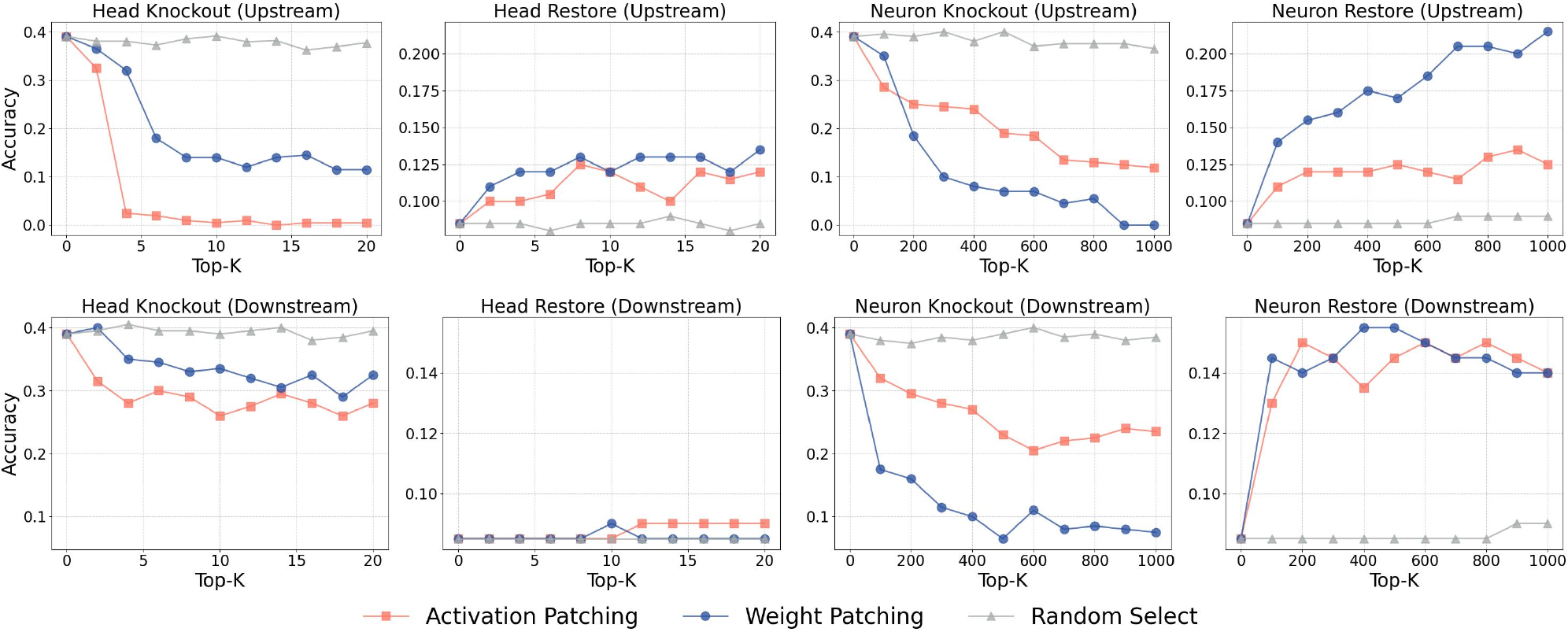}
  \vspace{-10pt}
  \caption{Causal validation of \textbf{Llama-2-13B} upstream and downstream modules on the \textbf{Title} task. }
  \label{sup:validation-upstream-13b}
\end{figure*}

\paragraph{Score Processing and Normalization.}
To compute the fusion weights $\alpha^{(k)}(c)$ for attention heads and MLP neurons, the raw attribution scores extracted from each expert are first passed through a ReLU activation (clamped at a minimum of $0.0$) to filter out negative interference. The non-negative scores are then normalized across the $K$ expert models. If the sum of scores for a specific component across all experts is strictly zero, the merging rule automatically falls back to a uniform distribution (i.e., assigning a weight of $\frac{1}{K}$ to each expert) to ensure parameter integrity.

\begin{figure*}[!th]
  \centering
  \includegraphics[width=0.95\linewidth]{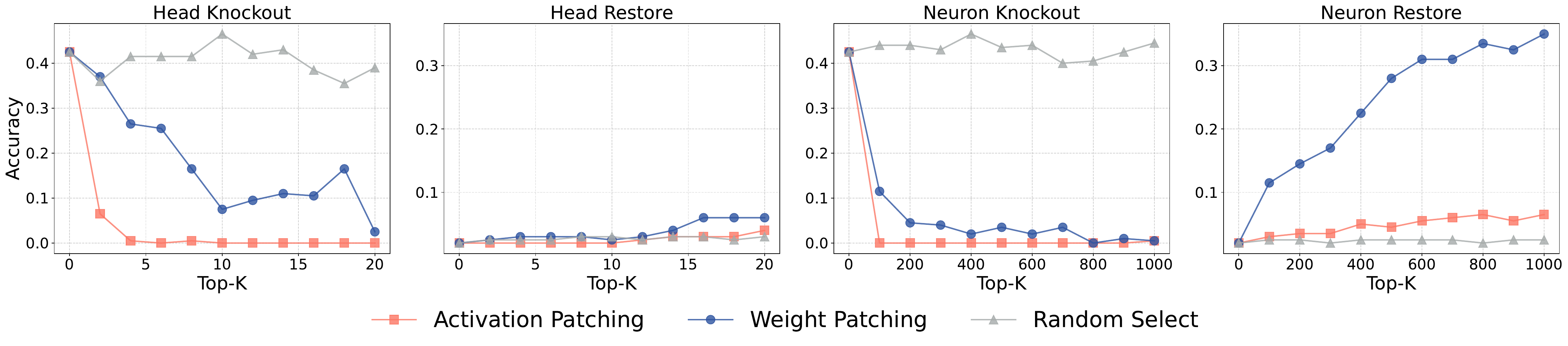}
  \vspace{-10pt}
  \caption{Causal validation of \textbf{Gemma2-2B} global modules on the \textbf{English Capital} task. }
  \label{sup:validation-gemma}
\end{figure*}

\begin{figure*}[!th]
  \centering
  \includegraphics[width=0.95\linewidth]{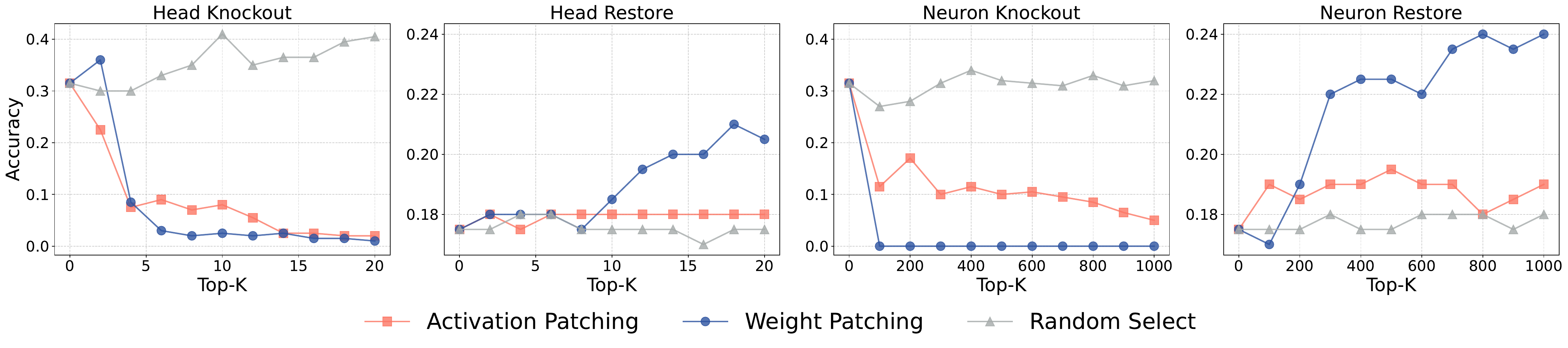}
  \vspace{-10pt}
  \caption{Causal validation of \textbf{Mistral-7B-v0.3} global modules on the \textbf{English Capital} task. }
  
  \label{sup:validation-mistral}
\end{figure*}

\begin{figure*}[!th]
  \centering
  \includegraphics[width=0.95\linewidth]{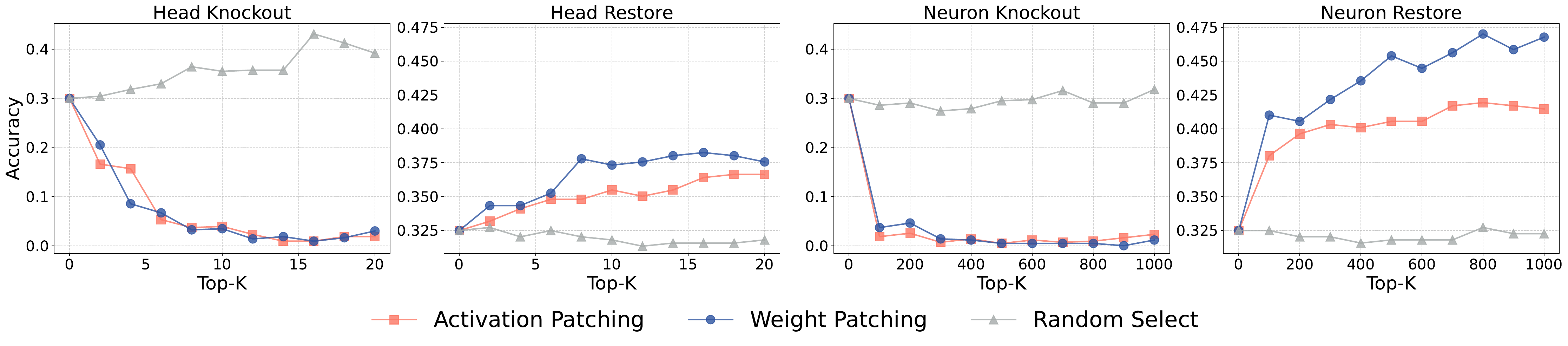}
  \vspace{-10pt}
  \caption{Causal validation of \textbf{Qwen2.5-3B} global modules on the \textbf{English Capital} task. }
  
  \label{sup:validation-qwen25}
\end{figure*}

\paragraph{Component-Level Mapping Rules.}
The normalized fusion weights are applied to the parameter matrices of the Transformer layers based on the following mapping rules:
\begin{itemize}
    \item \textbf{Attention sub-layers:} Head-level fusion weights are explicitly broadcast to the query ($W_Q$) and output ($W_O$) projections. For architectures utilizing Grouped-Query Attention (GQA), the weights for the key ($W_K$) and value ($W_V$) projections are obtained by averaging the fusion weights of the corresponding query heads within each KV group.
    \item \textbf{MLP sub-layers:} Neuron-level fusion weights are directly broadcast to the corresponding rows or columns of the gate, up, and down projections ($W_{\text{gate}}, W_{\text{up}}, W_{\text{down}}$).
\end{itemize}

\paragraph{Fallback Strategy for Structure-Agnostic Parameters.}
Parameters that do not logically decompose into head or neuron components---specifically token embeddings, layer normalizations (e.g., RMSNorm), and the final language modeling head---are merged via uniform parameter averaging across all $K$ experts. During this step, we account for potential vocabulary size mismatches caused by different tokenizer expansions during domain-specific supervised fine-tuning (SFT). If a shape mismatch is detected, we dynamically align the embedding matrices to the base model's original shape by truncating the expanded dimensions prior to averaging.

\paragraph{Hardware and Precision Settings.}
To manage the high memory demands of merging multiple 13B-scale models simultaneously, the state dictionaries of all expert models are loaded sequentially onto CPU memory, and the model objects are immediately garbage-collected to minimize peak RAM usage. The checkpoints are initially loaded in half-precision (FP16). During the layer-wise fine-grained fusion, the weight slices are temporarily cast to FP32 for the weighted accumulation step to prevent numerical underflow and precision loss. Upon completion of a layer's merge, the resulting parameters are cast back to FP16, ensuring that the final fused model remains memory-efficient for downstream deployment.

\section{Extended Experimental Results and Additional Ablations}
\subsection{Full Results for Vector-Anchor Validation}
\label{app:vector-anchor-val}

In Section~\ref{sec:exp_anchor} of the main text, we validated the vector-anchor interface using Llama-3.2-3B, demonstrating that injecting task vectors into mid-to-late layers yields the highest correction rates. To verify that this intermediate representation is a consistent mechanistic property rather than an artifact of a specific model scale, Fig.~\ref{sup:steering_plot} presents the corresponding layer-wise steering correction rates for the larger Llama-3.1-8B and Llama-2-13B models across the same six IFEval tasks. 

Consistent with our observations on the 3B model, the correction rate peaks consistently emerge in the mid-to-late layers across both the 8B and 13B architectures, while early lexical layers show minimal recovery. This cross-model stability further corroborates our conclusion in the main text: the extracted task vector is not merely a surface-level textual proxy, but a robust intermediate control representation. The consistency of these anchor locations justifies our use of a shared internal criterion for subsequent Weight Patching (WP) and Activation Patching (AP) analyses across different models in the Llama family.

\begin{figure*}[!th]
  \centering
  \includegraphics[width=0.95\linewidth]{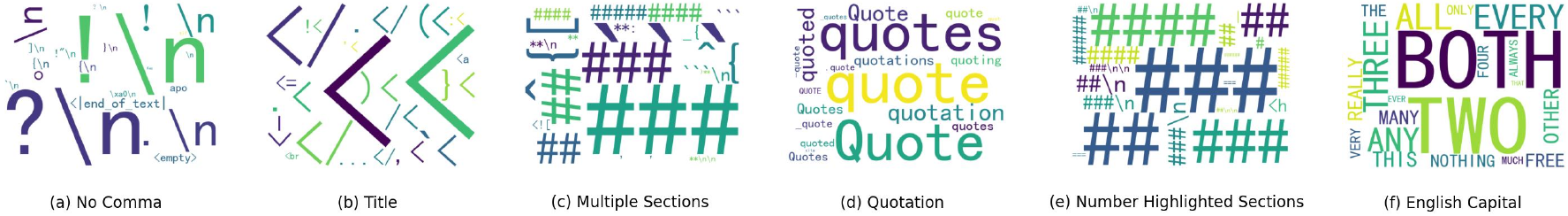}
  \vspace{-10pt}
  \caption{Vocabulary projection results of downstream neurons in the \textbf{Llama-3.2-3B} model, identified via weight patching across six tasks.}
  \label{sup:wp-downstream-word}
\end{figure*}

\begin{figure}[t]
  \centering
  \includegraphics[width=0.95\linewidth]{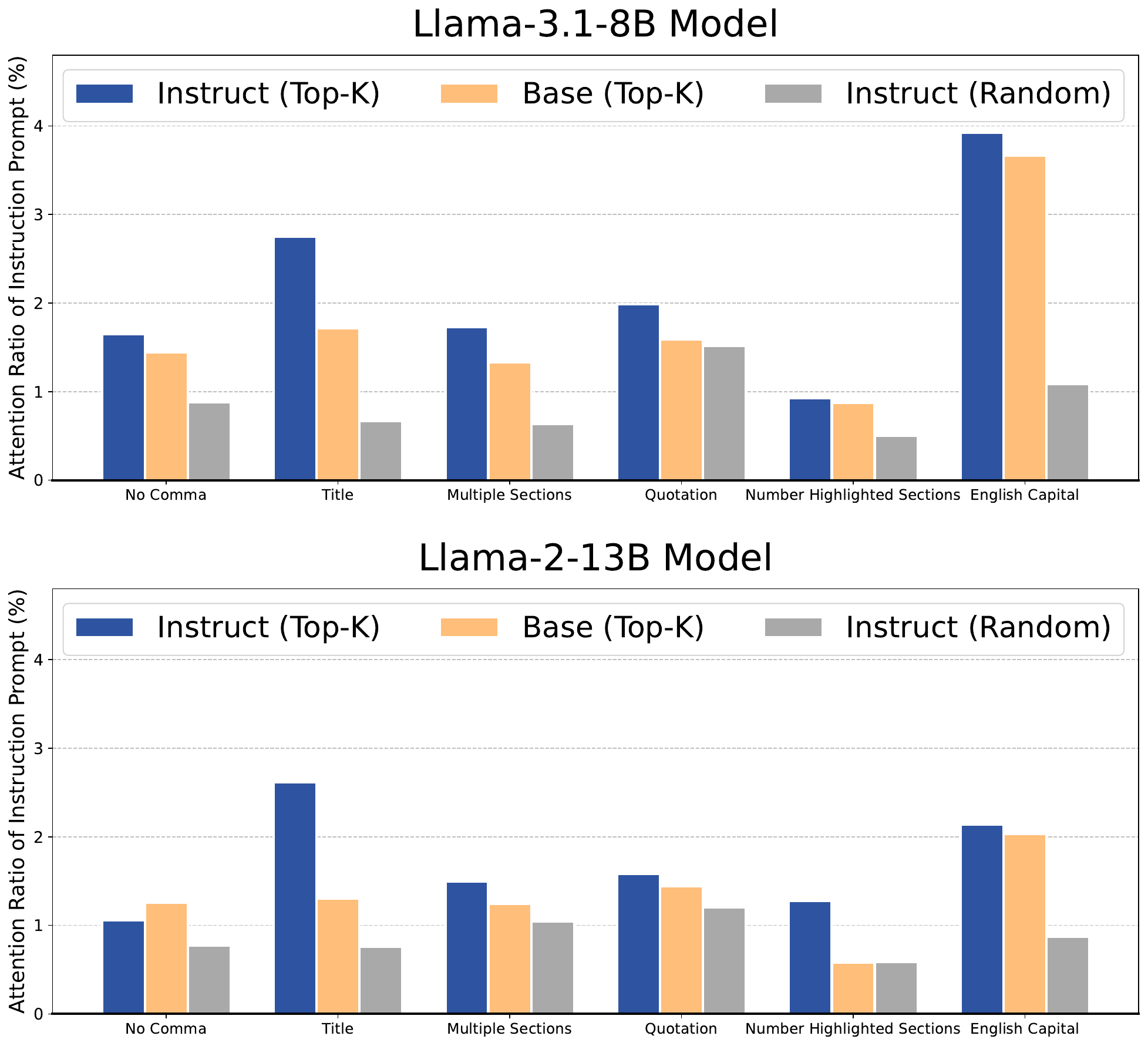}
  \vspace{-10pt}
  
  \caption{Selective attention of critical heads to instruction tokens.}
  \label{sup:attn-instruct}
\end{figure}

\subsection{Extended Results on Vocabulary Projection of Task Vectors}
\label{app:vocab_proj}

In Section~\ref{sec:exp_anchor} of the main paper, we demonstrated that the extracted task vectors serve as a stable, intermediate behavioral interface for instruction following, rather than acting as a fragile, text-level proxy. To substantiate this claim, we discussed the vocabulary-space projection of these vectors. Due to space constraints in the main text, we provide the comprehensive, layer-wise projection results for all six representative IFEval tasks in this section.

\begin{figure}[!th]
  \centering
  \includegraphics[width=\linewidth]{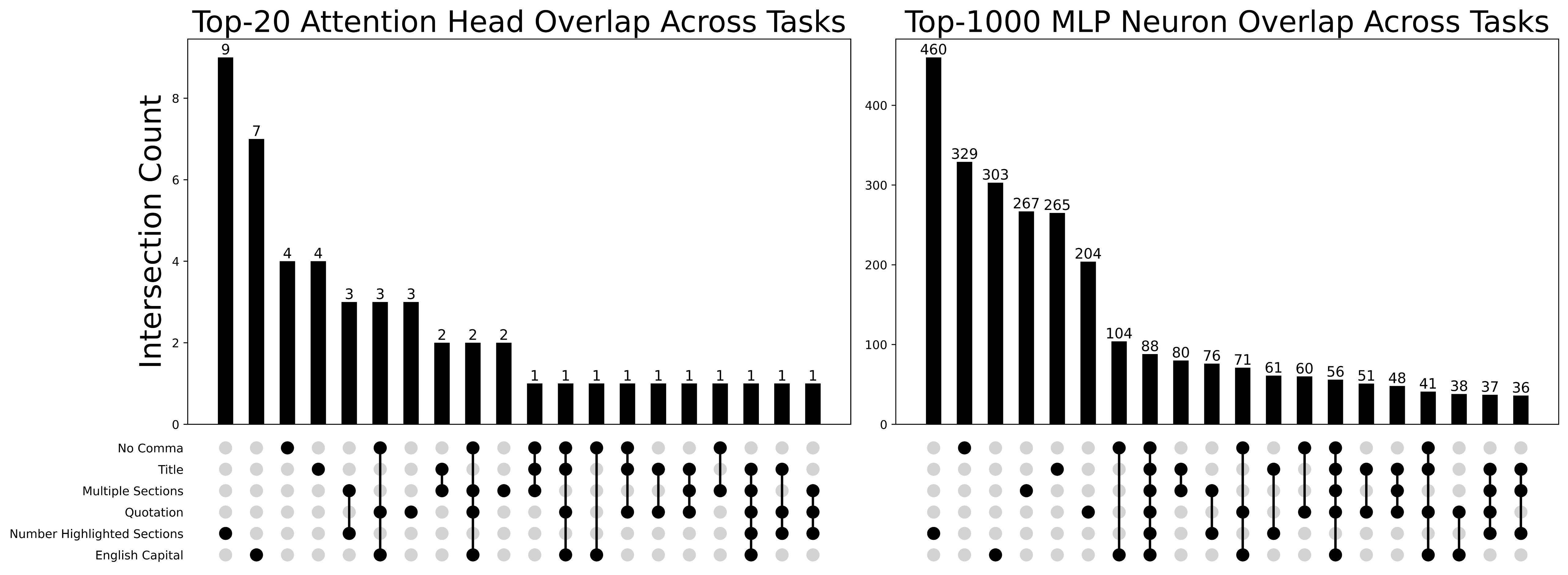}
  \vspace{-15pt}
  \caption{ Overlap of critical modules across tasks on \textbf{Llama-3.1-8B}. }
  \label{sup:task-overlap-8b}
\end{figure}

\begin{figure}[!th]
  \centering
  \includegraphics[width=\linewidth]{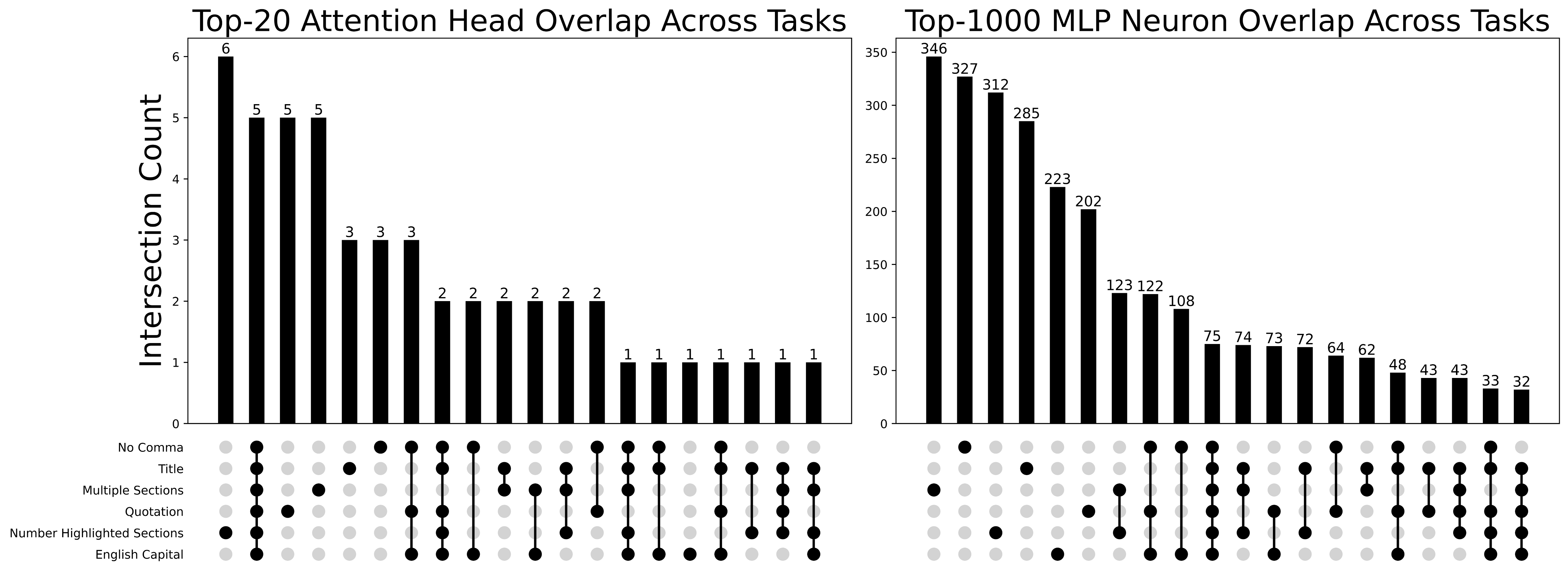}
  \vspace{-15pt}
  \caption{ Overlap of critical modules across tasks on \textbf{Llama-2-13B}. }
  \label{sup:task-overlap-13b}
\end{figure}

Tables~\ref{sup:task_proj_3b}, \ref{tab:task_proj_8b}, and \ref{tab:task_proj_13b} display the top projected tokens from shallow to deep layers for Llama-3.2-3B, Llama-3.1-8B, and Llama-2-13B, respectively. As illustrated in these tables, projecting the intermediate steering vectors into the vocabulary space offers only limited interpretability. While a few structurally explicit tasks occasionally exhibit partially recognizable, format-related cues (e.g., casing-related tokens in \emph{English Capital} or punctuation-related tokens in \emph{No Comma}), the projections for the majority of the tasks and layers remain diffuse, abstract, and largely disconnected from the explicit textual templates of the given prompts.

Crucially, this phenomenon remains highly consistent across different model scales (3B, 8B, and 13B). This robust cross-scale evidence reinforces our conclusion in the main text: the extracted task vector is not merely a surface residue of the prompt memorized in the vocabulary space. Instead, it functions as a deeply embedded, abstract control representation formed after early token-level processing. This validation ensures that the vector anchor reliably reflects the presence of instruction-conditioned control, forming the foundation for our mechanistic comparisons between Activation Patching (AP) and Weight Patching (WP).

\subsection{Extended Localization Heatmaps across Model Scales}
\label{app:extended_heatmaps}

\begin{figure}[!ht]
  \centering
  \includegraphics[width=0.92\linewidth]{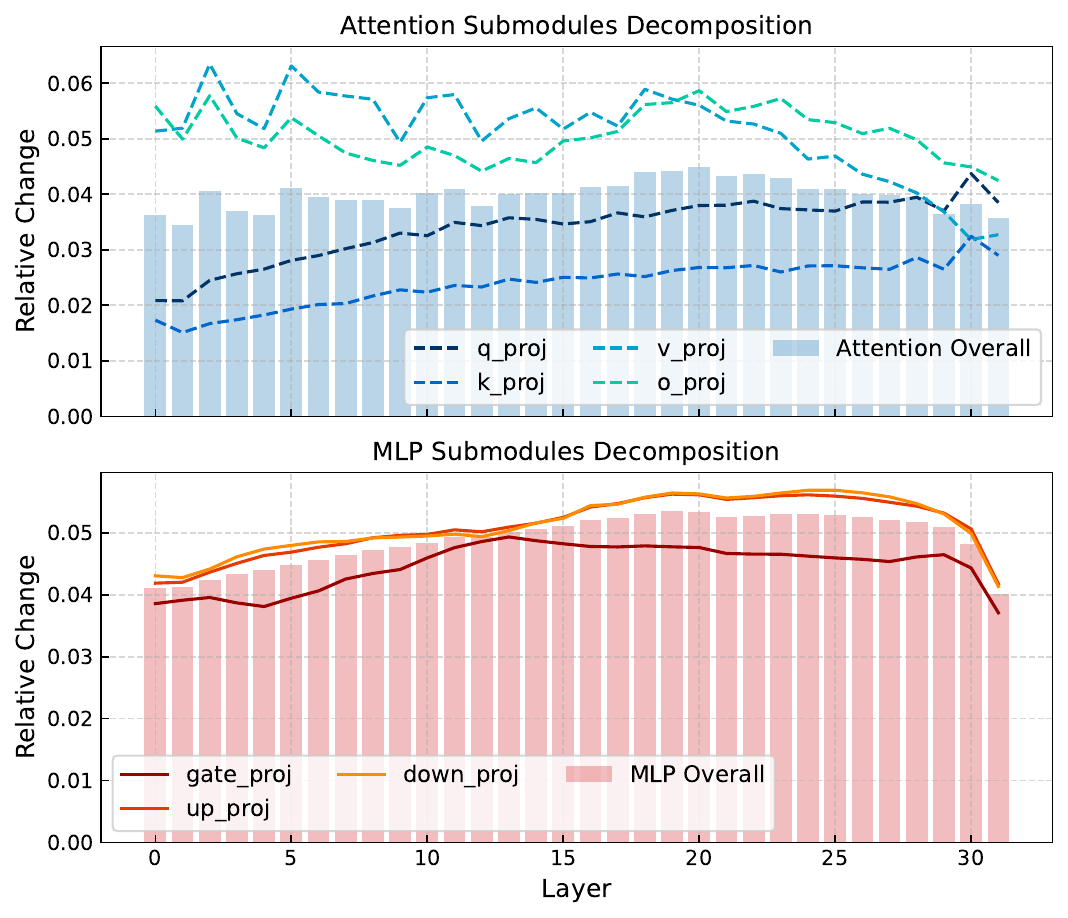}
  \vspace{-10pt}
  
  \caption{ Layer-wise parameter changes induced by instruction tuning on \textbf{Llama-3.1-8B}.}
  \label{sup:para-diff-8b}
\end{figure}

To verify that the source--aggregation separation observed in Section~\ref{sec:exp_source_convergence} is not idiosyncratic to the Llama-3.2-3B architecture, we extend our component-level localization analysis to larger scales. Figures~\ref{sup:heatmap-8b} and~\ref{sup:heatmap-13b} present the component importance heatmaps under both Activation Patching (AP) and Weight Patching (WP) for Llama-3.1-8B and Llama-2-13B, respectively, evaluated on the \emph{English Capital} task.

Consistent with the findings in the main text, the mechanistic division of labor remains structurally stable across different model sizes. In both the 8B and 13B models, Activation Patching (AP) predominantly highlights attention heads and adjacent neurons in the middle-to-late layers. This corroborates the interpretation that these components act as aggregation bottlenecks or routing hubs, whose internal states become critical once the instruction-conditioned information is already propagating during inference.

Conversely, Weight Patching (WP) strongly isolates parameter subsets within shallow-layer MLP neurons across both models. This distinct shift toward early layers reaffirms that the parameter weights most responsible for regenerating the control representation---the source-level capability carriers---are fundamentally distinct from the routing modules. Taken together, these supplementary heatmaps provide robust evidence that the hierarchical separation between parameter sources (revealed by WP) and activation routers (revealed by AP) is a general structural property of instruction-tuned LLMs, generalizing reliably from 3B up to 13B parameters.

\subsection{Source--aggregation Separation Across Model Scales}
\label{app:distribution_scales}

In Sec.~\ref{sec:exp_source_convergence} of the main text, we demonstrated a systematic spatial misalignment between the components localized by Activation Patching (AP) and Weight Patching (WP) on the Llama-3.2-3B model. This misalignment supports a ``source--aggregation separation'' hypothesis, wherein WP identifies shallow source-level parameter carriers, while AP highlights middle-layer routing and aggregation bottlenecks. To verify that this structural division of labor is a general mechanistic property rather than an artifact of a specific model scale, we extend this distributional analysis to larger models.

\begin{figure}[!ht]
  \centering
  \includegraphics[width=0.92\linewidth]{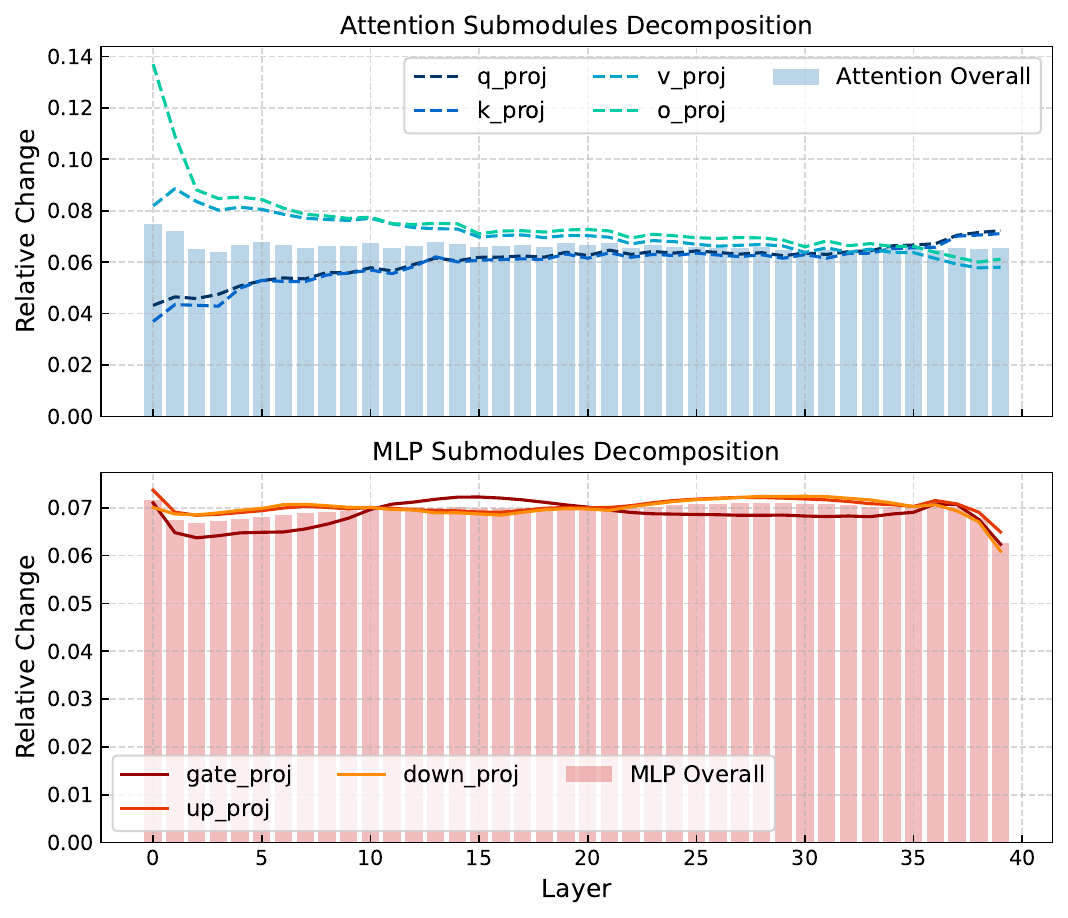}
  \vspace{-10pt}
  
  \caption{ Layer-wise parameter changes induced by instruction tuning on \textbf{Llama-2-13B}.}
  \label{sup:para-diff-13b}
\end{figure}

Figures~\ref{sup:distribution-8b} and~\ref{sup:distribution-13b} illustrate the layer-wise distributions and intersection sets of the top-ranked neurons identified by AP and WP on the \emph{English Capital} task for Llama-3.1-8B and Llama-2-13B, respectively. Consistent with the findings in the 3B model, the distributions remain highly disjoint across both larger scales. Specifically, WP consistently localizes the critical parametric changes to shallow layers, reflecting the early formulation of the control signal. In contrast, AP predominantly highlights neurons in the middle-to-late layers, where the instruction-conditioned information is already actively routed and integrated.

Furthermore, the direct overlap between the top-k components identified by the two intervention methods remains remarkably limited in both the 8B and 13B models. These supplementary results confirm that the observed source--aggregation separation generalizes robustly across different model sizes within the Llama family, reflecting a fundamental architectural regularity in how instruction-tuned LLMs organize procedural control during generation.

\subsection{Additional Causal Validation Results}
\label{app:causal_validation}

In Section~\ref{sec:exp_validation} of the main text, we established a hierarchical division of labor---source-level carriers, aggregation/routing interfaces, and downstream execution units---by analyzing the causal roles of modules on the \emph{English Capital} task using Llama-3.2-3B. To demonstrate that this source--aggregation--execution hierarchy is a robust mechanism rather than an artifact of a specific task or architecture, this section presents supplementary causal validation results along two distinct dimensions: cross-task generalization and cross-model generalization.

\begin{figure*}[htbp]
  \centering
  \includegraphics[width=0.98\linewidth]{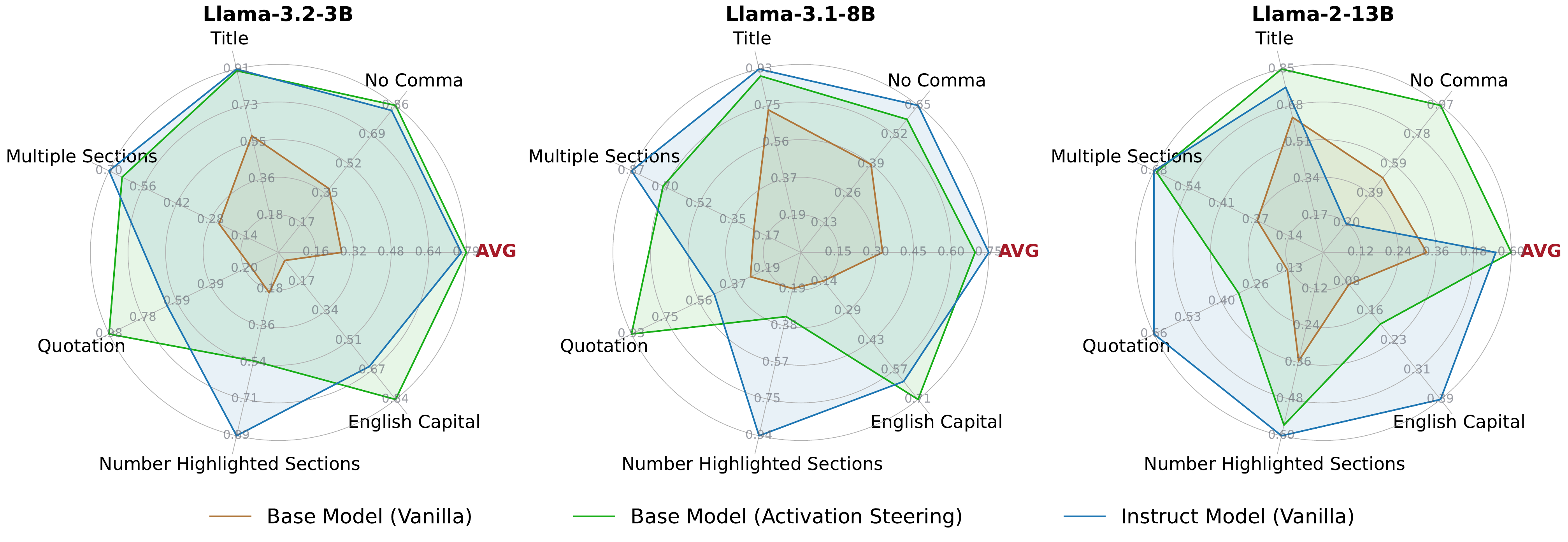}
  \vspace{-10pt}
  \caption{Absolute performance comparison (IFEval strict accuracy) across three model scales: Llama-3.2-3B, Llama-3.1-8B, and Llama-2-13B. The radar charts illustrate the performance of the Vanilla Base Model (brown), the Vanilla Instruct Model (blue), and the Base Model with optimal layer-wise activation steering (green). By injecting the extracted task vector, the base model successfully recovers, and in some cases slightly exceeds, the instruction-following capabilities of the fully fine-tuned instruct model across all six representative tasks.}
  \label{sup:base_steer_instruct}
\end{figure*}

\begin{table}[!th]
    \centering
    \vspace{-5pt}
    \caption{The intersection rate of neurons found by directly applying WP versus tracing WP from the top-20 heads identified by AP (Top-5000 Neurons). Metrics include the maximum overlap (\textit{Max}), the average of the top-5 overlaps (\textit{Top-5 Avg}), and the average overlap across all 20 heads (\textit{Top-20 Avg}).}
    \vspace{-8pt}
    \label{tab:neuron_intersection_top5000}
    
    \setlength{\aboverulesep}{0pt}
    \setlength{\belowrulesep}{0pt}
    \renewcommand{\arraystretch}{1.15} 
    \setlength{\tabcolsep}{4.5pt} 

    \begin{tabular}{lccc} 
        \whline
        \rowcolor{mygray}
        \textbf{Task} & \textbf{Max} & \textbf{Top-5 Avg} & \textbf{Top-20 Avg} \\
        \hline
        
        \multicolumn{4}{c}{\cellcolor{mygray!40}\textbf{Llama-3.2-3B}} \\
        \hline
        English Capital             & $0.761$ & $0.703$ & $0.480$ \\
        Multiple Sections           & $0.403$ & $0.380$ & $0.288$ \\
        Number Highlighted Sections & $0.409$ & $0.388$ & $0.286$ \\
        Title                       & $0.495$ & $0.446$ & $0.337$ \\
        No Comma                    & $0.415$ & $0.354$ & $0.247$ \\
        Quotation                   & $0.696$ & $0.620$ & $0.474$ \\
        \hline
        \textit{Avg}                & $0.530$ & $0.482$ & $0.352$ \\
        \hline
        
        \multicolumn{4}{c}{\cellcolor{mygray!40}\textbf{Llama-3.1-8B}} \\
        \hline
        English Capital             & $0.746$ & $0.676$ & $0.424$ \\
        Multiple Sections           & $0.550$ & $0.484$ & $0.358$ \\
        Number Highlighted Sections & $0.392$ & $0.322$ & $0.247$ \\
        Title                       & $0.533$ & $0.486$ & $0.346$ \\
        No Comma                    & $0.613$ & $0.551$ & $0.401$ \\
        Quotation                   & $0.639$ & $0.533$ & $0.403$ \\
        \hline
        \textit{Avg}                & $0.579$ & $0.509$ & $0.363$ \\
        \hline

        \multicolumn{4}{c}{\cellcolor{mygray!40}\textbf{Llama-2-13B}} \\
        \hline
        English Capital             & $0.635$ & $0.595$ & $0.400$ \\
        Multiple Sections           & $0.446$ & $0.431$ & $0.356$ \\
        Number Highlighted Sections & $0.421$ & $0.399$ & $0.296$ \\
        Title                       & $0.494$ & $0.463$ & $0.370$ \\
        No Comma                    & $0.539$ & $0.520$ & $0.373$ \\
        Quotation                   & $0.595$ & $0.573$ & $0.411$ \\
        \hline
        \textit{Avg}                & $0.522$ & $0.497$ & $0.368$ \\

        \whline
    \end{tabular}
\end{table}

\vspace{0.5em}
\noindent\textbf{Cross-Task and Cross-Scale Generalization.} First, we verify whether the hierarchical organization generalizes to other instruction-following behaviors, and whether this phenomenon remains consistent across different model capacities. To do so, we evaluate the \emph{Title} task across three different scales of the Llama family: Llama-3.2-3B, Llama-3.1-8B, and Llama-2-13B. As illustrated in Figures~\ref{sup:validation-upstream}, \ref{sup:validation-upstream-8b}, and \ref{sup:validation-upstream-13b}, we perform knockout (ablation) and restoration interventions on the top upstream and downstream modules identified by Activation Patching (AP) and Weight Patching (WP), compared against a random baseline. The results strictly corroborate the patterns observed in the main text across all three model sizes:

\textbf{Upstream Modules (Source vs. aggregation):} Across the 3B, 8B, and 13B models, the contrast between upstream heads and neurons remains distinct. Knocking out a small set of top AP-ranked upstream heads severely degrades task performance, yet restoring these same heads yields minimal recovery. This aligns with the conclusion that these middle-layer heads act as aggregation or routing bottlenecks---they are necessary for signal transmission but insufficient to generate the control representation from the base model. In contrast, WP-ranked upstream neurons show both pronounced degradation under knockout and significantly stronger recovery under restoration, reaffirming their role as genuine source-level parameter carriers of the instruction-conditioned signal regardless of model scale.

\textbf{Downstream Modules (Execution stage):} Consistent with the main findings and across varying parameter counts, downstream heads exhibit weak restorative capacity. Downstream neurons, however, demonstrate strong necessity and substantial sufficiency. Knocking out the top-ranked downstream neurons triggers an abrupt collapse in performance, while restoring them recovers a dominant fraction of the target behavior. This further substantiates the execution-stage hypothesis: once the abstract control representation is routed forward, downstream neurons serve as the primary units responsible for translating it into concrete textual outputs.

Overall, the targeted ablation and restoration trajectories on the \emph{Title} task mirror the exact functional signatures found in the \emph{English Capital} task. The consistency of these results across 3B, 8B, and 13B parameter scales provides robust evidence for both task-level and scale-level generalization.

\vspace{0.5em}
\noindent\textbf{Cross-Model Generalization.} Beyond varying the task and scaling parameters within a single lineage, it is crucial to verify whether this mechanistic division of labor is a fundamental property of modern LLMs or merely an idiosyncrasy of the Llama architecture. To this end, we extend our causal validation to three other widely used open-weight model families: Gemma2-2B, Mistral-7B-v0.3, and Qwen2.5-3B. 

Figures~\ref{sup:validation-gemma}, \ref{sup:validation-mistral}, and \ref{sup:validation-qwen25} present the global causal validation trajectories for these models on the \emph{English Capital} task. For this macroscopic analysis, we aggregate modules globally across all layers without the upstream/downstream split. The results demonstrate a remarkably consistent functional dichotomy across all three architectures:

Taken together, these supplementary evaluations confirm that the fundamental mechanistic organization identified in the main text---where sparse MLP neurons act as the genuine parameter-level sources and executors of instruction-conditioned control, while attention heads function as routing bottlenecks---is a general architectural property shared by diverse tasks, scales, and modern instruction-tuned LLMs.

\subsection{Vocabulary-Space Projection of Downstream Neurons}
\label{app:wp_downstream_proj}

In Section~\ref{sec:exp_validation} and Section~\ref{sec:exp_finegrained} of the main text, we established a hierarchical division of labor for instruction-following mechanisms: while upstream neurons act as source-level carriers and middle-layer heads serve as routing bottlenecks, downstream neurons function primarily as execution units that translate abstract control representations into concrete output behavior. To provide qualitative evidence for this execution role, we further investigate the semantic properties of the late-layer neurons localized by Weight Patching (WP).

Specifically, we project the weights of the top-ranked downstream neurons identified by WP on the Llama-3.2-3B model into the vocabulary space. By examining the tokens that maximally align with these neuron weights, we can visualize their functional semantics as word clouds.

\begin{figure*}[!ht]
  \centering
  \includegraphics[width=0.95\linewidth]{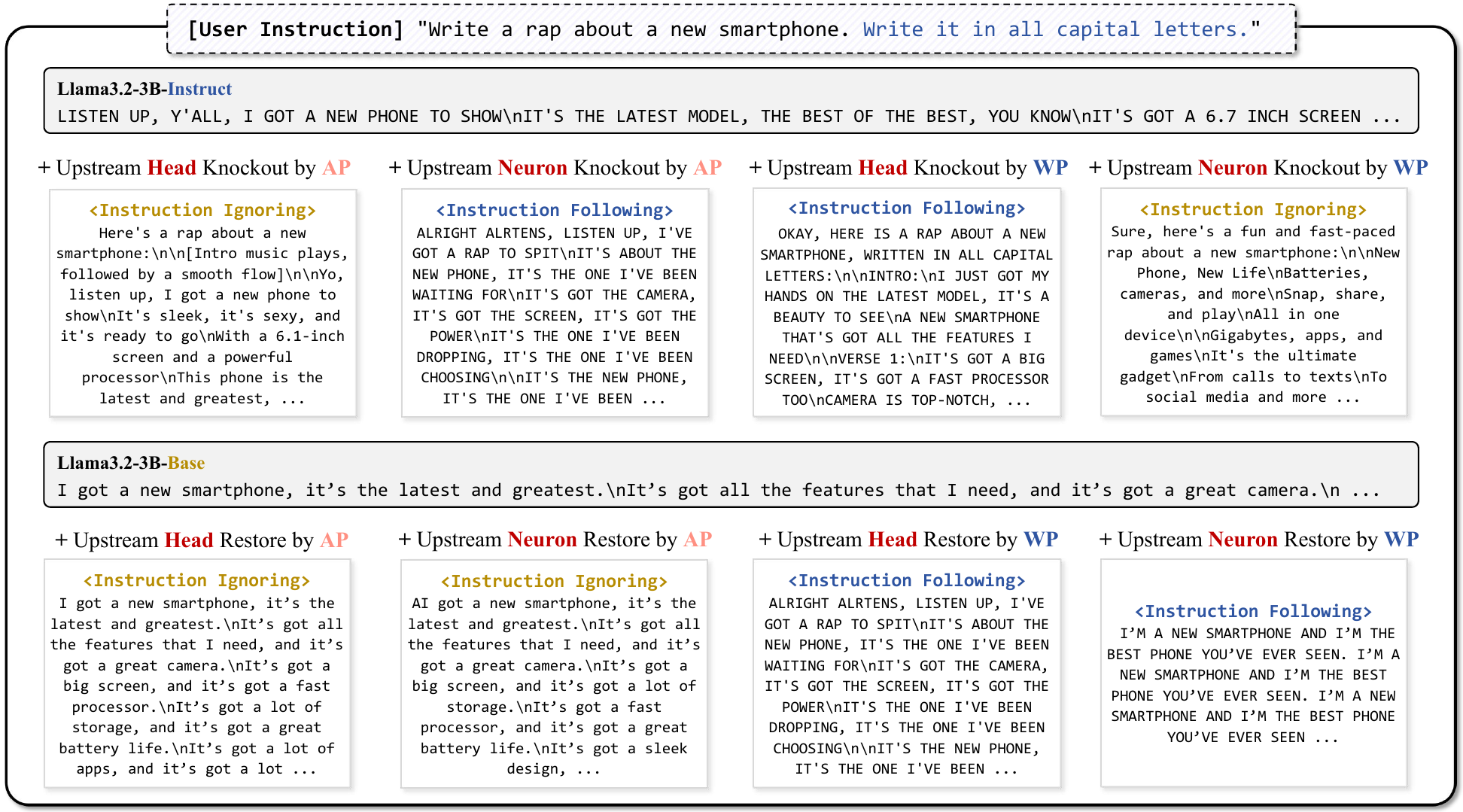}
  \vspace{-10pt}
  \caption{Case study of \textbf{Llama-3.2-3B} on the \textbf{Englist Capital} task. }
  \label{sup:case-capital}
\end{figure*}

\begin{figure*}[!ht]
  \centering
  \includegraphics[width=0.95\linewidth]{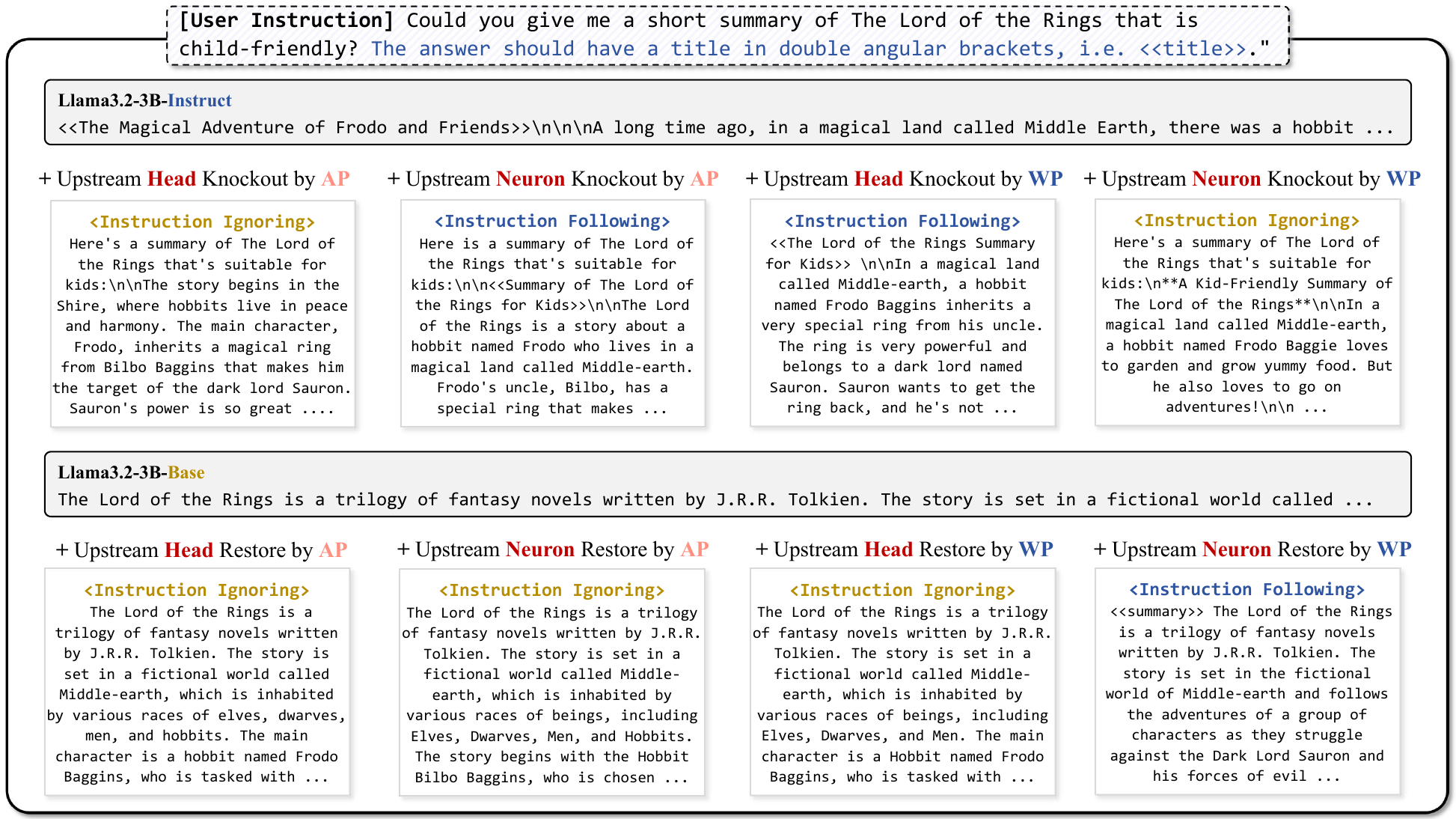}
  \vspace{-10pt}
  \caption{Case study of \textbf{Llama-3.2-3B} on the \textbf{Title} task. }
  \label{sup:case-title}
\end{figure*}

As shown in Fig.~\ref{sup:wp-downstream-word}, the vocabulary-space projections of these downstream neurons exhibit highly recognizable, task-specific lexical patterns across different instruction-following tasks. Unlike the intermediate vector anchors (discussed in Sec.~\ref{sec:exp_anchor}), which often remain diffuse and abstract, the tokens activated by these deep neurons are directly associated with the final formatting, stylistic, or structural constraints of the respective prompts (e.g., capitalization-related tokens for the \emph{English Capital} task). 

This strong alignment with output-space behavior provides direct visual corroboration for our causal and fine-grained analyses. It confirms that the specific parameter subsets localized by WP in the later layers are indeed responsible for the final lexical instantiation of the instruction-conditioned control, further validating the proposed source--aggregation--execution hierarchy.

\subsection{Cross-Scale Validation of Selective Attention in Critical Heads}
\label{app:attn-instruct}

In Section~\ref{sec:exp_finegrained} of the main text, we demonstrated that the critical attention heads identified in the Llama-3.2-3B model exhibit a strong selective preference for instruction-bearing tokens, supporting their functional role as reading and routing interfaces. To verify that this fine-grained functional signature is not an artifact of a specific model size but a general mechanistic property, we extend this attention allocation analysis to larger model scales (Llama-3.1-8B and Llama-2-13B), as presented in Fig.~\ref{sup:attn-instruct}.

Following the same protocol used for the main results, we first localize the top-$k$ critical attention heads for different instruction-following tasks using Weight Patching (WP). To quantify their reading behavior, we compute the ratio of the average attention weight per token allocated to the explicit instruction regions versus the non-instruction regions. We compare this selective attention ratio across three settings: (1) the critical heads in the instruction-tuned model, (2) the structurally matched heads in the corresponding base model, and (3) randomly sampled heads in the instruction-tuned model.

As illustrated in Fig.~\ref{sup:attn-instruct}, the strong selective preference consistently holds across both 8B and 13B models. The critical heads in the instruction-tuned models allocate substantially more attention to the instruction tokens compared to the baselines. In contrast, the exact same heads in the base models and the random heads exhibit a much lower, near-uniform ratio. This pronounced contrast at larger scales corroborates our conclusion in the main text: the attention heads localized by our method are specialized less for storing the downstream capability itself, and more for reading explicit rules from the input and routing this control signal into the internal pathway. This result confirms that the aggregation and routing roles of attention heads within the proposed source--aggregation--execution hierarchy generalize robustly across the Llama model family.

\subsection{Cross-Task Overlap of Critical Modules on Larger Models}
\label{app:cross_task_overlap}

In Section~\ref{sec:exp_generalization} of the main text, we established that while different instruction-following tasks share a common hierarchical organization, they rely on partially distinct local components. This conclusion was drawn from the limited overlap of critical modules localized via Weight Patching (WP) on the Llama-3.2-3B model. To verify whether this shared-vs.-specific structural property generalizes across model scales, we extend the cross-task overlap analysis to the Llama-3.1-8B and Llama-2-13B models.

Fig.~\ref{sup:task-overlap-8b} and Fig.~\ref{sup:task-overlap-13b} illustrate the intersection of the top-$k$ attention heads and sparse neurons across the six representative IFEval tasks for the 8B and 13B models, respectively. Consistent with the observations on the 3B model (Fig.~\ref{fig:task-overlap}), the overlap among top-ranked components remains notably limited across both larger architectures. For both attention heads and MLP neurons, only a negligible fraction of modules are shared across all or most tasks, whereas the majority of the critical components are highly task-specific.

These supplementary results confirm that the limited cross-task component intersection is not an artifact of a specific model capacity, but rather a robust characteristic of how instruction-tuned models encode behaviors. It further reinforces our core mechanistic conclusion: the Llama family models execute instruction-following behaviors through a generalized, shared source--aggregation--execution mechanism, while instantiating this template via distinctly specialized local parameter carriers for different procedural constraints.

\subsection{Layer-wise Parameter Changes across Different Model Scales}
\label{app:para_diff}

In Section~\ref{sec:exp_generalization} of the main text, we provided a static view of the parameter space for Llama-3.2-3B, demonstrating that post-training induces larger parameter shifts in MLP modules than in attention modules. To verify that this structural pattern is not an artifact of a specific model size but a consistent property of the instruction-tuning process, we extend this analysis to larger model scales: Llama-3.1-8B and Llama-2-13B.

Fig.~\ref{sup:para-diff-8b} and Fig.~\ref{sup:para-diff-13b} illustrate the layer-wise and component-wise parameter differences between the base and instruction-tuned models for Llama-3.1-8B and Llama-2-13B, respectively. Consistent with the observations on the 3B model, both larger models exhibit a pronounced disparity in parameter updates between different modules. Specifically, the MLP modules (representing neuron-level parameters) undergo substantially larger average parameter changes across layers compared to the attention modules (representing head-level parameters).

\begin{figure}[!ht]
    \centering

    \includegraphics[width=0.95\linewidth]{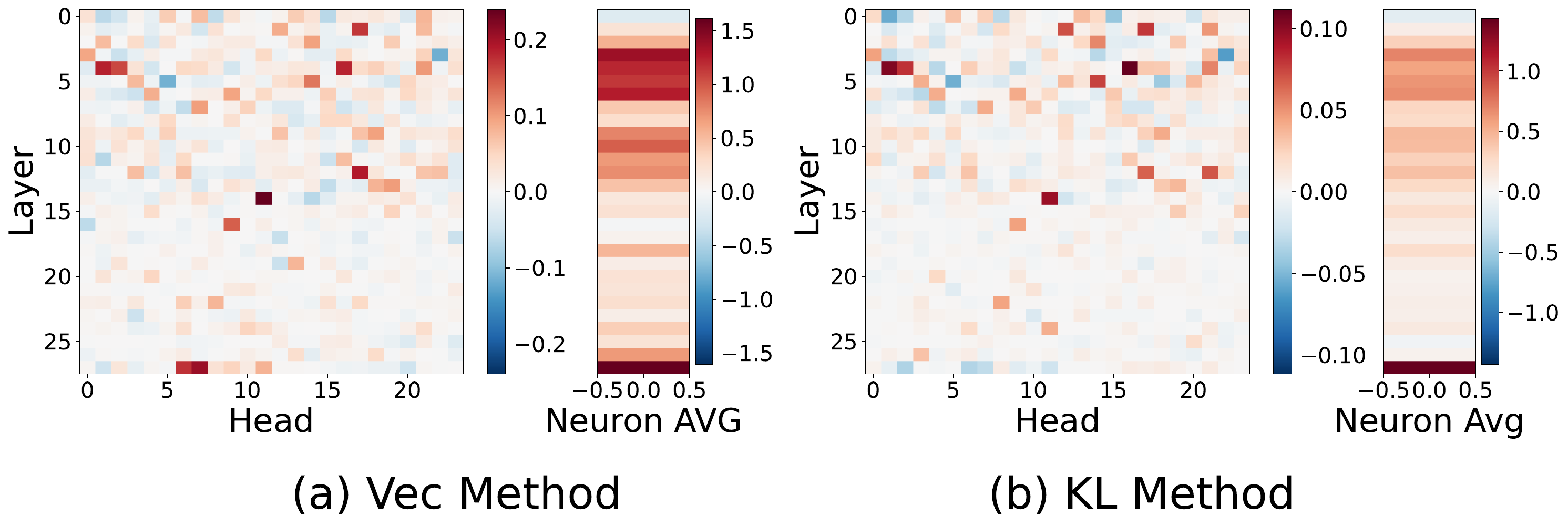}
    \vspace{-10pt}
  
    \caption{
     Heatmaps comparing component localization across distinct metrics (anchor similarity vs. KL divergence) on \textbf{Llama-3.1-8B}.
    }
    \label{sup:heatmap-vector_kl}
\end{figure}

This persistent trend across different model capacities and Llama generations further corroborates our causal conclusion derived from Weight Patching (WP): the acquisition of instruction-following behavior during post-training relies more heavily on writing capabilities into sparse, neuron-centric parameter subsets rather than uniformly updating the attention routing mechanisms. The consistency of these static parameter drift patterns provides strong supplementary evidence for the source--aggregation hierarchy discussed in the main text.

\subsection{Extended Analysis of Upstream Neuron Intersection}
\label{app:neuron_intersection_extended}

In the main text (Sec.~\ref{sec:exp_validation}), we demonstrated a structural linkage between the middle-layer aggregation heads identified by Activation Patching (AP) and the shallow source neurons identified by Weight Patching (WP). To ensure that this systematic alignment is robust and not merely an artifact of a specific, stringent sparsity threshold, we extend our upstream tracing analysis here to a broader population of components.

Specifically, Table~\ref{tab:neuron_intersection_top5000} presents the intersection rates when expanding the retrieval scale to the top-5000 neurons. Following the exact protocol described in the main paper, we trace the upstream neuronal suppliers for the top-20 AP-ranked heads using WP, and compute their overlap with the top-5000 neurons directly localized by WP. 

Consistent with our main findings, the results demonstrate that the substantial overlap is preserved even at this expanded scale. The intersection remains highly stable across multiple aggregation metrics, including the best-matching head (\textit{Max}), the top-5 heads average (\textit{Top-5 Avg}), and the overall average across all 20 heads (\textit{Top-20 Avg}). These extended results further corroborate the proposed hierarchical division of labor. They verify that the WP-recovered source neurons systematically act as upstream suppliers to the aggregation heads, and that this source-to-aggregation routing pathway is a stable, large-scale structural property of the instruction-following mechanism.

\subsection{Absolute Performance of the Vector-Anchor Interface}
\label{app:anchor_absolute}

In Section \ref{sec:exp_anchor} of the main text, we introduced the correction rate to quantify the fraction of the base-to-instruct performance gap recovered by activation steering. To provide a more comprehensive and absolute perspective on this validation step, Fig.~\ref{sup:base_steer_instruct} details the raw IFEval strict accuracy scores across all three evaluated model families (Llama-3.2-3B, Llama-3.1-8B, and Llama-2-13B) and all six representative instruction-following tasks.

\begin{figure}[!ht]
    \centering

    \includegraphics[width=0.95\linewidth]{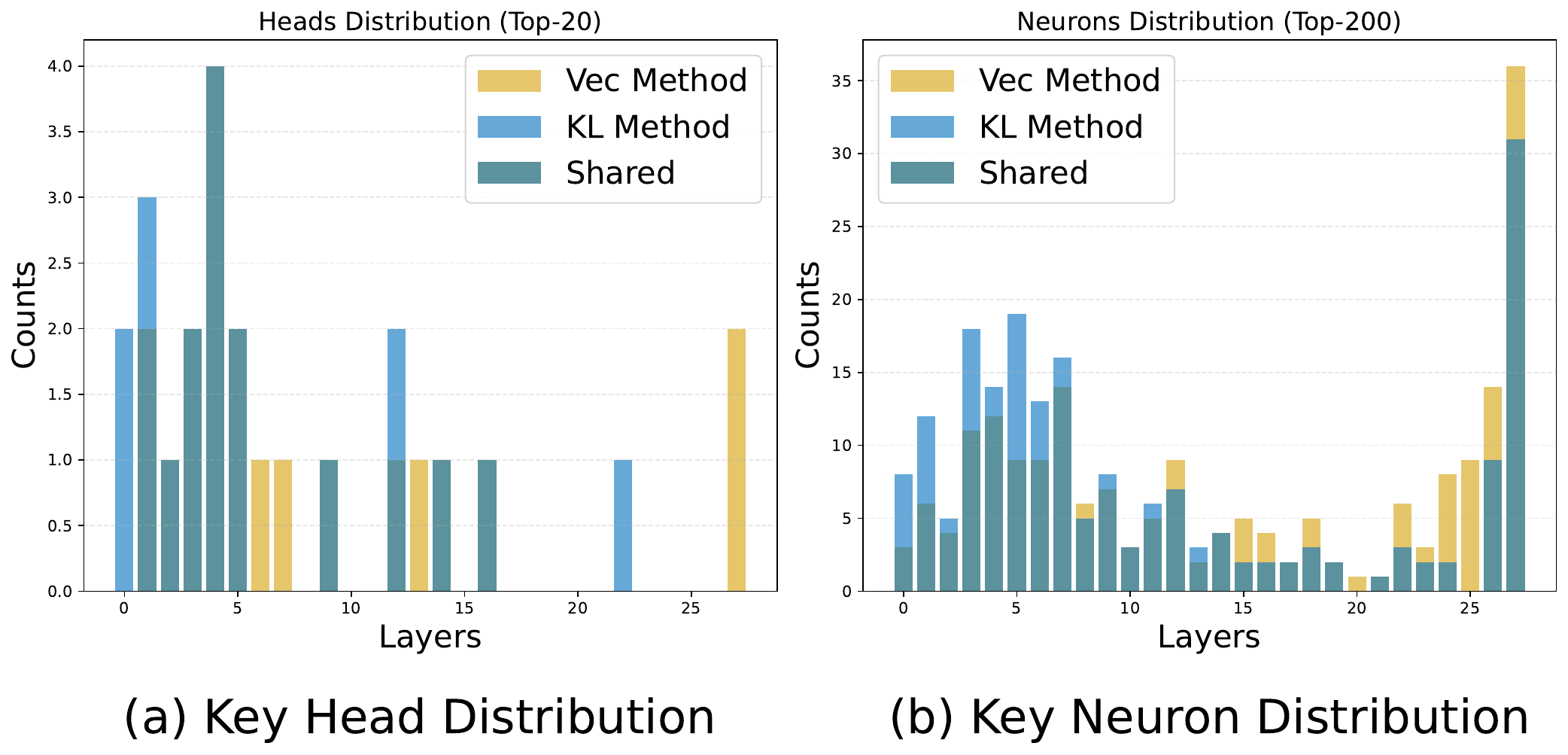}
    \vspace{-10pt}
  
    \caption{
    Layer distributions and overlaps of top-ranked components identified across distinct metrics (anchor similarity vs. KL divergence) on \textbf{Llama-3.1-8B}.
    }
    \label{sup:dist-vector_kl}
\end{figure}

As demonstrated in the radar charts, the vanilla base models (brown lines) inherently lack the ability to consistently follow explicit formatting constraints, often yielding near-zero strict accuracy. The instruction-tuned models (blue lines) establish the empirical upper bound for these specific capabilities. When the extracted task vector is injected into the identified optimal anchor layer of the base model (green lines), we observe a dramatic and consistent behavioral shift. Across all scales and tasks, the simple activation steering intervention successfully bridges the massive performance gap, closely approximating the behavior of the instruction-tuned models. 

Notably, in several instances—such as the \emph{Title} and \emph{No Comma} tasks for Llama-3.2-3B, and the \emph{Title} task for Llama-2-13B—the steered base model even slightly outperforms the vanilla instruct model. This phenomenon further corroborates our central premise in Section \ref{sec:exp_anchor}: the extracted task vector effectively captures a compact and pure control representation for instruction following, largely devoid of conflicting signals. These absolute performance gains across multiple model scales confirm that the vector-anchor interface provides a reliable and stable criterion for the subsequent comparative analyses of Activation Patching and Weight Patching.

\subsection{Case Studies.} 

To intuitively ground the quantitative causal validation results, Figures~\ref{sup:case-capital} and \ref{sup:case-title} present text generation case studies using Llama-3.2-3B on the \emph{English Capital} and \emph{Title} tasks, respectively. The generated outputs visually corroborate the source--aggregation dichotomy. In the knockout scenarios (top panels), ablating either the AP-ranked upstream heads or the WP-ranked upstream neurons causes the Instruct model to ignore the formatting constraints, confirming the necessity of both modules. However, a stark contrast emerges in the restoration scenarios (bottom panels). Injecting the AP-identified modules back into the Base model completely fails to induce instruction-following behavior. Conversely, restoring the WP-identified modules (particularly the WP-ranked neurons) successfully enables the Base model to generate the correct format (e.g., producing text in all capital letters or enclosing the title in double angular brackets). These qualitative examples explicitly demonstrate that Weight Patching successfully locates the genuine parameter-level carriers of the capability, whereas Activation Patching highlights necessary but insufficient routing bottlenecks.

\subsection{Consistency of Localization Across Evaluation Metrics}
\label{app:metric_consistency}

In our main methodology, we utilized the similarity to the intermediate vector anchor (the ``Vec Method'') as the primary evaluation metric during our patching experiments to isolate critical parameters and activations. To verify that our component localization results are robust and not an artifact of this specific proxy metric, we compare it against the traditional Kullback-Leibler (KL) divergence metric (the ``KL Method''), which measures the distributional shift over the entire output vocabulary space. 

Figure~\ref{sup:heatmap-vector_kl} presents a macroscopic comparison of the component importance heatmaps for both attention heads and MLP neurons on the Llama-3.1-8B model (evaluated on the \emph{English Capital} task). As visually evident, the spatial distributions of critical components are nearly identical across the two metrics. The exact same specific heads and shallow/deep neurons are highlighted as highly influential, regardless of whether the intervention effect is measured via internal structural similarity (Vec Method) or final output probability shift (KL Method).

To strictly quantify this alignment, Figure~\ref{sup:dist-vector_kl} illustrates the layer-wise distributions and the direct intersection of the top-ranked components (Top-20 for heads, Top-200 for neurons) identified by both methods. The bar charts demonstrate a remarkably high degree of overlap (represented by the dominant ``Shared'' teal bars). Both metrics precisely localize the same intermediate attention heads and the identical sparse subsets of source and execution MLP neurons. 

These supplementary results provide strong methodological validation. They confirm that the vector-anchor interface introduced in our framework is not only a valid proxy for traditional logit-based metrics like KL divergence, but it also yields highly stable and robust component localizations for tracing instruction-conditioned control mechanisms.

\end{document}